\documentclass{article}

\usepackage[preprint]{neurips_data_2022}

\usepackage[utf8]{inputenc} %
\usepackage[T1]{fontenc}    %
\usepackage{url}            %
\usepackage{booktabs}       %
\usepackage{amsfonts}       %
\usepackage{nicefrac}       %
\usepackage{microtype}      %
\usepackage[x11names]{xcolor}         %

\usepackage{caption}
\usepackage{subcaption}
\usepackage{mathtools}
\usepackage{colortbl}
\usepackage[export]{adjustbox}
\usepackage{tabularray}
\usepackage{tikzducks}
\usepackage{bbm}

\usepackage{graphbox} %
\usepackage{graphicx,booktabs,array}

\definecolor{bubblegum}{rgb}{0.99, 0.76, 0.8}
\definecolor{cambridgeblue}{rgb}{0.64, 0.76, 0.68}
\definecolor{lavender}{rgb}{0.69, 0.61, 0.85}
\definecolor{bananamania}{rgb}{0.98, 0.91, 0.71}
\definecolor{darkgray}{rgb}{0.6352941176470588, 0.6196078431372549, 0.6196078431372549}
\definecolor{darkgraylighter}{rgb}{0.9058823529411765, 0.9019607843137255, 0.9019607843137255}
\definecolor{palegray}{rgb}{0.9607843137254902, 0.9607843137254902, 0.9607843137254902}
\definecolor{lightbluewilds}{rgb}{0.8705882352941177, 0.9215686274509803, 0.9686274509803922}
\definecolor{lightpurplej}{rgb}{0.796078431372549, 0.7647058823529411, 0.8901960784313725}
\definecolor{lightgreenwilds}{rgb}{0.8862745098039215, 0.9450980392156862, 0.8509803921568627}
\definecolor{lightredwilds}{rgb}{1.0, 0.8705882352941177, 0.8901960784313725}
\definecolor{lightorangewilds}{rgb}{0.9803921568627451, 0.9019607843137255, 0.8392156862745098}
\definecolor{lightyellowwilds}{rgb}{1.0, 0.9450980392156862, 0.8}
\definecolor{lightpinkj}{rgb}{1.0, 0.8, 0.9568627450980393}

\newcommand\thickhline{\Xhline{3pt}}
\usepackage{makecell}
\usepackage[frozencache,cachedir=minted-cache]{minted}
\setminted[python]{breaklines}
\DeclarePairedDelimiter\floor{\lfloor}{\rfloor}
\usepackage{placeins}
\usepackage{bibunits}
\usepackage{placeins}
\setcitestyle{square,comma,numbers,sort}
\usepackage{hyperref}       %

\newtheorem{definition}{Definition}

\title{FLamby: Datasets and Benchmarks for Cross-Silo Federated Learning in Realistic Healthcare Settings}
\author{%
Jean Ogier du Terrail$^{1}$ \quad Samy-Safwan Ayed$^{2}$ \quad Edwige Cyffers$^3$ \quad Felix Grimberg$^4$ \\
\textbf{Chaoyang He}$^5$ \quad \textbf{Regis Loeb}$^1$ \quad \textbf{Paul Mangold}$^3$ \quad \textbf{Tanguy Marchand}$^1$ \\
\textbf{Othmane Marfoq}$^2$ \quad \textbf{Erum Mushtaq}$^6$ \quad \textbf{Boris Muzellec}$^1$ \quad \textbf{Constantin Philippenko}$^7$ \\
\textbf{Santiago Silva}$^2$ \quad \textbf{Maria Tele\'{n}czuk}$^1$ \quad \textbf{Shadi Albarqouni}$^{8, 9}$ \textbf{Salman Avestimehr}$^{5,6}$ \\
\textbf{Aurélien Bellet}$^3$ \quad \textbf{Aymeric Dieuleveut}$^7$ \quad \textbf{Martin Jaggi}$^4$ \\
\textbf{Sai Praneeth Karimireddy}$^{10}$ \textbf{Marco Lorenzi}$^2$ \quad \textbf{Giovanni Neglia}$^2$ \quad \textbf{Marc Tommasi}$^3$ \\
\textbf{Mathieu Andreux}$^1$ \\
$^1$Owkin, Inc, $^2$ Inria, Université C\^{o}te d’Azur, Sophia Antipolis, France \\ 
$^3$Univ. Lille, Inria, CNRS, Centrale Lille, UMR 9189 - CRIStAL, F-59000 Lille, France \\
$^4$EPFL \quad $^5$FedML, Inc. \quad $^6$University of Southern California \\
$^7$CMAP, UMR 7641, École Polytechnique, Institut Polytechnique de Paris \\
$^8$University Hospital Bonn \quad $^9$Helmholtz Munich \quad $^{10}$University of California, Berkeley\\
\texttt{\{jean.du-terrail, regis.loeb, tanguy.marchand, boris.muzellec, } \\
\texttt{maria.telenczuk, mathieu.andreux\}@owkin.com, \{samy-safwan.ayed, } \\
\texttt{edwige.cyffers, paul.mangold, othamne.marfoq, } \\
\texttt{santiago-smith.silva-rincon, aurelien.bellet, } \\
\texttt{marco.lorenzi, giovanni.neglia, marc.tommasi\}@inria.fr} \\
\texttt{\{felix.grimberg, martin.jaggi\}@epfl.ch},\\
\texttt{ch@fedml.ai}, \texttt{\{emushtaq, avestime\}@usc.edu},\\
\texttt{\{constantin.philippenko, aymeric.dieuleveut\}@polytechnique.edu},\\
\texttt{shadi.albarqouni@ukbonn.de}, \texttt{sp.karimireddy@berkeley.edu}\\
}

\begin{document}
\maketitle
\begin{bibunit}[plain]
\begin{abstract}
Federated Learning (FL) is a novel approach enabling several clients holding sensitive data to collaboratively train machine learning models, without centralizing data.
The cross-silo FL setting corresponds to the case of few ($2$--$50$) reliable clients,
each holding medium to large datasets,
and is typically found in applications such as healthcare, finance, or industry.
While previous works have proposed representative datasets for cross-device FL, few realistic healthcare cross-silo FL datasets exist, thereby slowing algorithmic research in this critical application.
In this work, we propose a novel cross-silo dataset suite focused on healthcare, FLamby (Federated Learning AMple Benchmark of Your cross-silo strategies), to bridge the gap between theory and practice of cross-silo FL.
FLamby encompasses~7 healthcare datasets with natural splits, covering multiple tasks, modalities, and data volumes, each accompanied with baseline training code.
As an illustration, we additionally benchmark standard FL algorithms on all datasets.
Our flexible and modular suite allows researchers to easily download datasets, reproduce results and re-use the different components for their research.
FLamby is available at~\url{www.github.com/owkin/flamby}.
\end{abstract}
\section{Introduction \label{sec:intro}}
Recently it has become clear that, in many application fields, impressive machine learning (ML) task performance can be reached by scaling the size of both ML models and their training data while keeping existing well-performing architectures mostly unaltered~\cite{sun2017revisiting, kaplan2020scaling, chowdhery2022palm, dalle2}.
In this context, it is often assumed that massive training datasets can be collected and centralized in a single client in order to maximize performance.
However, in many application domains, data collection occurs in distinct sites (further referred to as clients, e.g., mobile devices or hospitals), and the resulting local datasets cannot be shared with a central repository or data center due to privacy or strategic concerns~\cite{dwork2014algorithmic, burki2019pharma}.

To enable cooperation among clients given such constraints, Federated Learning (FL)~\cite{mcmahan2017communication, kairouz2019advances} has emerged as a viable alternative to train models across data providers without sharing sensitive data.
While initially developed to enable training across a large number of small clients, such as smartphones or Internet of Things (IoT) devices,
it has been then extended to the collaboration of fewer and larger clients, such as banks or hospitals.
The two settings are now respectively referred to as \textit{cross-device} FL and \textit{cross-silo} FL,
each associated with specific use cases and challenges~\cite{kairouz2019advances}.

On the one hand, cross-device FL leverages edge devices such as mobile phones and wearable technologies to exploit data distributed over billions of data sources~\cite{mcmahan2017communication, bonawitz2017practical,bhowmick2018protection, niu2020billion}. Therefore, it often requires solving problems related to edge computing~\cite{he2020group, lim2020federated, xia2021survey}, participant selection~\cite{kairouz2019advances, yang2021achieving, charles2021large,fraboni2021impact}, system heterogeneity~\cite{kairouz2019advances}, and communication constraints such as low network bandwidth and high latency~\cite{sattler2019sparse, lu2020low, haddadpour2021federated}.
On the other hand, cross-silo initiatives enable to untap the potential of large datasets previously out of reach.
This is especially true in healthcare, where the emergence of federated networks of private and public actors~\cite{rieke2020future, sheller2020federated, pati2021federated}, for the first time, allows scientists to gather enough data to tackle open questions on poorly understood diseases such as triple negative breast cancer~\cite{du2021collaborative} or COVID-19~\cite{dayan2021federated}.
In cross-silo applications, each silo has large computational power, a relatively high bandwidth, and a stable network connection, allowing it to participate to the whole training phase.
However, cross-silo FL is typically characterized by high inter-client dataset heterogeneity and biases of various types across the clients~\cite{pati2021federated, du2021collaborative}.

As we show in Section~\ref{sec:related_work}, publicly available datasets for the cross-silo FL setting are scarce.
As a consequence, researchers usually rely on heuristics to artificially generate heterogeneous data partitions from a single dataset and assign them to hypothetical clients.
Such heuristics might fall short of replicating the complexity of natural heterogeneity found in real-world datasets.
The example of digital histopathology~\cite{veta2014breast}, a crucial data type in cancer research, illustrates the potential limitations of such synthetic partition methods.
In digital histopathology, tissue samples are extracted from patients, stained, and finally digitized.
In this process, known factors of data heterogeneity across hospitals include patient demographics, staining techniques,
storage methodologies of the physical slides, and digitization processes~\cite{janowczyk2019histoqc,fu2020pan,howard2021impact}.
Although staining normalization~\cite{lahiani2020seamless, de2021deep} has seen recent progress, mitigating this source of heterogeneity,
the other highlighted sources of heterogeneity are difficult to replicate with synthetic partitioning~\cite{howard2021impact} and some may be unknown, which calls for actual cross-silo cohort experiments.
This observation is also valid for many other application domains, e.g. radiology~\cite{hahn2006adrenal}, dermatology~\cite{badano2015consistency}, retinal images~\cite{badano2015consistency} and
more generally computer vision~\cite{torralba2011unbiased}.

In order to address the lack of realistic cross-silo datasets, we propose FLamby, an open source cross-silo federated dataset suite with natural partitions focused on healthcare, accompanied by code examples, and benchmarking guidelines. Our ambition is that FLamby becomes the reference benchmark for  cross-silo FL, as LEAF~\cite{caldas2018leaf} is for cross-device FL.
To the best of our knowledge, apart from some promising isolated works to build realistic cross-silo FL datasets (see Section~\ref{sec:related_work}), our work is the first standard benchmark allowing to systematically study healthcare cross-silo FL on different data modalities and tasks.

To summarize, our contributions are threefold:
\begin{enumerate}
    \item We build an open-source federated cross-silo healthcare dataset suite including $7$ datasets.  These datasets cover different tasks (classification / segmentation / survival) in multiple application domains and with different data modalities and scale. Crucially, all datasets are partitioned using natural splits.
    \item We provide guidelines to help compare FL strategies in a fair and reproducible manner, and provide illustrative results for this benchmark.
    \item We make open-source code accessible for benchmark reproducibility and easy integration in different FL frameworks, but also to allow the research community to contribute to FLamby development, by adding more datasets, benchmarking types and FL strategies.
\end{enumerate}

This paper is organized as follows. 
Section~\ref{sec:related_work} reviews existing FL datasets and benchmarks, as well as client partition methods used to artificially introduce data heterogeneity.
In Section~\ref{sec:flamby}, we describe our dataset suite in detail, notably its structure and the intrinsic heterogeneity of each federated dataset. Finally, we define a benchmark of several FL strategies on all datasets and provide results thereof in Section~\ref{sec:experiments}. %
\section{Related Work \label{sec:related_work}}
In FL, data is collected locally in clients in different conditions and without coordination.
As a consequence, clients' datasets differ both in size (unbalanced) and in distribution (non-IID)~\cite{mcmahan2017communication}.
The resulting \emph{statistical heterogeneity} is a fundamental challenge in FL~\cite{li2020suvey, kairouz2019advances}, and it is necessary to take it into consideration when evaluating FL algorithms.
Most FL papers simulate statistical heterogeneity by artificially partitioning classic datasets, e.g., CIFAR-10/100~\cite{Krizhevsky09learningmultiple}, MNIST~\cite{lecun-mnisthandwrittendigit-2010} or ImageNet~\cite{deng2009imagenet}, on a given number of clients.
Common approaches to produce synthetic partitions of classification datasets include associating samples from a limited number of classes to each client~\cite{mcmahan2017communication}, Dirichlet sampling on the class labels~\cite{hsu2019measuring, yurochkin2019bayesian}, and using Pachinko Allocation Method (PAM)~\cite{li2006pachinko, reddi2020adaptive} (which is only possible when the labels have a hierarchical structure).
In the case of regression tasks, \cite{philippenko2020bidirectional} partitions the \emph{superconduct} dataset~\cite{caruana2004kdd} across 20 clients using Gaussian Mixture clustering based on T-SNE representations~\cite{van2008visualizing} of the features.
Such synthetic partition approaches may fall short of modelling the complex statistical heterogeneity of real federated datasets.
Evaluating FL strategies on datasets with natural client splits is a safer approach to ensuring that new strategies address real-world issues.

\begin{table}[t!]
\centering
  \caption{Overview of the datasets, tasks, metrics and baseline models in FLamby. For Fed-Camelyon16 the two different sizes refer to the size of the dataset before and after tiling.} \label{tab:dataset_summary}
\resizebox{\columnwidth}{!}{
\small
\color{white}\sffamily
 \begin{tabular}{@{} >{\color{black}} m{30mm} @{}  >{\color{black}}  m{30mm} @{} >{\color{black}} m{30mm} @{}  >{\color{black}} m{30mm} @{}  >{\color{black}} m{30mm} @{}  >{\color{black}}  m{30mm} @{}  >{\color{black}} m{35mm} @{}  >{\color{black}}  m{30mm} } 
 \thickhline
 \rowcolor{lightredwilds}
 \centering
 \textbf{Dataset} & \centering \textbf{Fed-Camelyon16} & \centering \textbf{Fed-LIDC-IDRI} & \centering \textbf{Fed-IXI} & \centering \textbf{Fed-TCGA-BRCA} & \centering \textbf{Fed-KITS2019} & \centering \textbf{Fed-ISIC2019} & \centering \textbf{Fed-Heart-Disease} \tabularnewline
  \thickhline

 \rowcolor{darkgraylighter}
\centering \textbf{Input (x)} & \centering Slides & \centering CT-scans & \centering T1WI & \centering Patient info. & \centering CT-scans & \centering Dermoscopy & \centering Patient info. \tabularnewline
  \thickhline
  \rowcolor{palegray}
\centering \textbf{Preprocessing} & \centering Matter extraction \newline + tiling  & \centering Patch Sampling & \centering Registration & \centering None & \centering Patch Sampling & \centering Various image \newline transforms & \centering Removing missing data \tabularnewline

   \thickhline
 \rowcolor{darkgraylighter}
\centering \textbf{Task type} & \centering binary \newline classification & \centering 3D segmentation & \centering 3D segmentation & \centering survival & \centering 3D segmentation & \centering multi-class \newline classification & \centering binary classification \tabularnewline
    \thickhline
 \rowcolor{palegray}
\centering \textbf{Prediction (y)} & \centering Tumor on slide & \centering Lung Nodule Mask & \centering Brain mask & \centering Risk of death & \centering Kidney and tumor masks & \centering Melanoma class & \centering  Heart disease \tabularnewline
  \thickhline
  \rowcolor{darkgraylighter}
 \textbf{Center extraction} & \centering Hospital & \centering Scanner Manufacturer & \centering Hospital & \centering Group of Hospitals & \centering Group of Hospitals & \centering Hospital & \centering Hospital \tabularnewline
   \thickhline
  \rowcolor{lightbluewilds} 
\centering \textbf{Thumbnails} 
& \vspace{2mm} \centering \adjincludegraphics[width=25mm, height=25mm, valign=b]{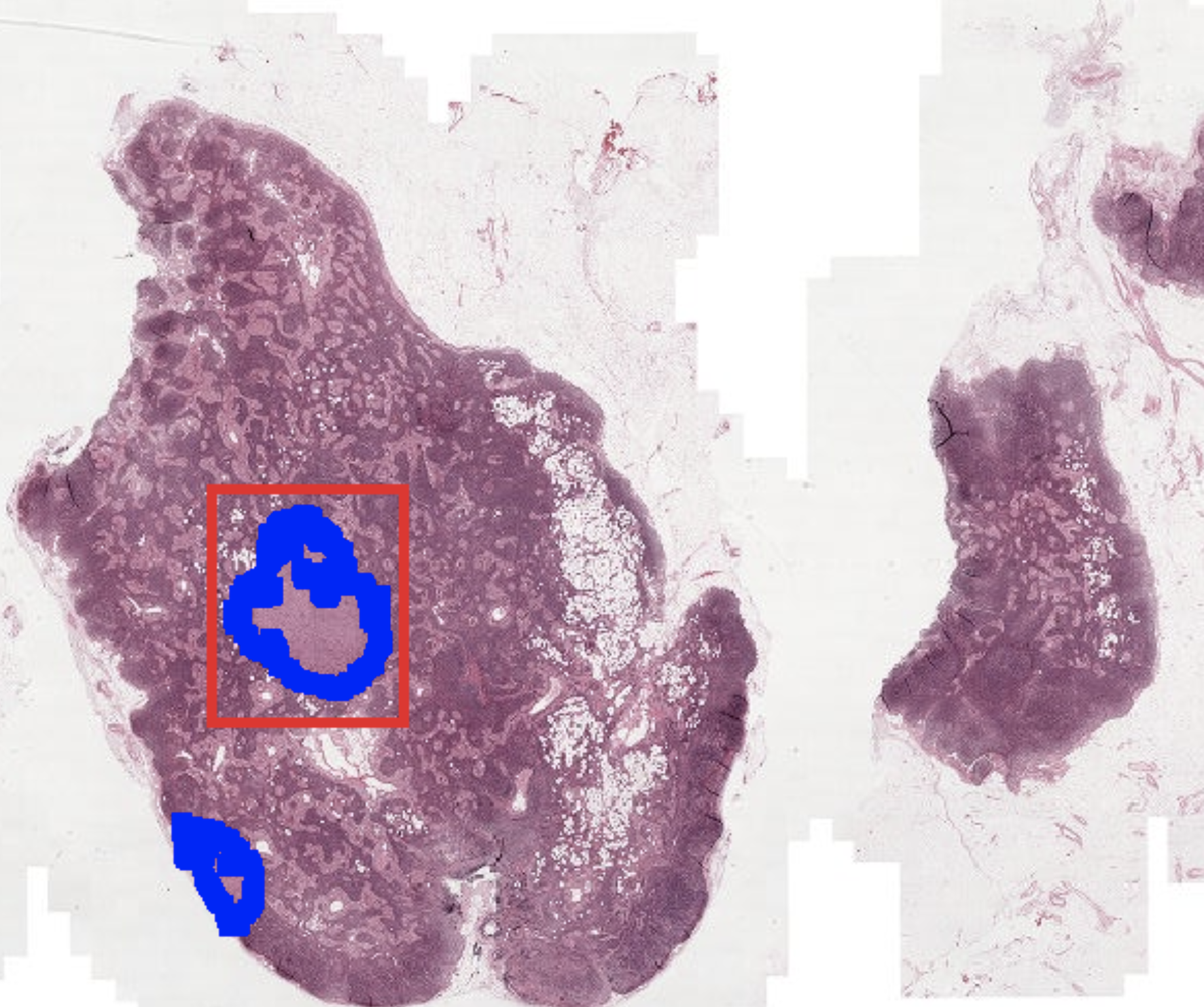} \vspace{2mm}
& \vspace{2mm} \centering \includegraphics[width=25mm, height=25mm, valign=b]{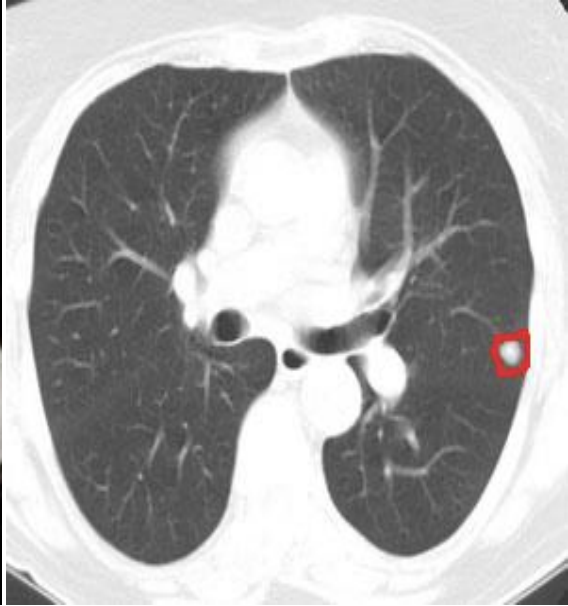} \vspace{2mm} 
& \vspace{2mm} \centering \includegraphics[width=25mm, height=25mm, valign=b]{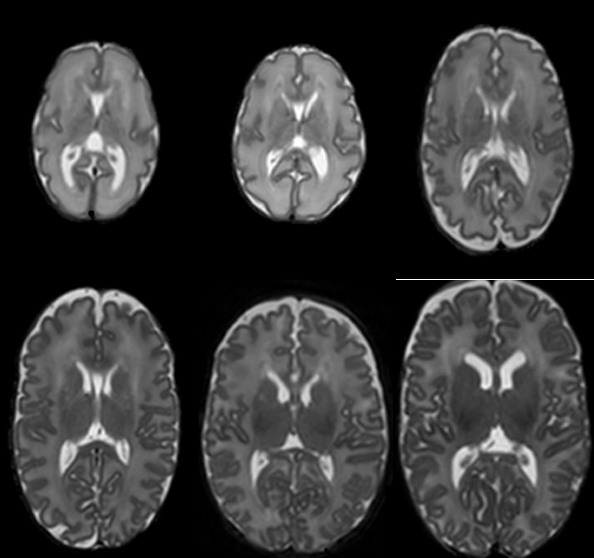} \vspace{2mm}
& \vspace{2mm} \centering \includegraphics[width=25mm, height=25mm, valign=b]{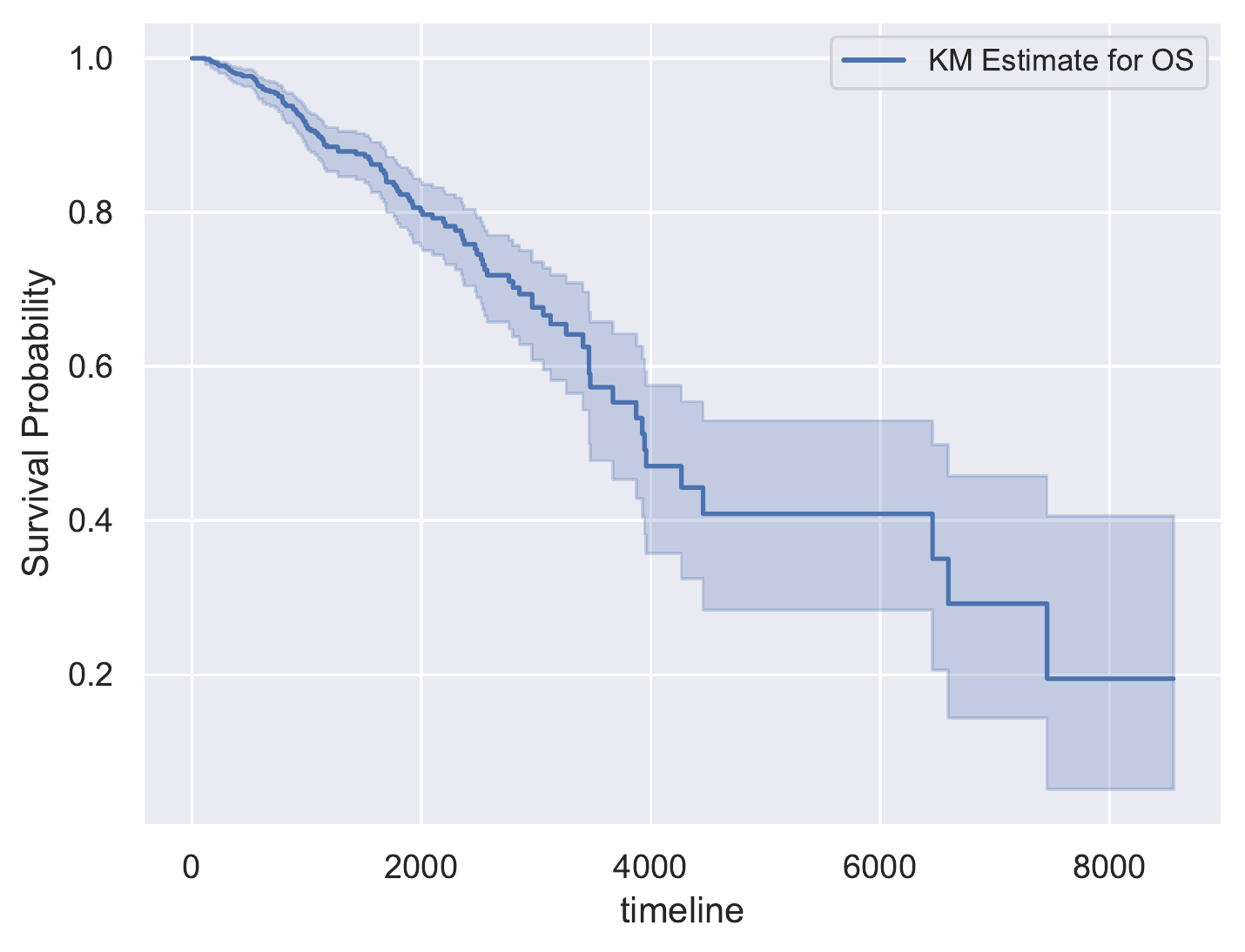} \vspace{2mm}
& \vspace{2mm} \centering \includegraphics[width=25mm, height=25mm, valign=b]{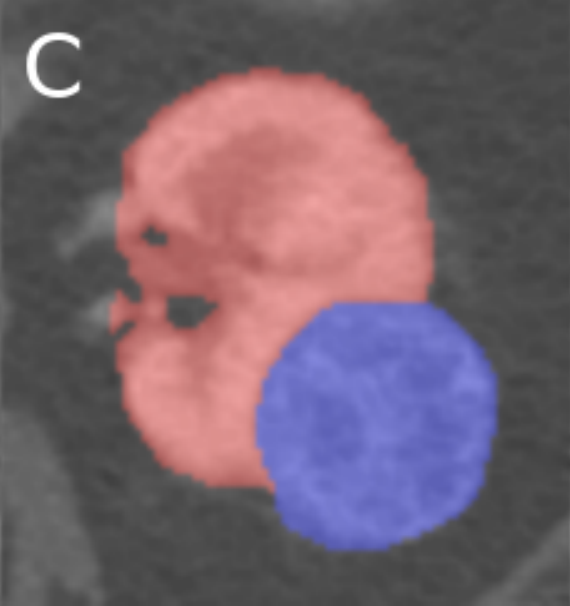} \vspace{2mm} 
& \vspace{2mm} \centering \includegraphics[width=25mm, height=25mm, valign=b]{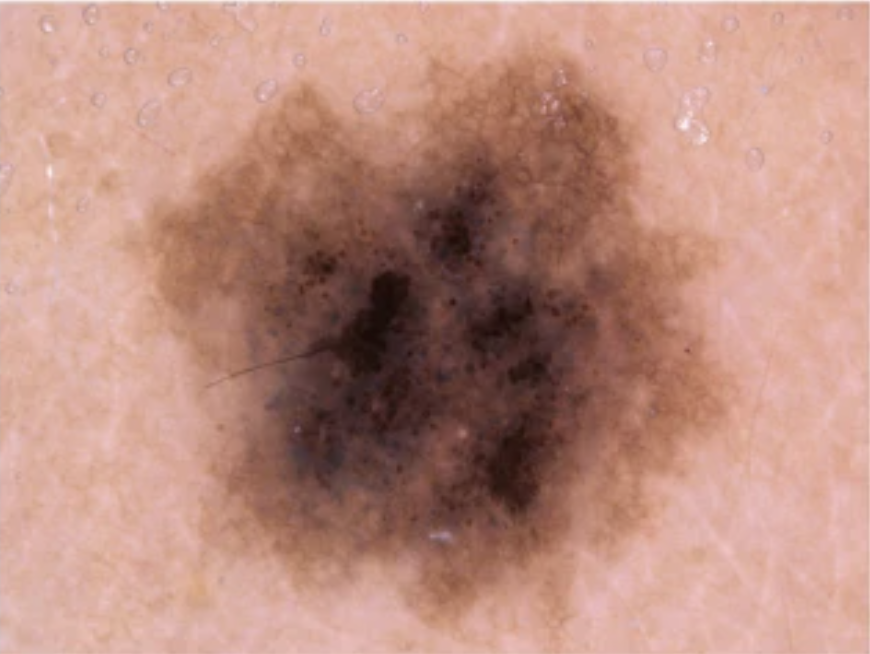} \vspace{2mm}
& \vspace{2mm} \centering \includegraphics[width=25mm, height=25mm, valign=b]{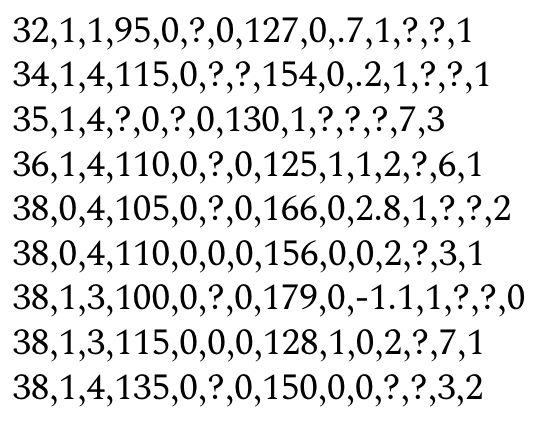} \vspace{2mm} \tabularnewline
  \thickhline
 \rowcolor{palegray}
\centering \textbf{Original paper} 
& \centering Litjens \textit{et al.} \newline 2018  
& \centering Armato \textit{et al.} \newline 2011 
& \centering Perez \textit{et al.} \newline 2021
& \centering \hspace{2mm} Liu \textit{et al.} \newline 2018
& \centering Heller \textit{et al.} \newline 2019
& \centering Tschandl \textit{et al.} 2018 / Codella \textit{et al.} 2017 / Combalia \textit{et al.} 2019
& \centering Janosi \textit{et al.} \newline 1988 \tabularnewline
  \thickhline
 \rowcolor{darkgraylighter}
\centering \textbf{\# clients} & \centering 2 & \centering 4 & \centering 3 & \centering 6 & \centering 6 & \centering 6 & \centering 4  \tabularnewline
  \thickhline
 \rowcolor{palegray}

\centering \textbf{\# examples} & \centering 399 & \centering 1,018 & \centering 566 & \centering 1, 088 & \centering 96 & \centering 23, \centering 247 & \centering 740  \tabularnewline
  \thickhline
 \rowcolor{darkgraylighter}
\centering \textbf{\# examples per center} & \centering 239, 150  & \centering 670, 205, 69, 74 & \centering 311, 181, 74 &  \centering 311, 196, 206, 162, 162, 51 & \centering 12, 14, 12, 12, \quad \quad 16, 30 & \centering 12413, 3954, 3363, 2259, 819, 439  & \centering 303, 261, 46, 130 \tabularnewline
  \thickhline
 \rowcolor{palegray}
\centering \textbf{Model} & \centering DeepMIL~\cite{deepmil} & \centering Vnet~\cite{milletari2016v, adaloglou2019MRIsegmentation} & \centering 3D U-net~\cite{cciccek20163d} & \centering Cox Model~\cite{cox1972regression} & \centering nnU-Net \cite{isensee2021nnu} & \centering efficientnet~\cite{DBLP:journals/corr/abs-1905-11946} \newline + linear layer & \centering Logistic Regression \tabularnewline
 \thickhline

 \rowcolor{darkgraylighter}
\centering \textbf{Metric} & \centering AUC & \centering DICE & \centering DICE & \centering C-index & \centering DICE & \centering Balanced Accuracy & \centering Accuracy \tabularnewline
 \thickhline
 \rowcolor{palegray}
\centering \textbf{Size} & \centering 50G (850G total) & \centering 115G & \centering 444M & \centering 115K & \centering 54G & \centering 9G & \centering 40K \tabularnewline
 \thickhline
 \rowcolor{darkgraylighter}

\centering \textbf{Image resolution} & \centering  0.5~{\textmu}m / pixel & { \centering  $\sim$1.0 × 1.0 × 1.0 \par \hspace{6mm} mm / voxel} & {\centering $\sim$ 1.0 × 1.0 × 1.0\par \hspace{6mm} mm / voxel} & \centering NA & {\centering $\sim$1.0 × 1.0 × 1.0 \par \hspace{6mm} mm / voxel} & \centering $\sim$0.02 mm / pixel & \centering NA \tabularnewline
 \thickhline 
 \rowcolor{palegray}
\centering \textbf{Input dimension} & \centering 10, 000 x 2048  & \centering 128 x 128 x 128 & \centering 48 x 60 x 48 &  \centering 39 & \centering 64 x 192 x 192 & \centering 200 x 200 x 3 & \centering 13 \tabularnewline
\hline 
\end{tabular}}
\end{table}

For \emph{cross-device} FL, the LEAF dataset suite~\cite{caldas2018leaf} includes five datasets with natural partition, spanning a wide range of machine learning tasks: natural language modeling (Reddit~\cite{volskeetal2017tl}), next character prediction (Shakespeare~\cite{mcmahan2017communication}), sentiment analysis (Sent140~\cite{go2009twitter}), image classification (CelebA~\cite{liu2015deep}) and handwritten-character recognition (FEMNIST~\cite{cohen2017emnist}).
TensorFlow Federated \cite{bonawitz2019towards} complements LEAF and provides three additional naturally split federated benchmarks, i.e., StackOverflow~\cite{tenso2019stack}, Google Landmark v2~\cite{hsu2020federated} and iNaturalist~\cite{horn2018inaturalist}.
Further, FLSim~\cite{flsim} provides cross-device examples based on LEAF and CIFAR10~\cite{Krizhevsky09learningmultiple} with a synthetic split, and
FedScale~\cite{lai2022fedscale} introduces a large FL benchmark focused on mobile applications.
Apart from iNaturalist, the aforementioned datasets target the cross-device setting.

To the best of our knowledge, no extensive benchmark with natural splits is available for \emph{cross-silo} FL.
However, some standalone works built cross-silo datasets with real partitions.
\cite{gong2021ensemble} and \cite{marfoq20neurips} partition Cityscapes~\cite{Cordts2016Cityscapes} and iNaturalist~\cite{horn2018inaturalist}, respectively, exploiting the geolocation of the picture acquisition site.
\cite{icml2020_3152} releases a real-world, geo-tagged dataset of common mammals on Flickr.
\cite{luo2019real} gathers a  federated cross-silo benchmark for object detection created using street cameras.
\cite{corinzia2019variational} partitions Vehicle Sensor Dataset~\cite{duarte2004vehicle} and Human Activity Recognition dataset~\cite{anguita2013public} by sensor and by individuals, respectively. \cite{luofediris} builds an iris recognition federated dataset across five clients using multiple iris datasets~\cite{wei2007nonlinear, zhang2010contact, zhang2018deep,phillips2008iris}.
While FedML~\cite{he2020fedml} introduces several cross-silo benchmarks~\cite{he2021fedcv,yuchen2022fednlp,he2021fedgraphnn}, the related client splits are synthetically obtained with Dirichlet sampling and not based on a natural split.
Similarly, FATE~\cite{fate} provides several cross-silo examples but, to the best of our knowledge, none of them stems from a natural split.

In the medical domain, several works use natural splits replicating the data collection process in different hospitals: the works \cite{andreux2020siloed, chakravarty2021federated, baheti2020federated, kaissis2021end, xie2022fedmed, chang2018distributed} respectively use the Camelyon datasets~\cite{litjens20181399,bejnordi2017diagnostic,bandi2018detection}, the CheXpert dataset~\cite{irvin2019chexpert}, LIDC dataset~\cite{armato2011lidc}, the chest X-ray dataset~\cite{kermany2018identifying}, the IXI dataset~\cite{xie2022fedmed}, the Kaggle diabetic retinopathy detection dataset~\cite{graham2015kaggle}.
Finally, the works~ \cite{andreux2020federated, gunesli2021feddropoutavg, lu2022federated} use the TCGA dataset~\cite{tomczak2015cancer} by extracting the Tissue Source site metadata. %

Our work aims to give more visibility to such isolated cross-silo initiatives by regrouping seven medical datasets, some of which listed above, in a single benchmark suite. We also provide reproducible code alongside precise benchmarking guidelines in order to connect past and subsequent works for a better monitoring of the progress in cross-silo FL.

\section{The FLamby Dataset Suite \label{sec:flamby}}
\subsection{Structure Overview}
The FLamby datasets suite is a Python library organized in two main parts: datasets with corresponding baseline models, and FL strategies with associated benchmarking code.
The suite is modular, with a standardized simple application programming interface (API) for each component, enabling easy re-use and extensions of different components.
Further, the suite is compatible with existing FL software libraries, such as FedML~\cite{he2020fedml}, Fed-BioMed~\cite{silva2020fed}, or Substra~\cite{galtier2019substra}.
Listing~\ref{list:code_example} provides a code example of how the structure of FLamby allows to test new datasets and strategies in a few lines of code,
and Table~\ref{tab:dataset_summary} provides an overview of the FLamby datasets.

\paragraph{Dataset and baseline model.}
The FLamby suite contains datasets with a natural notion of client split, as well as a predefined task and associated metric.
A train/test set is predefined for each client to enable reproducible comparisons.
We further provide a baseline model for each task, with a reference implementation for training on pooled data.
For each dataset, the suite provides documentation, metadata and helper functions to: 1. download the original pooled dataset; 2. apply preprocessing if required, making it suitable for ML training; 3. split each original pooled dataset between its natural clients; and 4. easily iterate over the preprocessed dataset.
The dataset API relies on PyTorch~\cite{paszke2019pytorch}, which makes it easy to iterate over the dataset with natural splits as well as to modify these splits if needed.

\paragraph{FL strategies and benchmark.}
FL training algorithms, called \textit{strategies} in the FLamby suite, are provided for simulation purposes.
In order to be agnostic to existing FL libraries, these strategies are provided in plain Python code.
The API of these strategies is standardized and compatible with the dataset API, making it easy to benchmark each strategy on each dataset.
We further provide a script performing such a benchmark for illustration purposes.
We stress the fact that it is easy to alternatively use implementations from existing FL libraries.

\subsection{Datasets, Metrics and Baseline Models}
\label{subsec:datasets}
\begin{figure}[t!]
\centering
\begin{subfigure}[b]{0.45\textwidth}
\includegraphics[max width=\textwidth]{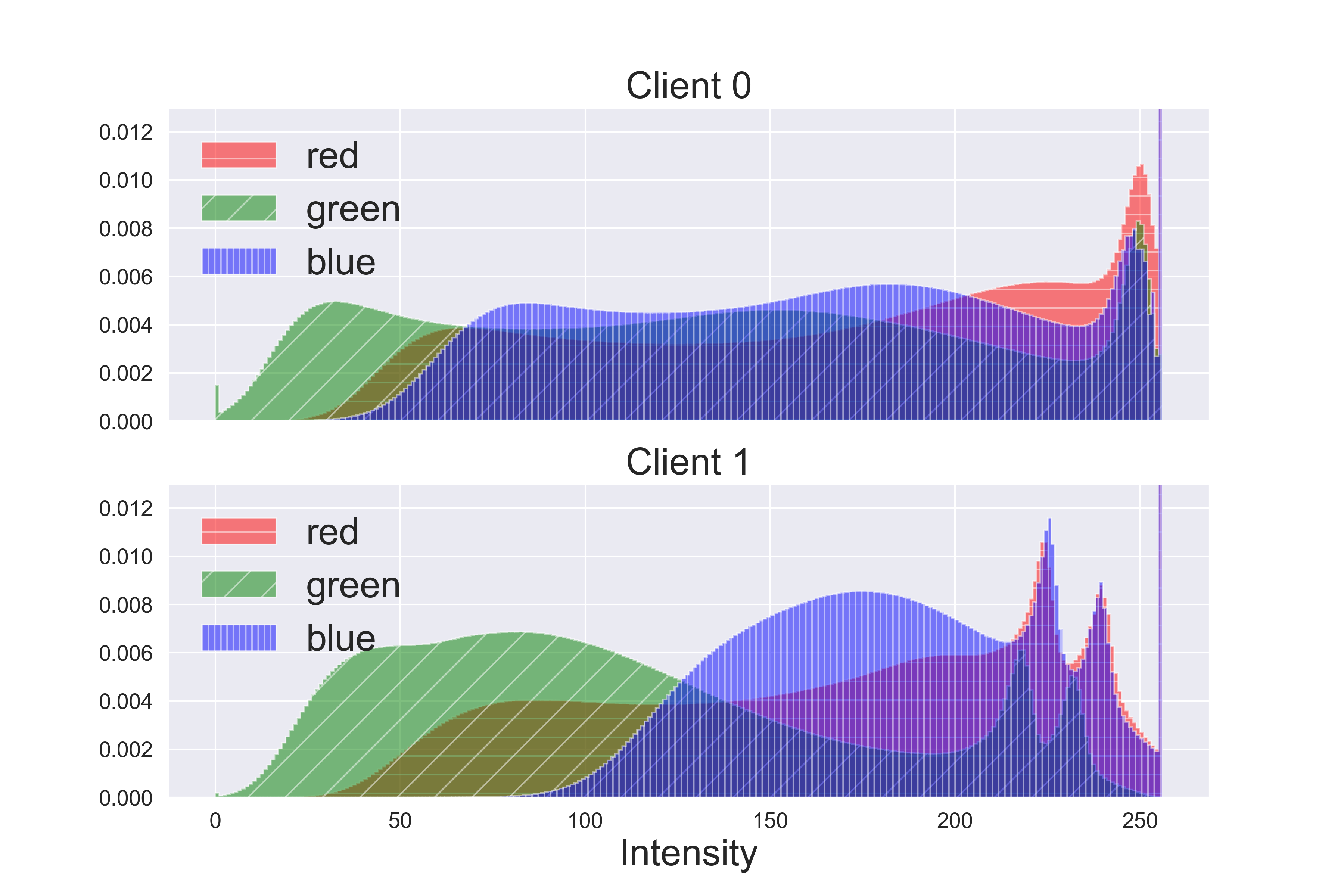}
\caption{Fed-Camelyon16}
\label{subfig:heterogeneity_camelyon16}
\end{subfigure}
\begin{subfigure}[b]{0.34\textwidth}
\includegraphics[max width=\textwidth]{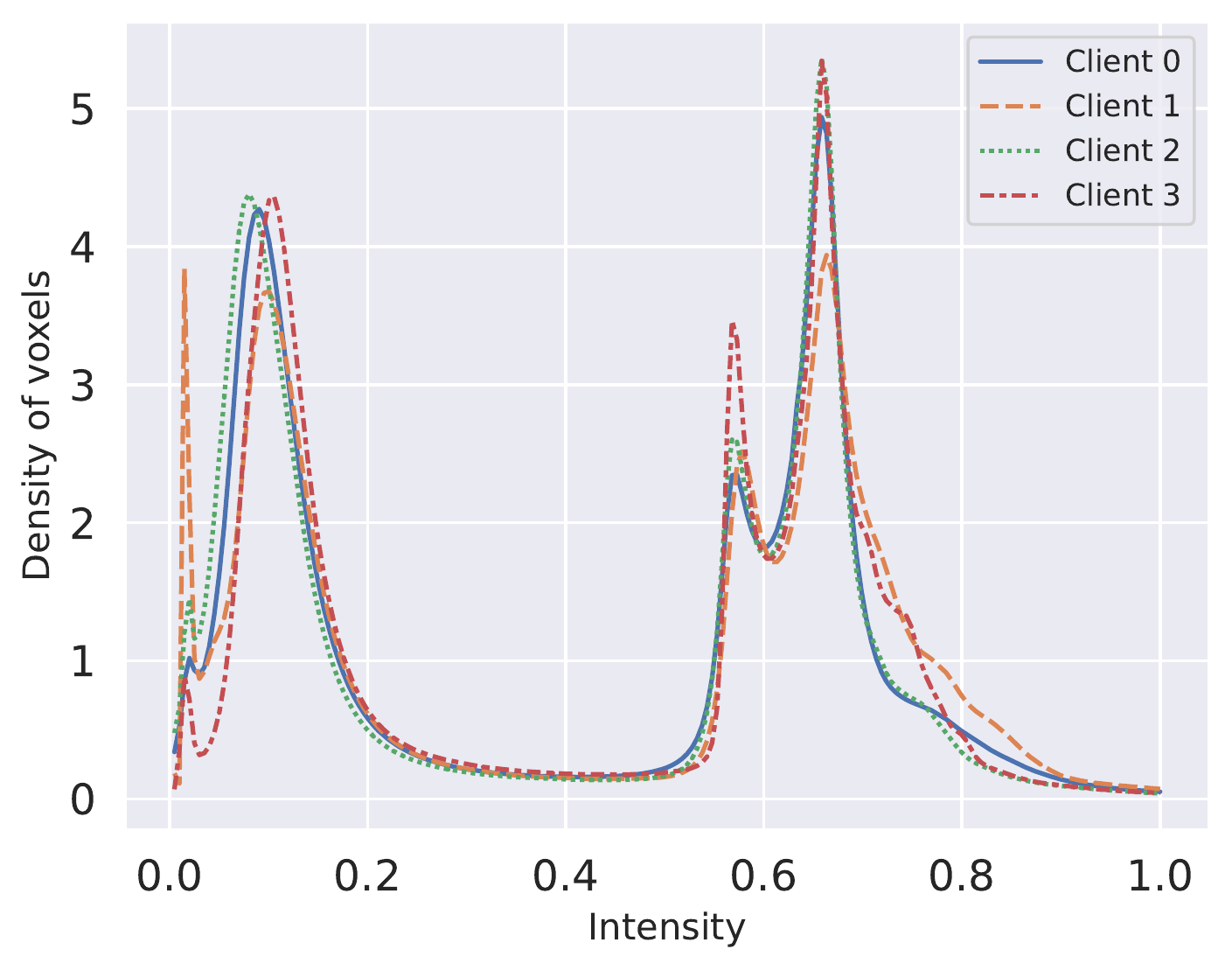}
\caption{Fed-LIDC-IDRI}
\label{subfig:heterogeneity_lidc}
\end{subfigure}
\begin{subfigure}[b]{0.45\textwidth}
\includegraphics[max width=\textwidth]{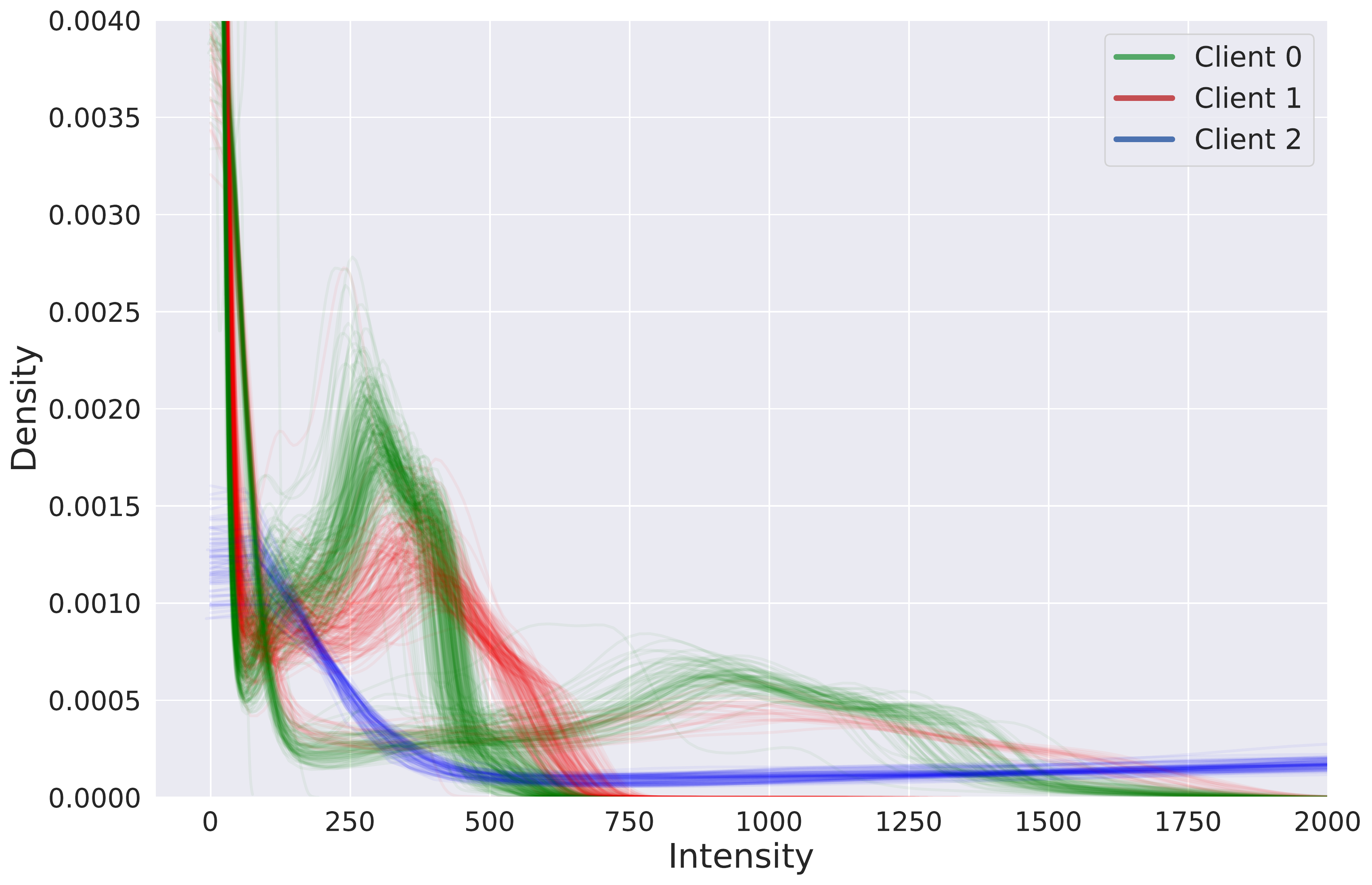} 
\caption{Fed-IXI}
\label{subfig:heterogeneity_ixi}
\end{subfigure}
\begin{subfigure}[b]{0.45\textwidth}
\includegraphics[max width=\textwidth]{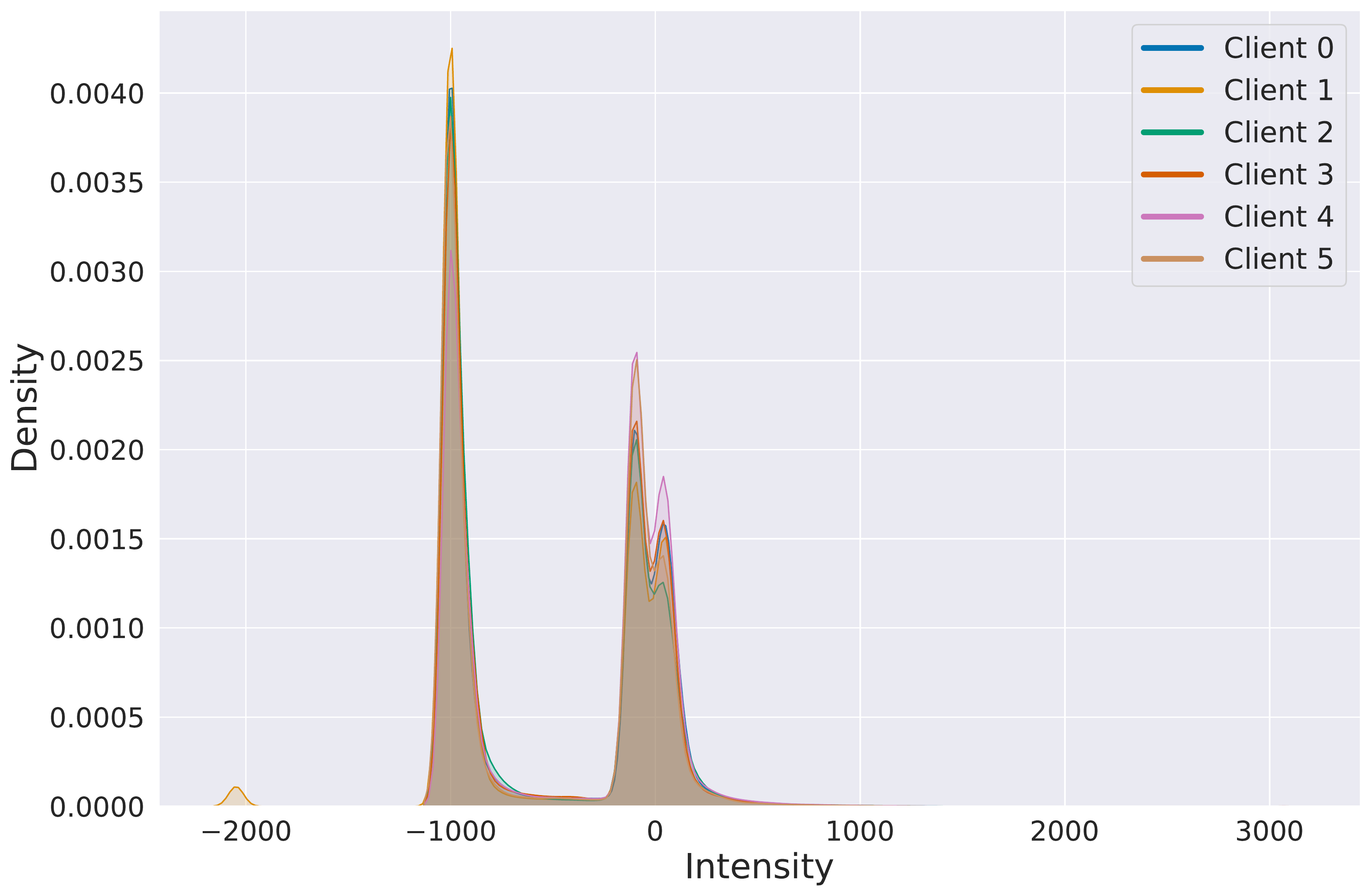}
\caption{Fed-KITS2019}
\label{subfig:heterogeneity_kits}
\end{subfigure}
\begin{subfigure}[b]{0.45\textwidth}
\includegraphics[max width=\textwidth]{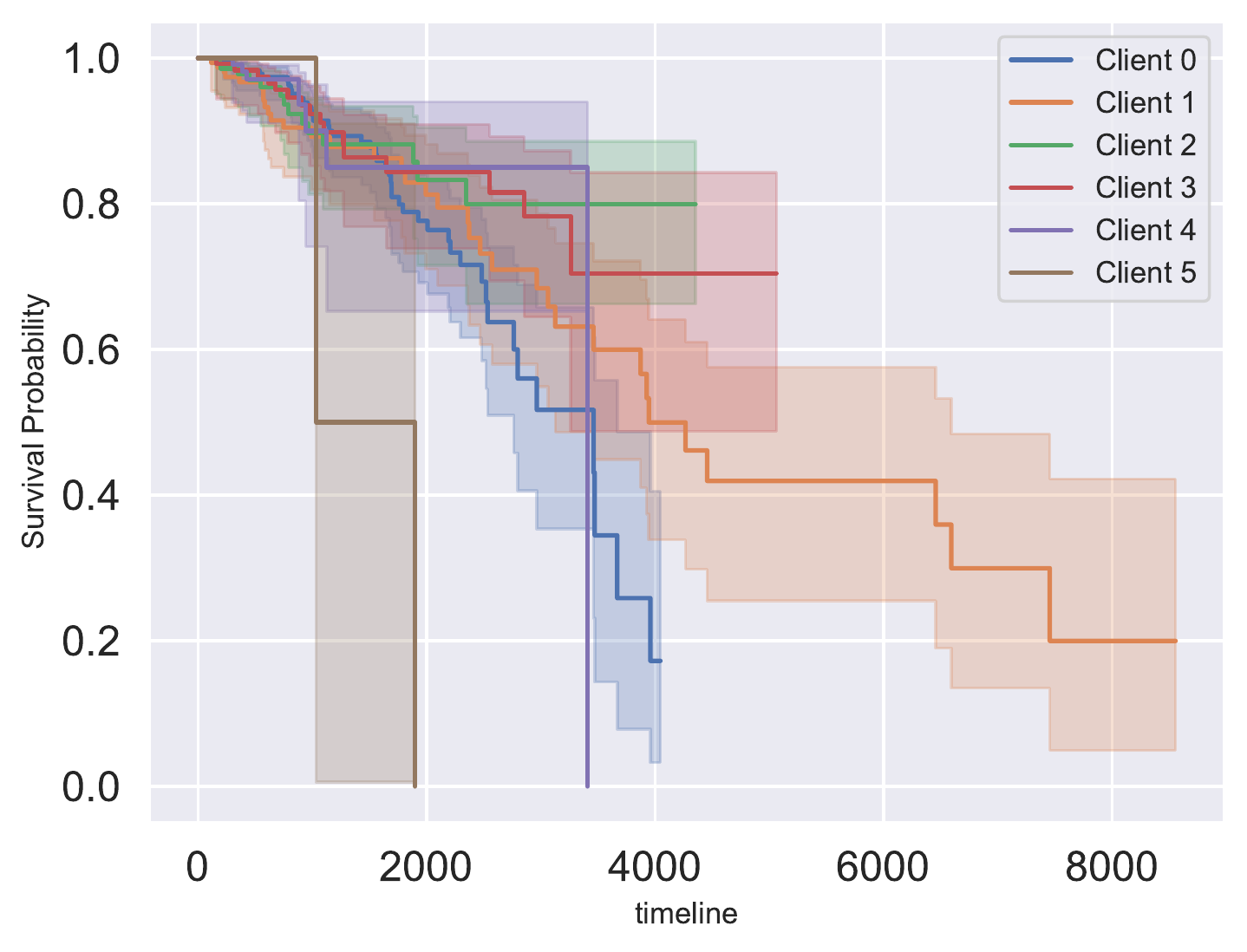} 
\caption{Fed-TCGA-BRCA}
\label{subfig:heterogeneity_tcga}
\end{subfigure}
\begin{subfigure}[b]{0.38\textwidth}
\includegraphics[max width=\textwidth]{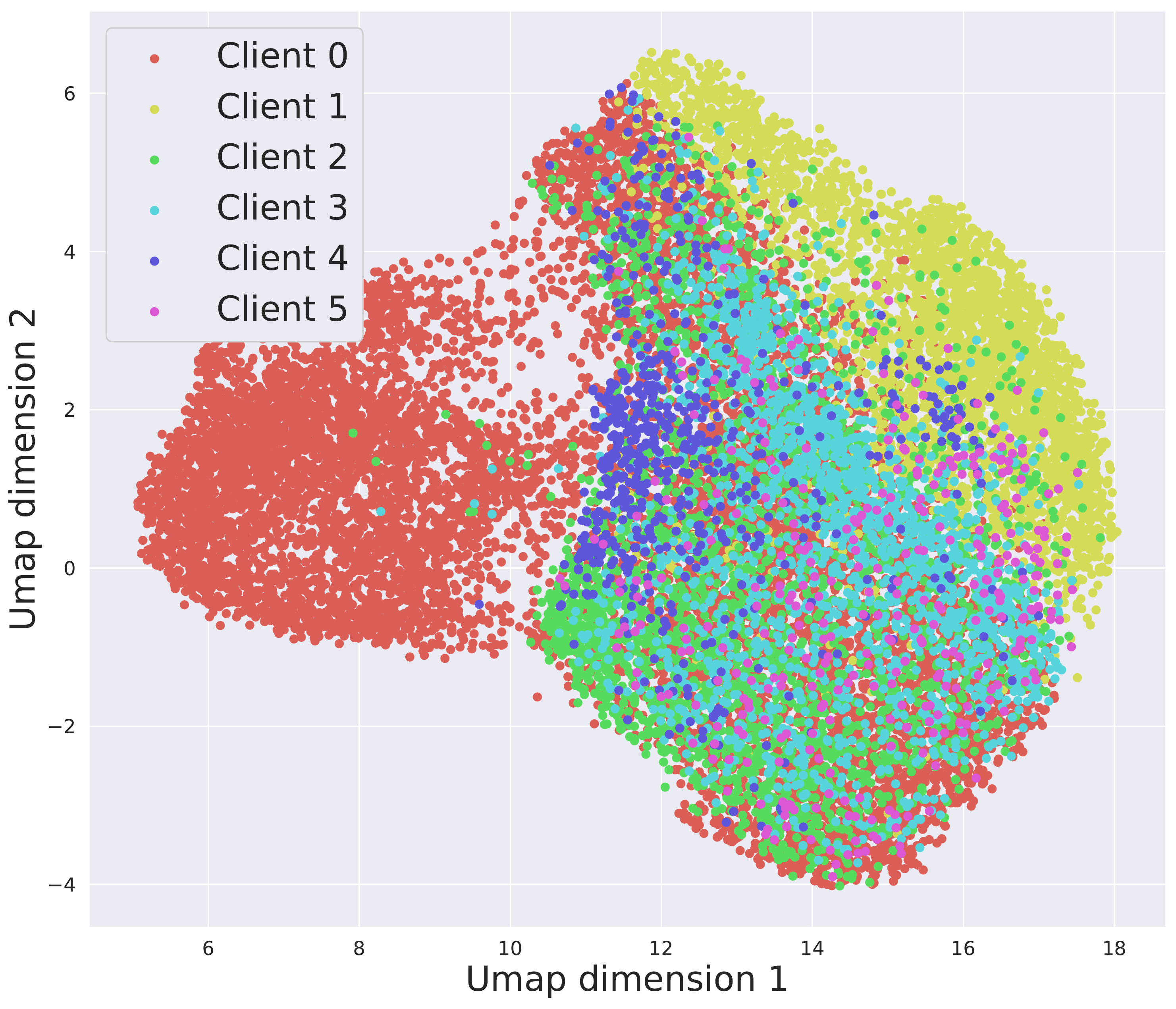}
\caption{Fed-ISIC2019}
\label{subfig:heterogeneity_isic}
\end{subfigure}
\begin{subfigure}[b]{0.45\textwidth}
\includegraphics[max width=\textwidth]{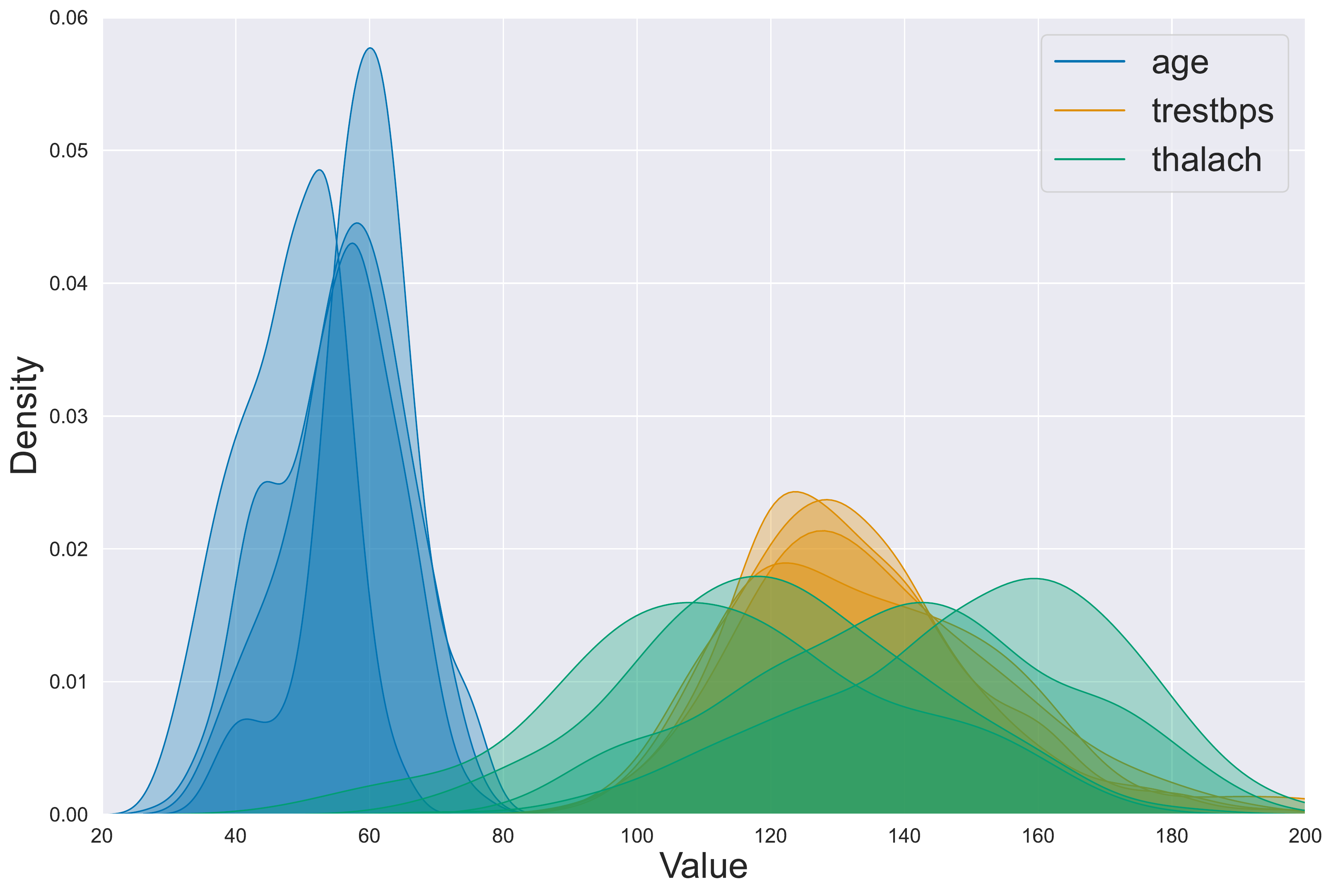}
\caption{Fed-Heart-Disease}
\label{subfig:heterogeneity_heart}
\end{subfigure}
\caption{Heterogeneity of FLamby datasets. Best seen in color.
\ref{subfig:heterogeneity_camelyon16}: Color histograms per client.
\ref{subfig:heterogeneity_lidc},
\ref{subfig:heterogeneity_ixi} and \ref{subfig:heterogeneity_kits}: Voxel intensity distribution per client.
\ref{subfig:heterogeneity_tcga}: Kaplan-Meier survival curves per client.
\ref{subfig:heterogeneity_isic}: UMAP of deep network features of the raw images, colored by client.
\ref{subfig:heterogeneity_heart}: Per-client histograms of several features.
Differences between client distributions are sometimes obvious and sometimes subtle. Some clients are close in the feature space, some are not and different types of heterogeneity are observed with different data modalities.}
    \label{fig:hetero}
    \vskip-.5cm
\end{figure}

We provide a brief description of each dataset in the FLamby dataset suite,
which is summarized in Table~\ref{tab:dataset_summary}.
In Section~\ref{subsec:heterogeneity}, we further explore the heterogeneity of each dataset,
as displayed in Figure~\ref{fig:hetero}.
\paragraph{Fed-Camelyon16.} Camelyon16~\cite{litjens20181399} is a histopathology dataset of 399 digitized breast biopsies' slides with or without tumor collected from two hospitals: Radboud University Medical Center (RUMC) and University Medical Center Utrecht (UMCU).
By recovering the original split information we build a federated version of Camelyon16 with \textbf{2} clients.
The task consists in binary classification of each slide, which is challenging due to the large size of each image ($10^5 \times 10^5$ pixels at 20X magnification), and measured by the Area Under the ROC curve (AUC).

As a baseline, we follow a weakly-supervised learning approach.
Slides are first converted to bags of local features, which are one order of magnitude smaller in terms of memory requirements, and a model is then trained on top of this representation.
For each slide, we detect regions with a matter-detection network and then extract features from each tile with an ImageNet-pretrained Resnet50, following state-of-the-art practice~\cite{courtiol2018classification, lu2021data}.
Note that due to the imbalanced distribution of tissue in the different slides, a different number of features is produced for each slide: we cap the total number of tiles to $10^5$ and use zero-padding for consistency.
We then train a DeepMIL architecture~\cite{ilse2018attention}, using its reference implementation~\cite{deepmil} and hyperparameters from~\cite{dehaene2020self}.
We refer to Appendix~\ref{app:camelyon} for more details.
\paragraph{Fed-LIDC-IDRI.} LIDC-IDRI~\cite{armato2011lidc, lidcdata, clark2013cancer} is an image database~\cite{clark2013cancer} study with 1018 CT-scans (3D images) from The Cancer Imaging Archive (TCIA),
proposed in the LUNA16 competition~\cite{setio2017validation}.
The task consists in automatically segmenting lung nodules in CT-scans, as measured by the DICE score~\cite{dice1945measures}.
It is challenging because lung nodules are small, blurry, and hard to detect.
By parsing the metadata of the CT-scans from the provided annotations, we recover the manufacturer of each scanning machine used, which we use as a proxy for a client.
We therefore build a \textbf{4}-client federated version of this dataset, split by manufacturer.
Figure~\ref{subfig:heterogeneity_lidc} displays the distribution of voxel intensities in each client.

As a baseline model, we use a VNet~\cite{milletari2016v} following the implementation from~\cite{adaloglou2019MRIsegmentation}.
This model is trained by sampling 3D-volumes into 3D patches fitting in GPU memory.
Details of the sampling procedure are available in Appendix~\ref{app:lidc}.
\paragraph{Fed-IXI.} This dataset is extracted from the Information eXtraction from Images - IXI database~\cite{ixi}, and has been previously released by Perez \textit{et} al.~\cite{ixitiny,perez2021torchio} under the name of \emph{IXITiny}.
IXITiny provides a database of brain T1 magnetic resonance images (MRIs) from \textbf{3} hospitals (Guys, HH, and IOP).
This dataset has been adapted to a brain segmentation task by obtaining
spatial brain masks using a state-of-the-art unsupervised brain segmentation tool~\cite{iglesias2011robust}.
The quality of the resulting supervised segmentation task is measured by the DICE score~\cite{dice1945measures}.

The image pre-processing pipeline includes volume resizing to $48 \times 60 \times 48$ voxels, and sample-wise intensity normalization.
Figure~\ref{subfig:heterogeneity_ixi} highlights the heterogeneity of the raw MRI intensity distributions between clients.
As a baseline, we use a 3D U-net~\cite{cciccek20163d} following the implementation of~\cite{iximodel}.
Appendix~\ref{app:ixi} provides more detailed information about this dataset, including demographic information,
and about the baseline.
\paragraph{Fed-TCGA-BRCA.} The Cancer Genome Atlas (TCGA)'s Genomics Data Commons (GDC) portal~\cite{tcga} contains multi-modal data (tabular, 2D and 3D images) on a variety of cancers collected in many different hospitals.
Here, we focus on clinical data from the BReast CAncer study (BRCA), which includes features gathered from 1066 patients.
We use the Tissue Source Site metadata to split data based on extraction site, grouped into geographic regions to obtain large enough clients.
We end up with \textbf{6} clients: USA~(Northeast, South, Middlewest, West), Canada and Europe,
with patient counts varying from~51 to~311.
The task consists in predicting survival outcomes~\cite{jenkins2005survival} based on the patients' tabular data~(39 features overall), with the event to predict being death.
This survival task is akin to a ranking problem with the score of each sample being known either directly or only by lower bound (right censorship).
The ranking is evaluated by using the concordance index (C-index) that measures the percentage of correctly ranked pairs while taking censorship into account.

As a baseline, we use a linear Cox proportional hazard model~\cite{cox1972regression}
to predict time-to-death for patients.
Figure~\ref{subfig:heterogeneity_tcga} highlights the survival distribution heterogeneity between the different clients.
Appendix~\ref{app:tcga} provides more details on this dataset.
\paragraph{Fed-KITS2019.} 
The KiTS19 dataset ~\cite{heller2020state, heller2019kits19} stems from the Kidney Tumor Segmentation Challenge 2019 and contains CT scans of 210 patients along with the segmentation masks from 79 hospitals.
We recover the hospital metadata and extract a \textbf{6}-client federated version of this dataset by removing hospitals with less than $10$ training samples.
The task consists of both kidney and tumor segmentation, labeled 1 and 2, respectively, and we measure the average of Kidney and Tumor DICE scores~\cite{dice1945measures} as our evaluation metric.

The preprocessing pipeline comprises intensity clipping followed by intensity normalization, and resampling of all the cases to a common voxel spacing of 2.90x1.45x1.45 mm.
As a baseline, we use the nn-Unet library~\cite{isensee2021nnu} to train a 3D nnU-Net, combined with multiple data augmentations including scaling, rotations, brightness, contrast, gamma and Gaussian noise with the batch generators framework~\cite{isensee2020batchgenerators}. Appendix~\ref{app:kits} provides more details on this dataset.
\paragraph{Fed-ISIC2019.} The ISIC2019 dataset~\cite{tschandl2018ham10000,codella2018skin,
combalia2019bcn20000} contains dermoscopy images collected in~4 hospitals.
We restrict ourselves to~23,247 images from the public train set due to metadata availability reasons, which we re-split into train and test sets.
The task consists in image classification among~8 different melanoma classes, with high label imbalance (prevalence ranging 49\% to less than 1\% depending on the class).
We split this dataset based on the imaging acquisition system used:
as one hospital used 3 different imaging technologies throughout time, we end up with a \textbf{6}-client federated version of ISIC2019.
We measure classification performance through balanced accuracy, defined as the average recall on each class.

As an offline preprocessing step, we follow recommendations and code from~\cite{aman} by resizing images to the same shorter side while maintaining their aspect ratio, and by normalizing images' brightness and contrast through a color consistency algorithm.
As a baseline classification model, we fine-tune an EfficientNet~\cite{DBLP:journals/corr/abs-1905-11946} pretrained on ImageNet, with a weighted focal loss~\cite{DBLP:journals/corr/abs-1708-02002} and with multiple data augmentations.
Figure~\ref{subfig:heterogeneity_isic} highlights the heterogeneity between the different clients prior to preprocessing.
Appendix~\ref{app:isic} provides more details on this dataset.
\paragraph{Fed-Heart-Disease.} 
The Heart-Disease dataset~\citep{janosi1988heart} was collected in \textbf{4} hospitals in the USA, Switzerland and Hungary.
This dataset contains tabular information about 740 patients distributed among these four clients.
The task consists in binary classification to assess the presence or absence of heart disease.
We preprocess the dataset by removing missing values and encoding non-binary categorical variables as dummy variables, which gives 13 relevant attributes.
As a baseline model, we use logistic regression.
Appendix~\ref{app:heart} provides more details on this dataset.
\subsection{Federated Learning Strategies in FLamby}
\label{subsec:strategies}
The following standard FL algorithms, called \textit{strategies}, are implemented in FLamby.
We rely on a common API for all strategies, which allows for efficient benchmarking of both datasets and strategies, as shown in Listing~\ref{list:code_example}.
As we focus on the cross-silo setting, we restrict ourselves to strategies with full client participation.

\textbf{FedAvg~\cite{mcmahan2017communication}.} FedAvg is the simplest FL strategy.
It performs iterative round-based training, each round consisting in local mini-batch updates on each client followed by parameter averaging on a central server.
As a convention, we choose to count the number of local updates in batches and not in local epochs in order to match theoretical formulations of this algorithm; this choice also applies to strategies derived from FedAvg.
This strategy is known to be sensitive to heterogeneity when the number of local updates grows~\cite{li2020federated, karimireddy2020scaffold}.

\textbf{FedProx~\cite{li2020federated}.} In order to mitigate statistical heterogeneity, FedProx builds on FedAvg by introducing a regularization term to each local training loss,
thereby controlling the deviation of the local models from the last global model.

\textbf{Scaffold~\cite{karimireddy2020scaffold}.} Scaffold mitigates client drifts using control-variates and by adding a server-side learning rate. We implement a full-participation version of Scaffold that is optimized to reduce the number of bits communicated between the clients and the server.

\textbf{Cyclic Learning~\cite{chang2018distributed, sheller2018multi}.} Cyclic Learning performs local optimizations on each client in a sequential fashion, transferring the trained model to the next client when training finishes. Cyclic is a simple sequential baseline to other federated strategies.
For Cyclic, we define a round as a full cycle throughout all clients.
We implement both such cycles in a fixed order or in a shuffled order at each round.

\textbf{FedAdam~\cite{reddi2020adaptive}}, \textbf{FedYogi~\cite{reddi2020adaptive}}, \textbf{FedAdagrad~\cite{reddi2020adaptive}.} FedAdam, FedYogi and FedAdagrad are generalizations of their respective single-centric optimizers (Adam~\cite{kingma2014adam}, Yogi~\cite{zaheer2018adaptive} and Adagrad~\cite{lydia2019adagrad}) to the FL setting.
In all cases, the running means and variances of the updates are tracked at the server level.
\subsection{Dataset Heterogeneity}
\label{subsec:heterogeneity}
We qualitatively illustrate the heterogeneity of the datasets of FLamby.
For each dataset, we compute a relevant statistical distribution for each client, which differs due to the differences in tasks and modalities of the datasets.
We comment the results displayed in Figure~\ref{fig:hetero} in the following.
Appendix~\ref{app:heterogeneity} provides a more quantitative exploration of this heterogeneity.

For the \textbf{Fed-Camelyon16} dataset, we display the color histograms (RGB values) of the raw tissue patches in each client.
We see that the RGB distributions of both clients strongly differ.
For both \textbf{Fed-LIDC-IDRI} and \textbf{Fed-KITS2019} datasets, we display histograms of voxel intensities.
In both cases, we do not note significant differences between clients.
For the \textbf{Fed-IXI} dataset, we display the histograms of raw T1-MRI images, showing visible differences between clients.
For \textbf{Fed-TCGA-BRCA}, we display Kaplan-Meier estimations of the survival curves~\cite{kaplan1958nonparametric} in each client.
As detailed in Appendix~\ref{app:tcga}, pairwise log-rank tests demonstrate significant differences between some clients, but not all.
For the \textbf{Fed-ISIC2019}, we use a 2-dimensional UMAP~\cite{mcinnes2018umap} plot of the features extracted from an Imagenet-pretrained Efficientnetv1 on the raw images.
We see that some clients are isolated in distinct clusters, while others overlap, highlighting the heterogeneity of this dataset.
Last, for the \textbf{Fed-Heart-Disease} dataset, we display histograms for a subset of features (age, resting blood pressure and maximum heart rate), showing that feature distributions vary between clients.

\begin{listing}[t!]
\tiny
\begin{minted}[frame=single,]{Python}
# Import relevant dataset, strategy, and utilities
from flamby.datasets.fed_camelyon16 import FedCamelyon16, Baseline, BaselineLoss, NUM_CLIENTS, metric
from flamby.strategies import FedProx
from flamby.utils import evaluate_model_on_tests, get_nb_max_rounds

# Define number of local updates and number of rounds
num_updates = 100
num_rounds = get_nb_max_rounds(num_updates)
# Dataloaders for train and test
training_dataloaderss = [
    DataLoader(FedCamelyon16(center=i, train=True, pooled=False), batch_size=BATCH_SIZE, shuffle=True)
    for i in range(NUM_CLIENTS)
]
test_dataloaders = [
    DataLoader(FedCamelyon16(center=i, train=False, pooled=False), batch_size=BATCH_SIZE, shuffle=False)
    for i in range(NUM_CLIENTS)
]
# Define local model and loss
model_baseline = Baseline()
loss_baseline = BaselineLoss()
# Define and train strategy
strategy = FedProx(training_dataloaders, model_baseline, loss_baseline, torch.optim.SGD, LR, num_updates, num_rounds)
model_final = strategy.run()[0]
# Evaluate final FL model on test sets
results_per_client = evaluate_model_on_tests(model_final, test_dataloaders, metric)
\end{minted}
\caption{Code example from the FLamby dataset suite: on the Fed-Camelyon16 dataset, we use the FedProx Federated Learning strategy to train the pre-implemented baseline model.\label{list:code_example}}
\end{listing}
\section{FL Benchmark Example with FLamby \label{sec:experiments}}
In this section, we detail the guidelines we follow to perform a benchmark and provide results thereof.
These guidelines might be used in the future to facilitate fair comparisons between potentially novel FL strategies and existing ones.
However, we stress that FLamby also allows for any other experimental setup thanks to its modular structure, as we showcase in Appendices~\ref{app:diff-privacy} and~\ref{app:personalized_fl}.
The FLamby suite further provides a script to automatically reproduce this benchmark based on configuration files.

\textbf{Train/test split.}
We use the per-client train/test splits, including all clients for training.
Performance is evaluated on each local test dataset, and then averaged across the clients.
We exclude model personalization from this benchmark: therefore, a single model is evaluated at the end of training.
We refer to Appendix~\ref{app:personalized_fl} for more results with model personalization.

\textbf{Hyperparameter tuning and Baselines.}
We distinguish two kinds of hyperparameters: those related to the machine learning (ML) part itself, and those related to the FL strategy.
We tune these parameters separately, starting with the machine learning part.
All experiments are repeated with~5 independent runs, except for FED-LIDC-IDRI where only 1 training is performed due to a long training time.

For each dataset, the ML hyperparameters include the model architecture, the loss and related hyperparameters, including local batch size.
These ML hyperparameters are carefully tuned with cross-validation on the pooled training data.
The resulting ML model gives rise to the \textbf{pooled baseline}.
We use the same ML hyperparameters for training on each client individually, leading to \textbf{local baselines}.

For the FL strategies, hyperparameters include e.g. local learning rate, server learning rate, and other relevant quantities depending on the strategies.
For each dataset and each FL strategy, we use the same model as in the pooled and local baselines, with fixed hyperparameters.
We then only optimize FL strategies-related hyperparameters.

\begin{figure}[h!]
    \centering
    \includegraphics[width=\linewidth, trim={0.5cm, 0cm, 0.5cm, 0cm}, clip]{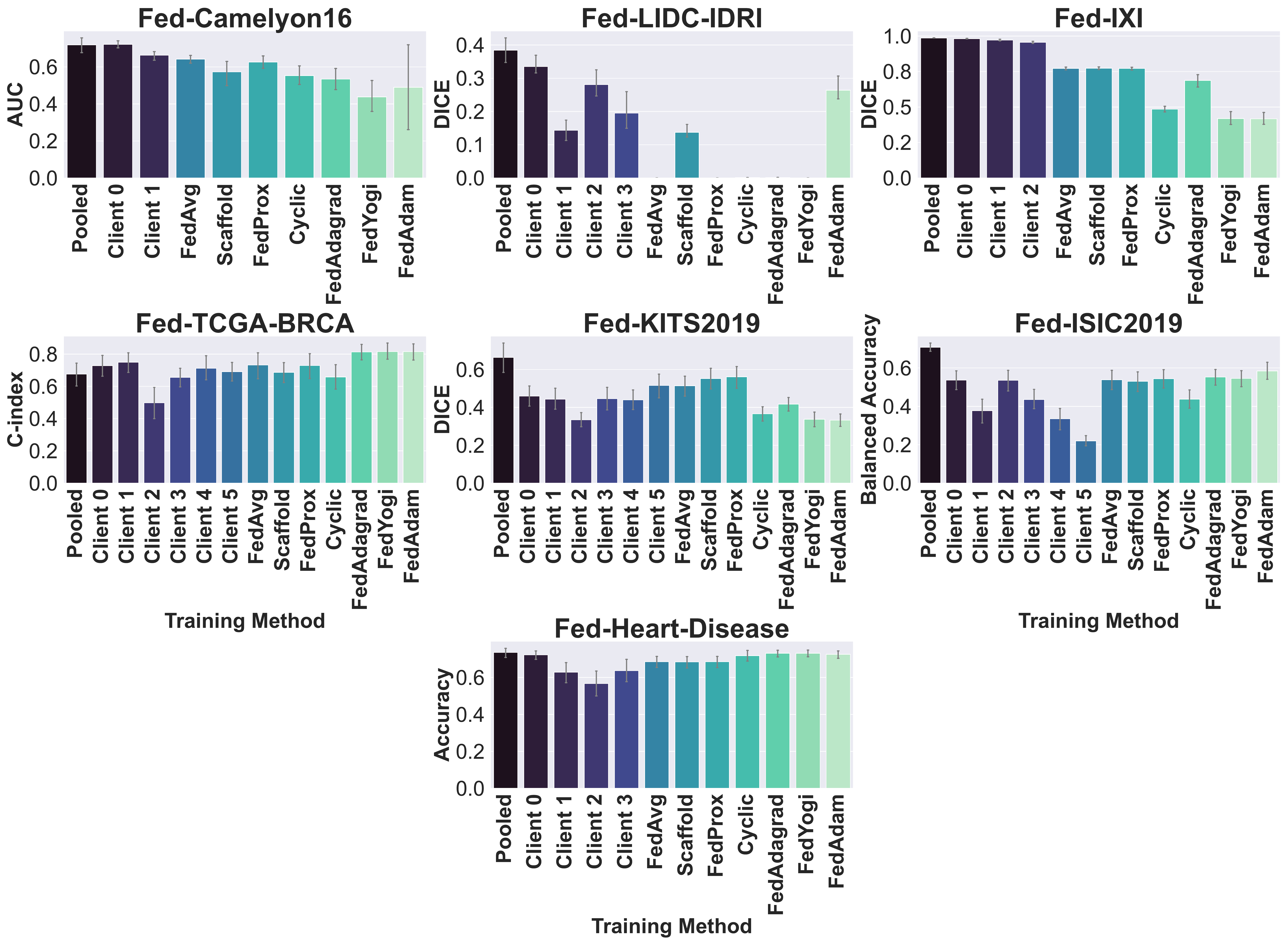}
    \caption{Benchmark results on FLamby for each dataset.
    For all metrics, higher is better, see Section~\ref{subsec:datasets} for metric details and Section~\ref{sec:experiments} for experimental details.
    For Fed-LIDC-IDRI, multiple strategies fail converging, leading to zero DICE.
    Except for Fed-TCGA-BRCA and Fed-Heart-Disease, federated strategies fall short of reaching the pooled performance,
    but improve over the local ones.
    }
    \label{fig:benchmark_results}
\end{figure}

\textbf{Federated setup.}
For all strategies and datasets, the number of rounds $T_{\max}$ is fixed to perform approximately as many epochs on each client as is necessary to get good performance when data is pooled.
Note that, as we use a single batch size $B$ and a fixed number $E$ of local steps, the notion of epoch is ill-defined; we approximate it as follows.
Given $n_{epochs}^{P}$, the number of epochs required to train the baseline model for the pooled dataset, $n_{T}$ the total number of samples in the distributed training set, $K$ the number of clients, we define
\begin{equation}
    T_{\max} = n_{epochs}^{P} \cdot \floor{ n_{T} / K / B / E}
\end{equation}
where $\floor{\cdot}$ denotes the floor operation.
In our benchmark, we use $E=100$ local updates for all datasets.
Note that this restriction in the total number of rounds may have an impact on the convergence of federated strategies.
We refer to Appendix~\ref{app:experiments} for more details on this benchmark.

\textbf{Benchmark results.}
The test results of the benchmark are displayed in Figure~\ref{fig:benchmark_results}.
Note that test results are uniformly averaged over the different local clients.
We observe strikingly different behaviours across datasets.

No local training or FL strategy is able to reach a performance on par with the pooled training, except for Fed-TCGA-BRCA and Fed-Heart-Disease.
It is remarkable that both of them are tabular, low-dimensional datasets, with only linear models.
Still, for Fed-KITS2019 and Fed-ISIC2019, some FL strategies outperform local training, showing the benefit of collaboration, but falling short of reaching pooled performance.
For Fed-Camelyon16, Fed-LIDC-IDRI and Fed-IXI, the current results do not indicate any benefit in collaborative training.

Among FL strategies, we note that for the datasets where an FL strategy outperforms the pooled baselines, FedOpt variants (FedAdagrad, FedYogi and FedAdam) reach the best performance.
Further, the Cyclic baseline systematically underperforms other strategies.
Last, but not least, FedAvg does not reach top performance among FL strategies, except for Fed-Camelyon16 and Fed-IXI, it remains a competitive baseline strategy.

These results show the difficulty of tuning properly FL strategies, especially in the case of heterogeneous cross-silo datasets.
This calls for the development of more robust FL strategies in this setting.

\section{Conclusion \label{sec:conclusion}}
In this article we introduce FLamby, a modular dataset suite and benchmark, comprising multiple tasks and data modalities, and
reflecting the heterogeneity of real-world healthcare cross-silo FL use cases.
This comprehensive benchmark is needed to advance the understanding of cross-silo healthcare data collection on FL performance.

Currently, FLamby is limited to healthcare datasets.
In the longer run and with the help of the FL community, it could be enriched with datasets from other application domains to better reflect the diversity of cross-silo FL applications, which is possible thanks to its modular design.
Regarding machine learning backends, FLamby only provides PyTorch~\cite{paszke2019pytorch} code: supporting other backends, such as TensorFlow~\cite{tensorflow2015-whitepaper} or JAX~\cite{jax2018github}, is a relevant future direction if there is such demand from the community. 
Further, our benchmark currently does not integrate all constraints of cross-silo FL, especially privacy aspects, which are important in this setting.

In terms of FL setting, the benchmark mainly focuses on the heterogeneity induced by natural splits.
In order to make it more realistic, future developments might include in depth study of Differential Privacy (DP) training~\cite{dwork2014algorithmic},
cryptographic protocols such as Secure Aggregation~\cite{bonawitz2017practical}, Personalized FL~\cite{fallah2020personalized},
or communication constraints~\cite{sattler2019sparse} when applicable.
As we showcase in Appendices~\ref{app:diff-privacy} for DP and~\ref{app:personalized_fl} for personalization,
the structure of FLamby makes it possible to quickly tackle such questions.
We hope that the scientific community will
use FLamby for cross-silo research purposes on real data, and 
contribute to further develop it, making it a reference for this research topic.

 \begin{ack}
The authors thank the anonymous reviewers, ethics reviewer, and meta-reviewer
for their feedback and ideas, which significantly improved the paper and the repository.
The authors listed as Owkin, Inc. employees are supported by Owkin, Inc.
The works of \mbox{E.M.} and \mbox{S.A.} is supported, in part, by gifts from Intel and Konica Minolta.
This work was supported by the Swiss State Secretariat for Education, Research
and Innovation (SERI) under contract number 22.00133, by the Inria Explorator
Action FLAMED and by the French National Research Agency (grant ANR-20-CE23-0015,
project PRIDE and ANR-20-THIA-0014 program \Verb"AI_PhD@Lille"). This project has
also received funding from the European Union’s Horizon 2020 research and innovation
programme under grant agreement No 847581 and is co-funded by the Region
Provence-Alpes-C\^{o}te d’Azur and IDEX UCAJEDI. \mbox{A.D.}'s research was supported
by the \textit{Statistics and Computation for AI} ANR Chair and by \textit{Hi!Paris}. 
\mbox{C.P.} received support from \textit{Accenture Labs} (Sophia Antipolis, France).
\end{ack}

\newpage

\medskip
\typeout{}
\putbib[refs]
\end{bibunit}
\newpage
\section*{Checklist}
\begin{enumerate}

\item For all authors...
\begin{enumerate}
  \item Do the main claims made in the abstract and introduction accurately reflect the paper's contributions and scope?
   \answerYes{} 
  \item Did you describe the limitations of your work?
    \answerYes{See Section~\ref{sec:conclusion}.}
  \item Did you discuss any potential negative societal impacts of your work?
     \answerYes{See Section~\ref{sec:broader_impact}.}
  \item Have you read the ethics review guidelines and ensured that your paper conforms to them?
    \answerYes{We point out that we not collect data ourselves. We did a thorough background check on each dataset regarding compliance with these guidelines. We refer to the detailed appendix of each dataset for specific details.}
\end{enumerate}

\item If you are including theoretical results...
\begin{enumerate}
  \item Did you state the full set of assumptions of all theoretical results?
   \answerNA{Our work does not contain theoretical results.}
	\item Did you include complete proofs of all theoretical results?
    \answerNA{Our work does not contain theoretical results.}
\end{enumerate}

\item If you ran experiments (e.g. for benchmarks)...
\begin{enumerate}
  \item Did you include the code, data, and instructions needed to reproduce the main experimental results (either in the supplemental material or as a URL)?
    \answerYes{See Abstract.}
  \item Did you specify all the training details (e.g., data splits, hyperparameters, how they were chosen)?
    \answerYes{See supplementary and code provided.}
	\item Did you report error bars (e.g., with respect to the random seed after running experiments multiple times)?
    \answerYes{We reported error bars as we report the average error on the local test sets across multiple seeds.
    For the largest one, we did not use multiple seeds, but observed empirically a smaller variance in the results due to larger local test set sizes.}
	\item Did you include the total amount of compute and the type of resources used (e.g., type of GPUs, internal cluster, or cloud provider)?
    \answerYes{See Appendix~\ref{app:computing_resources}}
\end{enumerate}

\item If you are using existing assets (e.g., code, data, models) or curating/releasing new assets...
\begin{enumerate}
  \item If your work uses existing assets, did you cite the creators?
    \answerYes{See Section~\ref{sec:flamby}.}
  \item Did you mention the license of the assets?
    \answerYes{See code and supplementary}.
  \item Did you include any new assets either in the supplemental material or as a URL?
   \answerYes{See link in abstract.}
  \item Did you discuss whether and how consent was obtained from people whose data you're using/curating?
     \answerYes{We are only repurposing existing assets. We did a thorough background check on each dataset on this issue. We refer to the detailed appendix of each dataset
    for specific details.}
  \item Did you discuss whether the data you are using/curating contains personally identifiable information or offensive content?
    \answerYes{We are only repurposing existing assets. We did a thorough background check on each dataset on this issue. We refer to the detailed appendix of each dataset
    for specific details.
    }
\end{enumerate}

\item If you used crowdsourcing or conducted research with human subjects...
\begin{enumerate}
  \item Did you include the full text of instructions given to participants and screenshots, if applicable?
     \answerNA{We are only repurposing existing assets.}
  \item Did you describe any potential participant risks, with links to Institutional Review Board (IRB) approvals, if applicable?
     \answerNA{We are only repurposing existing assets.}
  \item Did you include the estimated hourly wage paid to participants and the total amount spent on participant compensation?
     \answerNA{We are only repurposing existing assets.}
\end{enumerate}

\end{enumerate} 
\newpage %
\resetlinenumber
\appendix
\setcounter{page}{1}
\begin{bibunit}[alpha]
\section{Broader Impact \label{sec:broader_impact}}

As this study solely involves the repurposing of existing open-source materials and benchmarking, there are limited risks associated within the study itself.
However, it should be noted that all datasets included in this study could be subject to biases originated during the collection process, such as gender or ethnicity biases.
Unfortunately, on the images' datasets (5 datasets out of 7), the sources of such potential biases cannot be easily checked, since data were properly pseudonymised and image-based medical records cannot be straightforwardly tied back to a particular ethnicity or gender by non-medical experts.
Nevertheless, as our work exposes more clearly some metadata (e.g. geographical origin) of the datasets, it might help revealing underlying geographical biases, and thus help building more heterogeneous benchmarks, as expected in real scenarios for FL.

As we focused on simplicity and ease of use, the current benchmark does not encompass privacy metrics.
However, privacy is of paramount important in healthcare cross-silo FL, and we urge the community not to ignore these aspects.
Thanks to the modularity of FLamby, it is easy to add privacy components, as we show in a DP example in Appendix~\ref{app:diff-privacy}.
Thus, we hope FLamby will help tackle privacy questions in healthcare cross-silo FL. %
\FloatBarrier
\section{Datasets repository and Authors Statement}
\subsection{Dataset repository.}
The code is now available at \hyperlink{https://github.com/owkin/FLamby}{https://github.com/owkin/FLamby}

The code respects best practices for reproducibility and dataset sharing.
The installation process is detailed and allow to install only requirements of specific datasets.
Regarding code readability, the code is linted with black and flake8 and most functions have docstrings.
Documentation is automatically generated from markdown with sphinx, including tutorials.
Unit tests ensure FL strategies perform correctly.

Regarding licenses,
all datasets documented in this repository come with links towards data terms or licenses. 
Every time a user downloads a dataset for the first time, he or she is prompted with a link towards the data terms or license, and has to explicitly agree to it in order to proceed.

\subsection{Maintenance plan}
We will follow a maintenance plan to ensure the code remains correct and the datasets provided by the suite follow adequate standards.
In particular, this maintenance plan encompasses:
\begin{itemize}
    \item Fixing bugs affecting the correctness of the code, whether brought out by the community or ourselves;
    \item Ensuring security updates in the dependencies are performed;
    \item Regarding datasets, reviewing, on a monthly basis, potential updates of the datasets referenced in the suite, including but not limited to patients opting out or ethical concerns raised by the work.
    Such modification may go to the extent of a full revocation of the related dataset if need be;
    \item Reviewing contributions from the community, whether they are related to the benchmark or to incorporating new datasets to the suite, ensuring they are at the highest standards.
\end{itemize}

\subsection{Authors statement.}
As authors of this repository and article we bear all responsibility in case of violation of rights and licenses.
We have added a disclaimer on the repository to invite original datasets creators to open issues regarding any license related matters.

\section{Fed-Camelyon16}
\label{app:camelyon}
\subsection{Description}
Camelyon16~\cite{litjens20181399} is a histopathology dataset of 399 digitized breast biopsies' slides with or without tumor collected from two hospitals: Radboud University Medical Center (RUMC) and University Medical Center Utrecht (UMCU). The client information can be read directly from the training slides as the first 170 slides belong to RUMC and the others to UMCU. For the test slides we use a manual approach based on clustering to recover the centres and visual inspection. 
The slides split are summarized Table~\ref{tab:cam16}

\begin{table}
    \centering
    \caption{Information for the different clients in Camelyon16 \\ ~}
    \label{tab:cam16}
    \begin{tabular}{llccc}
    \toprule
      \textbf{Number} & \textbf{Client} & \textbf{Dataset size} & \textbf{Train} & \textbf{Test} \\
    \toprule
    0 & RUMC & 243 & 169 & 74 \\
    \midrule
    1 & UMCU & 156 & 101 & 55\\
    \bottomrule
    \end{tabular}
\end{table}
\subsection{License and Ethics}
The Camelyon data is open access (CC0)\footnote{\url{https://camelyon17.grand-challenge.org/Data/}}.

The collection of the data was approved by the local ethics committee (Commissie Mensgebonden Onderzoek regio Arnhem - Nijmegen) under 2016-2761, and the need for informed consent was waived~\cite{litjens20181399}.
\subsection{Download and preprocessing}
As the original dataset is stored in Google Drive, we provide code relying on the Google drive API's python SDK to batch download all the images (800GB) efficiently. It requires the user to have a Google account and to setup a service account. Detailed instructions are provided in the repository.

Once all tif images have been downloaded we use the histolab package\cite{histolab} to tile the slides with patches of size 224x224 at the second level of the image pyramid corresponding to $\approx0.5$~{\textmu}m / pixel.
We only keep tiles with sufficient amount of tissue on them thanks to the \mintinline{python}{check_tissue=True} option of the \mintinline{python}{GridTiler} histolab object.
We then perform Imagenet preprocessing~\cite{he2016deep} and extract a 2048 feature vector from an Imagenet-pretrained Resnet50~\cite{he2016deep} on each patch.
As slides have different amount of matter this produces a variable number of features per slide. We subsequently save those features in the numpy format~\cite{van2011numpy}.
\subsection{Task}
Each of this slide represented as a bag of features has a binary label indicating the presence of a tumour on the breast. The task is to predict if a slide contains a tumour or not so it is framed as a binary classification problem under the Multiple Instance Learning paradigm\cite{deepmil}.
\subsection{Baseline, loss function and evaluation}
\paragraph{Loss function}
We use a traditional binary cross entropy loss~\cite{good1992rational} and evaluate the performance with the Area under the ROC curve or AUC~\cite{bradley1997use}.  
\paragraph{Baseline Model}
We use the DeepMIL\cite{deepmil} architecture that uses attention to learn to weight patch features importance in an end to end fashion.
The network architecture is specified in the code. The model trains in approximately 5 minutes on a P100.
\paragraph{Optimization parameters}
We use a batch size of 16 with Adam~\cite{kingma2014adam} with a learning rate of $0.001$.
Both sets of hyperparameters mentioned above used for the network architecture and optimization are taken from~\cite{dehaene2020self}, we change the number of pooled epochs to 45 in order to be able to do more than one synchronization rounds when performing federated experiments.

\paragraph{Hyperparameter Search}
For the pooled dataset benchmark we use the configuration described above without further tuning. For FL strategies we use the following hyperparameter grid:
for clients' learning rates (all strategies) \{1e-5, 1e-4, 1e-3, 1e-2, 1e-1, 1.0\};
for server size learning rate (for Scaffold and FedOpt strategies) \{1e-3, 1e-2, 1e-1, 1.0, 10.0\}, and for FedProx only,
$\mu$ belongs to \{1e-2, 1e-1, 1.0\}.

\section{Fed-LIDC-IDRI}
\label{app:lidc}

\subsection{Description} 

LIDC-IDRI~\cite{armato2011lidc, lidcdata, clark2013cancer} is part of The Cancer Imaging Archive (TCIA) database~\cite{clark2013cancer} with 1009 lung CT-scans (3D images), on which radiologists annotated the presence of nodules.

We split the dataset in 4 different clients that correspond to different medical imagery machine manufacturers, which were previously shown to be a source of heterogeneity in CT image quality~\cite{favazza2015cross}. We end up with 661 samples gathered by GE Medical Systems, 205 by Siemens, 69 by Toshiba, and 74 by Philips scanner.
These datasets are further split in training and testing sets that contain respectively 80\% and 20\% of the data. This split is stratified according to clients, so that proportions are respected.
The exact distribution of the samples between clients are given in Table~\ref{tab:lidc-sizes}

\begin{table}[h!]
    \centering
    \caption{Information for the different clients in LIDC IDRI. \\ ~}
    \label{tab:lidc-sizes}
    \begin{tabular}{llccc}
    \toprule
      \textbf{Number} & \textbf{Client} & \textbf{Dataset size} & \textbf{Train} & \textbf{Test} \\
    \toprule
    0 & GE MEDICAL SYSTEMS & 661 & 530 & 131 \\
    \midrule
    1 & Philips & 74 & 59 & 15\\
    \midrule
    2 & SIEMENS & 205 & 164 & 41 \\
    \midrule
    3 & TOSHIBA & 69 & 55 & 14\\

    \bottomrule
    \end{tabular}
\end{table}

\subsection{License and Ethics}
The users of this data must abide by the Data Usage Policies listed on the TCIA webpage under LIDC (links are provided in the README of the LIDC dataset in FLamby repository). It is licensed under a Creative Commons Attribution 3.0 Unported License.

Data was anonymized in each local center before being uploaded to the central repository~\cite{armato2011lidc}.
Further, as per the terms of use of TCIA\footnote{\url{https://wiki.cancerimagingarchive.net/display/Public/Data+Usage+Policies+and+Restrictions}},
``users must agree not to generate
and use information in a manner that could allow the identities of research participants
to be readily ascertained''.

\subsection{Download and preprocessing}

Instructions in the README.md of the LIDC-IDRI dataset in FLamby repository allow to download images and average annotation masks from the TCIA initiative.
Flamby code then permits conversion from DICOMs to nifti files to facilitate further analysis.

\subsubsection{Preprocessing and sampling}

Raw CT scans have varying dimensions which must be standardized prior to training.
Therefore, as a first step we resize them to a common $(384, 384, 384)$ shape by cropping dimensions in excess and reflection-padding missing dimensions. During training, this operation is performed in the same way both on the CT scans and the ground truth masks.

Next, the images are normalized. CT scan voxels are originally expressed in the Hounsfield unit (HU)~\cite{feeman2010mathematics} scale: roughly $-1,000$ HU for air, $0$ HU for water, and $1,000$ HU for bone. We clip the images to the $[-1024, 600]$ range, add $1024$, and then divide voxels by $1624$ to obtain values ranging in $[0, 1]$.

\subsection{Task}

We benchmark federated learning strategies on a nodule segmentation task using a VNet~\cite{milletari2016v}. More precisely, we aim to maximize the DICE coefficient \cite{dice1945measures} between predictions and the annotated ground truths. For reference, the baseline model trained on the pooled training set achieves a DICE of $41\%$ on the pooled test set.

\subsection{Baseline, loss function and evaluation}

\paragraph{Sampling} 
The resulting images of size $(384, 384, 384)$ are too voluminous to fit in the memory of most GPUs. Hence, during training we feed the model with sampled patches of size $(128, 128, 128)$. We sample $2$ patches from each (image, mask) pair. This implies that batches are constituted of two~$(128, 128, 128)$ patches drawn from the same CT scan. 
As lung nodules are relatively small and rare, there is a strong class imbalance in the LIDC dataset. To alleviate this issue, we ensure that one of the sampled patches contains nodule voxels (by centering it on a nodule voxel drawn at random), and sample the other completely at random. To account for possible nodules at the edges of CT scans, a padding of half the patch size is applied to each dimension of the image prior to sampling.

\paragraph{Loss function} 

Our objective is to maximize the DICE coefficient~\cite{dice1945measures}. However, we observed that maximizing DICE alone during training yielded poor results at inference time on regions that do not contain nodules.
To force the model to account for class imbalance, we added a small balanced cross-entropy term \citep[see][]{jadon2020survey}. Hence, we minimize the following loss:
\begin{equation}
    \mathcal{L}(\mathbf{y}, \hat{\mathbf{y}}) = (1 - \mathrm{DICE}(\mathbf{y}, \hat{\mathbf{y}})) + 0.1 \times \mathrm{BCE}(\mathbf{y}, \hat{\mathbf{y}}),
\end{equation}
with
\begin{equation}
     \mathrm{DICE}(\mathbf{y}, \hat{\mathbf{y}}) =  \frac{2 \sum_{i=1}^n y_i \hat{y}_i}{\sum_{i=1}^n y_i  + \sum_{i=1}^n \hat{y}_i},
\end{equation}
and
\begin{equation}
     \mathrm{BCE}(\mathbf{y}, \hat{\mathbf{y}}) = - \alpha \sum_{i=1}^n y_i \log(\hat{y}_i) - \sum_{i=1}^n (1 - y_i) \log(1 - \hat{y}_i),
\end{equation}
where $\alpha = \left(\max(\frac{1}{n}\sum_{i=1}^n y_i, 10^{-7})\right)^{-1}  - 1.$

\paragraph{Baseline Model}

We implement a VNet \cite{milletari2016v}, following the architecture proposed therein. During training, we use dropout ($p=0.25$). The final layer produces a single output, which is passed through a sigmoid function to encode the probability that each voxel corresponds to a nodule. The model trains in approximately 48 hours on a P100.

\paragraph{Optimization parameters} 

We optimize the VNet using RMSprop, with an initial learning rate of~$10^{-2}$. We run $100$ epochs, multiplying the learning rate by $0.95$ every $10$ epochs.

\paragraph{Hyperparameters search}

LIDC FL trainings take approximately 70 hours on a P100 so because of time constraints we could not use an extensive grid search as for other datasets. The final parameters we use are reported in Section~\ref{sec:hp}.  

\section{Fed-IXI}
\label{app:ixi}
\subsection{Description}
IXI Tiny~\cite{ixitiny} is a light version of the dataset IXI, a multimodal brain imaging dataset of almost~600 subjects~\cite{ixi}.
This lighter version provides T1-weighted brain MR images for a subset of 566 subjects, along with a set of corresponding brain image segmentations labels, taking the form of binary image masks.

Brain image masks isolate the brain pixels from the other head components, such as the eyes, skin, and fat.
For the supervised task, brain image segmentation masks (labels) were obtained through automatic whole-brain extraction on the T1-weighted MRI data, using the unsupervised brain extraction tool ROBEX \cite{iglesias2011robust}.

The images come from three different London hospitals: Guys (Guy’s Hospital, manufacturer code~0), HH (Hammersmith Hospital, manufacturer code 1), both using a Philips 1.5T system, and IOP (Institute of Psychiatry, manufacturer code 2), using a GE 1.5T system.
We split this dataset in training and testing sets, respectively containing 80\% and 20\% of the data. The split is also stratified according to hospitals to preserve data proportions. In other words, we define one test set on each hospital.
Table~\ref{tab:ixi-demographics} provides demographic information for this dataset.

\begin{table}
    \centering
    \caption{Demographics information for Fed-IXI. \\ ~}
    \label{tab:ixi-demographics}
    \begin{tabular}{llccc}
    \toprule
      \textbf{Hospital Name} &  & & &   \\
     &  \textbf{Sex}   &  \textbf{Dataset size} &                \textbf{Age} & \textbf{Age Range}    \\
    \toprule
    Guys & Female & 184 &  53.23 $\pm$ 15.25 &   20 - 80 \\
         & Male   &  144 &  51.02 $\pm$ 17.26 &   20 - 86 \\
    \midrule
    HH & Female &  93 &  50.28 $\pm$ 16.93 &   20 - 81 \\
       & Male  &  85 &  44.43 $\pm$ 15.67 &   20 - 73 \\
    \midrule
    IOP & Female & 44 &  43.90 $\pm$ 18.43 &   19 - 86 \\
        & Male  &  24 &  39.57 $\pm$ 12.46 &   23 - 70 \\
    \bottomrule
    \end{tabular}
\end{table}

\subsection{License and Ethics}
This dataset is licensed under a Creative Commons Attribution Share Alike 3.0 Unported (CC BY-SA~3.) license.

The dataset website does not provide any information regarding data collection ethics.
However, the original dataset was collected as part of the
IXI - Information eXtraction from Images (EPSRC GR/S21533/02) project,
and thus funded by UK Research and Innovation (UKRI).
As part of its terms and conditions\footnote{\url{https://www.ukri.org/wp-content/uploads/2022/04/UKRI-050422-FullEconomicCostingGrantTermsConditions-Apr2022.pdf}},
the UKRI demands that all funded
projects are ``carried out in accordance with all applicable ethical, legal and
regulatory requirements'' (RGC 2.2).

\subsection{Downloading and preprocessing}

We provide a helper script to download the dataset from an Amazon S3 bucket.

\paragraph{Preprocessing and sampling}
We use a fixed preprocessing step that is performed once.
Brain scans are first geometrically aligned to a common anatomical space (MNI template) through affine registration estimated with NiftyReg \cite{niftyreg}. Images are then reoriented using ITK \cite{itk}.
Finally, intensities are normalized in each image (based on the entire image histogram), and the image volumes are resized from 83x44x55 to 48×60×48 voxels.

\subsection{Task}

The task is to segment the brain on the volume.
The prediction is evaluated with the DICE score, which is the symmetric of the DICE loss with respect to $1/2$.
\subsection{Baseline, loss function and evaluation}
\paragraph{Loss function}
The model was directly trained for the DICE loss~\cite{dice1945measures}, defined as
\begin{equation*}
    \ell_{DICE} = 1 - S_{DICE} =
    1 - \frac{2 \mathrm{TP}}{2 \mathrm{TP} + \mathrm{FP} + \mathrm{FN} + \epsilon},
\end{equation*}
where $\mathrm{TP}$, $ \mathrm{FP}$, and $ \mathrm{FN}$ stand for the true positive rate, false positive rate, and false negative rate, respectively, and $\epsilon = 10^{-9}$ ensures numerical stability.
\paragraph{Baseline Model}
We use a UNet model taking the individual T1 image as input, to predict the associated binary brain mask. The UNet model is a standard type of convolution neural network architecture commonly used in biomedical image segmentation tasks \cite{ronneberger2015u}. It is specifically used to perform semantic segmentation, meaning that each voxel of the image volume is classified. We can also refer to this task as a dense prediction. The model trains in approximately 5 minutes on a P100.

\paragraph{Optimization parameters}
The UNet is optimized with a batch size of $2$ and a learning rate of $10^{-3}$ with the AdamW optimizer.
The best architecture used batch normalization, max-pooling, linear upsampling, zero-padding of size 1, PReLU activation functions, and 3 encoding blocks.  

\paragraph{Hyperparameters search}
We do not change parameters for the pooled baseline.
For FedAvg and Cyclic, we optimized the learning rate over the values \{0.1, 0.01, 0.001, 0.0001, 0.00001\}.
For FedYogi, FedAdam, FedAdagrad and Scaffold, our search grid space was \{0.1, 0.01, 0.001, 0.0001, 0.00001\} and \{10, 1, 0.1, 0.01, 0.001\} for the learning rate and the server learning rate respectively.
For FedProx, our search space contained \{0.1, 0.01\} and \{1, 0.1, 0.01\} sets for learning rate and $\mu$ respectively.

\section{Fed-TCGA-BRCA}
\label{app:tcga}

\subsection{Description}

Our dataset comes from The Cancer Genome Atlas (TCGA)'s Genomics Data Commons (GDC) portal~\cite{tcga} more specifically from the BReast CAncer study (BRCA), which includes features gathered from 1066 patients. We use the material produced by Liu \textit{et al.}~\cite{liu2018integrated} as a base file that we further preprocess with one-hot encoding following~\cite{andreux2020federated}.
This produces a lightweight tabular dataset with 39 input features.
Patients' labels are overall survival time and event status with the event being death.
We use the Tissue Source Site metadata to split data based on extraction site, grouped into geographic regions to obtain large enough clients.
We end up with \textbf{6} clients: USA~(Northeast, South, Middlewest, West), Canada and Europe,
with patient counts varying from~51 to~311.
Our train-test split of the data is stratified per client and event.
Table~\ref{tab:tcga-details} provides details per client for this dataset.
Table~\ref{tab:tcga-logrank} provides results of pair-wise log-rank tests between the different clients.

\subsection{License and Ethics}
The data terms can be found on the GDC website\footnote{\url{https://gdc.cancer.gov/access-data/data-access-processes-and-tools}}.
In particular, these terms bind users as to ``not attempt to identify individual human
research participants from whom the data were obtained''.

As per the TCGA policies\footnote{\url{https://www.cancer.gov/about-nci/organization/ccg/research/structural-genomics/tcga/history/policies}}, special care was devoted to ensure privacy protection of research subjects, including but not limited to HIPAA compliance.
Note that we do not use the genetic part of TCGA whose access is restricted due to its sensitivity.

\subsection{Downloading and preprocessing}

The pooled TCGA-BRCA dataset requires no downloading or extra-preprocessing as the preprocessed data is now a part of the Flamby repository.

\begin{table}[h!]
    \centering
    \caption{Information for the different clients (geographical regions) in Fed-TCGA-BRCA. \\ ~}
    \label{tab:tcga-details}
    \begin{tabular}{llcccc}
    \toprule
      \textbf{Number}&\textbf{Client}&\textbf{Dataset size}&\textbf{Train}&\textbf{Test}&\textbf{Censorship ratio}\\
    \toprule
    0 & USA Northeast & 311 & 248 & 63 & 81\\
    \midrule
    1 & USA South & 196 & 156 & 40 & 80\\
    \midrule
    2 & USA West & 206 & 164 & 42 & 89\\
    \midrule
    3 & USA Midwest & 162 & 129 & 33 & 88\\
    \midrule
    4 & Europe & 162 & 129 & 33 & 94\\
    \midrule
    5 & Canada & 51 & 40 & 11 & 94\\
    
    \bottomrule
    \end{tabular}
\end{table}

\begin{table}[h!]
    \centering
    \caption{Pairwise log-rank $p$-values on the Fed-TCGA-BRCA clients. Some clients have significant differences for a 10\% significance threshold.}
    \label{tab:tcga-logrank}
\begin{tabular}{llllll}
\toprule
{} &   Client 1 &   Client 2 &   Client 3 &  Client 4 &   Client 5 \\
Compared with &           &           &           &           &           \\
\midrule
Client 0     &  0.289682 &  0.066374 &  0.039892 &  0.576926 &  0.200366 \\
Client 1     &       &  0.192075 &  0.161797 &   0.92917 &  0.541251 \\
Client 2     &       &       &  0.954475 &  0.720912 &  0.256973 \\
Client 3     &       &      &       &  0.576374 &  0.127662 \\
Client 4     &       &     &      &   &  0.441106 \\
\bottomrule
\end{tabular}
\end{table}

\subsection{Task}

The task consists in predicting survival outcomes~\cite{jenkins2005survival} based on the patients' clinical tabular data~(39 features overall).
This survival task is akin to a ranking problem with the score of each sample being known either directly or only by lower bound.
Indeed, some patients leave the study before the event of interest is observed, and are labelled  as right-censored. Survival analysis aims at solving this type of ranking problem while leveraging  right-censored data.
The censoring ratio in the TCGA-BRCA study is $86\%$.

The ranking is evaluated by using the concordance index (C-index) that measures the percentage of correctly ranked pairs while taking censorship into account:
\begin{equation}
\mathrm{C-index} = 
\mathbb{E}_{\substack{i: \delta_i = 1 \\ j: t_j > t_i}}  \big[ \mathbbm{1}_{[ \eta_j < \eta_i ]}  \big]
\end{equation}
where $\eta_i$ is a risk score assigned by our model to a patient $i$. In our case of linear Cox proportional hazard models we use $\eta_i = \beta^T x_i $,
where $x_i$ are the features for patient $i$
and $\beta$ the learned weights, see Section~\ref{app:subsec:survival_analysis_background}.

\paragraph{Optimization parameters}
For the pooled dataset benchmark, we use the Adam optimizer ~\cite{kingma2014adam}, with a learning rate of $0.1$ and a batch size of $8$ for $30$ epochs.

\subsection{Baseline, loss function and evaluation}
\label{app:subsec:survival_analysis_background}
\paragraph{Survival analysis background}
Let $T$ be the random time-to-death taken from the patient's population under study. The survival function $S$ is defined as:
\begin{equation}
S(t)=Pr[T>t]
\end{equation}
A patient is characterized by its vector of covariates $x$ (clinical data in our case), an observed time point $t$ and an indicator $\delta \in \{0,1\}$ where $\delta = 0$ if the event has been censored.
A key quantity characterizing the distribution of $S$ is the hazard function $h$. It is the instantaneous rate of occurrence of the event given that it has not yet happened for a patient:
\begin{equation}
h(t,x) = \lim_{dt \to 0} \frac{Pr[t < T < t + dt | x , T > t]}{dt}
\end{equation}

\paragraph{Loss function}

The simplest model in survival analysis is the linear Cox proportional hazard~\cite{cox1972regression}. This model assumes:
\begin{equation}
h(t,x) = h_0(t) \exp(\beta^T x)
\end{equation}
where $h_0$ is the baseline hazard function (common to all patients and dependent on time only) and $\beta$ is the vector of parameters of our linear model.
$\beta$ is estimated by minimization of the negative Cox partial log-likelihood, which compares relative risk ratios:
\begin{equation}
L(\beta) = - \sum_{i: \delta_i = 1} \Big[ \beta^T x_i - \log \big( \sum_{j: t_j \geq t_i} \exp( \beta^T x_j ) \big) \Big]
\end{equation}
where $i$ and $j$ index patients.

We minimize the negative Cox partial log-likelihood by gradient descent w.r.t. $\beta$.

As explained in~\cite{andreux2020federated} the Cox partial log-likelihood is not separable with respect to the samples: this means it cannot be expressed as a sum of terms each dependent on a single sample. Hence it is not separable with respect to the clients either.
In this work, for simplicity, we decide to ignore this fact in the baseline: we treat each client's negative Cox partial log-likelihood independently of the others and apply any federated learning strategy logic to the resulting local gradients. Please refer to~\cite{andreux2020federated} for a more rigorous treatment of the federated survival analysis problem.

\paragraph{Baseline Model}
As a baseline, we use the aforementioned linear Cox proportional hazard model~\cite{cox1972regression}. The model trains in a matter of seconds on modern CPUs.

\paragraph{Hyperparameter Search}

All the federated learning strategies are tested with the SGD optimizer.
We performed a grid search for the federated learning strategies hyperparameters .
For FedAvg and Cyclic, we optimized the learning rate over the values \{0.1, 0.01, 0.001, 0.0001, 0.00001\}.
For FedYogi, FedAdam, FedAdagrad and Scaffold, our search grid space was \{0.1, 0.01, 0.001, 0.0001, 0.00001\} and \{10, 1, 0.1, 0.01, 0.001\} for the learning rate and the server learning rate respectively.
For FedProx, our search space contained \{0.1, 0.01\} and \{1, 0.1, 0.01\} sets for learning rate and $\mu$ respectively.
The chosen hyperparameters from the HP search can be found in Section~\ref{sec:hp}.

\section{Fed-KiTS19}
\label{app:kits}
\subsection{Description}

\begin{figure}[t!]
\centering
\begin{subfigure}[b]{0.45\textwidth}
\includegraphics[max width=\textwidth]{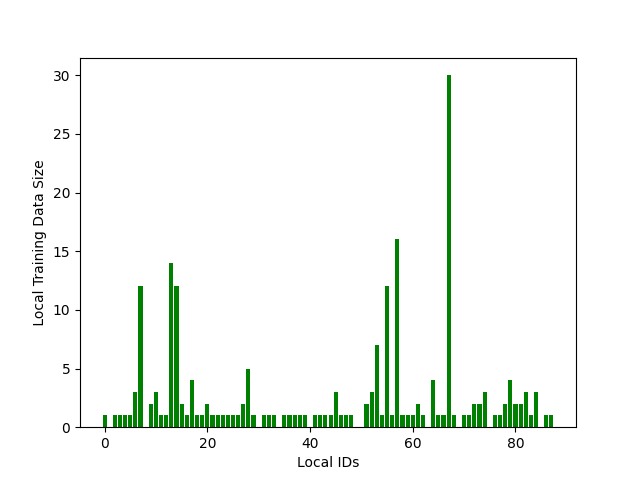}
\caption{Original KiTS19}
\label{subfig:kits19_org}
\end{subfigure}
\begin{subfigure}[b]{0.45\textwidth}
\includegraphics[max width=\textwidth]{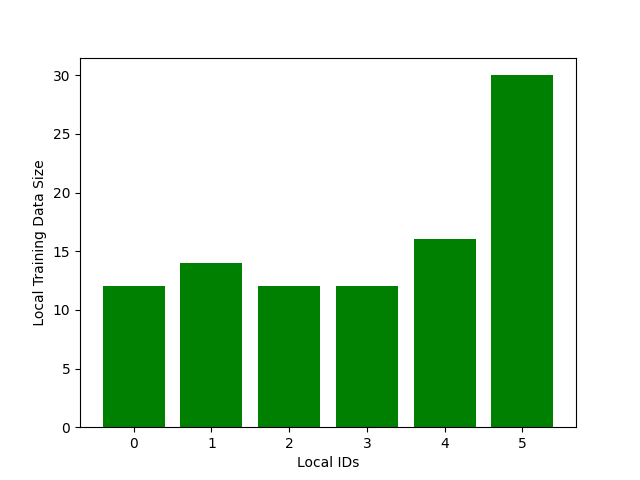}
\caption{Fed-KiTS19}
\label{subfig:fed_kits19}
\end{subfigure}
\label{data_distribution}
\caption{Patient distributions across hospitals of the original KiTS19 Dataset and the derived Data distribution of Fed-KiTS19} 
\end{figure}

The KiTS19 dataset ~\cite{heller2020state, heller2019kits19} stems from the Kidney Tumor Segmentation Challenge 2019 and contains CT scans of 210 patients along with the segmentation masks from 77 hospitals\footnote{It is important to note that KiTS19 dataset does not come with the hospital information. We obtained the data distribution per client from one of the organizers of this challenge, Nicholas Heller, over email communication. We acknowledge the help of Nicholas Heller for sharing this valuable resource with us that helped us explore federated learning strategies with this dataset for the first time.}.
We only consider the training dataset of this challenge as the segmentation masks are not provided for the test dataset. We recover the hospital metadata and extract a \textbf{6}-client federated version of this dataset by removing hospitals with less than $10$ training samples. Figures~\ref{subfig:kits19_org} and~~\ref{subfig:fed_kits19} show the repartition of patients per client before and after this client selection respectively. Table~\ref{tab:kits19-details} provides further details of the train and test split at each selected client.

\subsection{License and Ethics}
This dataset is licensed under a Attribution-NonCommercial-ShareAlike 4.0 International (CC-BY-NC-SA) license\footnote{\url{https://github.com/neheller/kits19/blob/master/LICENSE}}.

The dataset collection was approved by the Institutional Review Board at the University of
Minnesota as Study 1611M00821~\cite{heller2019kits19}.
\subsection{Downloading and Preprocessing}
\label{subsec:preprocesskits}
We use the official KiTS19 repository\footnote{\url{https://github.com/neheller/kits19}} to download the KiTS19 data. Next, we preprocess this dataset. The first step of the preprocessing is to clip the intensity. We clip the values of each image to the [5th percentile, 95th percentile] range, where 5th percentile and 95th percentile are calculated on the image intensities of each patient's case separately. After this step, we apply z-scale normalization, where we subtract the mean and divide by the standard deviation of the image intensities. Since KiTS19 dataset comes with inhomogeneous voxel spacing even for the patients data from the same silos, we resample the voxel spacings to the target spacing of 2.90x1.45x1.45 mm for all the samples.
\subsection{Task}
The task consists of both kidney and tumor segmentation, labeled 1 and 2, respectively. The background is labeled as 0. The score we consider on this dataset is the average of Kidney and Tumor DICE scores~\cite{dice1945measures}.

\subsection{Baseline, loss function and evaluation}
\paragraph{Sampling}
The image size distribution of the samples of the KiTS19 dataset is heterogeneous. After the resampling detailed in section \ref{subsec:preprocesskits}, the median patient's data size is [116, 282, 282]. To make our model's computation memory efficient, we extract a patch of size [64, 192, 192] from each sample during the model training. The number of voxels belonging to the foreground classes (i.e. either Kidney or Tumor) is small compare to the number of voxels belonging to the background class. Therefore, we oversample the foreground classes when taking patches of the samples. More precisely, we use batches of size 2. Each batch contains one patch with the foreground oversampled. Furthermore, we split each silo's data into training and validation data with 80\% and 20\% split, respectively. All this pre-processing and patching is done using the nnU-Net library~\cite{isensee2021nnu}. 

\begin{table}
    \centering
    \caption{Information for the selected clients in Fed-KiTS19. \\ ~}
    \label{tab:kits19-details}
    \begin{tabular}{lccc}
    \toprule
      \textbf{Local ID Number}&\textbf{Dataset size}&\textbf{Train}&\textbf{Test}\\
    \toprule
    0 & 12 & 9 & 3 \\
    \midrule
    1 &  14 & 11 & 3\\
    \midrule
    2  & 12 & 9 & 3\\
    \midrule
    3 & 12 & 9 & 3\\
    \midrule
    4 & 16 & 12 & 4\\
    \midrule
    5 & 30 & 24 & 6\\
    
    \bottomrule
    \end{tabular}
\end{table}

\paragraph{Loss function}
We use the same loss function as proposed by nnU-Net ~\cite{isensee2021nnu} for the KiTS19 dataset which is based on DICE~\cite{dice1945measures} and on the Cross Entropy loss. Both losses are summed with equal weight as shown in Equation~\eqref{eq:kits19_loss}, 

\begin{equation}
    \mathcal{L}(\mathbf{y}, \hat{\mathbf{y}}) = \mathrm{CE}(\mathbf{y}, \hat{\mathbf{y}}) - \mathrm{DICE}(\mathbf{y}, \hat{\mathbf{y}}),
    \label{eq:kits19_loss}
\end{equation}
with
\begin{equation}
     \mathrm{DICE}(\mathbf{y}, \hat{\mathbf{y}}) =  \frac{2 \sum_{l=1}^2\sum_{i=1}^n y^{l}_i \hat{y}^{l}_i +\epsilon}{\sum_{l=1}^2(\sum_{i=1}^n y^{l}_i  + \sum_{i=1}^n \hat{y}^{l}_i) +\epsilon},
\end{equation}
and
\begin{equation}
     \mathrm{CE}(\mathbf{y}, \hat{\mathbf{y}}) =  -\sum_{i=1}^n\sum_{l=1}^2 y^{l}_i \log \hat{y}^{l}_i,
\end{equation}
where $\epsilon$ value is $1e^{-5}$ and n is the set of all pixels and 2 signifies the 2 class labels here, Kidney and Tumor, and where $y^{l}_i$
 is the one-hot encoding (0 or 1) for the label l and pixel $i$ and $y^{l}_i$  
 is the predicted probability for the same label l and pixel $i$.
 \paragraph{Baseline Model}
 During the training, we use nnU-Net~\cite{isensee2021nnu}, with the architecture proposed therein for the KiTS19 dataset. We chose convolution kernels of sizes [[3,3,3],[3,3,3],[3,3,3],[3,3,3],[3,3,3]] and pool kernels of sizes [[2,2,2],[2,2,2],[2,2,2],[2,2,2],[1,2,2]].
  The model trains in under 24 hours on a P100.
  
 \paragraph{Optimization parameters}
 In addition, we use Adam optimizer ~\cite{kingma2014adam} with a learning rate of 0.0003 for 500 epochs to train our model. To evaluate the performance of the trained model, we evaluate the DICE score on the validation data for both classes, Kidney and tumor, and report the average of these two scores. We note that with 8000 epochs we can obtain higher performances, however at the expense of computational cost.
\paragraph{Hyperparameter Search}
For the pooled strategy results, we use the Adam Optimizer and 0.0003 learning rate, as used in nnU-Net work for KiTS19 dataset \cite{isensee2021nnu}. For Cyclic and FedAvg, we optimized the learning rate over the values \{0.3, 0.03, 0.003, 0.0003, 0.00003\} and found that a learning rate of 0.3 provided the best results for both strategies.
For FedYogi, FedAdam, FedAdagrad and Scaffold, our search grid space was \{0.1, 0.01\} and \{0.001, 0.01, 0.1, 1\} for the learning rate and the server learning rate respectively. In the best setting, the learning rate was $0.1$ for all these strategies, and the server learning rate $0.1$ for FedAdagrad, $0.01$ for FedYogi and FedAdam and $1$ for Scaffold. Likewise, for FedProx, our search space contained \{0.1, 0.01\} and \{0.001, 0.01, 0.1, 1\} sets for learning rate  and $\mu$ respectively, and the best set of hyperparameters was $0.1$ for the learning rage and $0.001$  for $\mu$.

\section{Fed-ISIC2019}
\label{app:isic}

\subsection{Dataset description}

The ISIC2019 challenge dataset~\cite{tschandl2018ham10000,codella2018skin,combalia2019bcn20000} contains 25,331 dermoscopy images collected in~4 hospitals.
To the best of our knowledge, it is the largest public dataset of high-quality images of skin lesions.
We restrict ourselves to~23,247 images from the public train set due to metadata availability reasons, which we re-split into train and test sets. The train-test split is static.

We split this dataset into 6 clients corresponding to different sites where images were taken with different imaging technologies. The ViDIR Group, Medical University of Vienna, Austria uses 3 different imaging systems representing evolving clinical practice: a Heine Dermaphot system using an immersion fluid, a DermLite™ FOTO and a MoleMax HD machine which gives rise to 3 clients.
On top of this, the skin cancer practice of Cliff Rosendahl in Queensland, Australia, the Hospital Cl\'{i}nic de Barcelona, Spain and the Memorial Sloan Kettering Cancer Center, New York give rise to 3 other different clients making a total of 6 clients.
The biggest client counts 12413 images while the smallest counts 439. Table~\ref{tab:isic-sizes} provides details about the size of the different clients.
\begin{table}[h!]
    \centering
    \caption{Information for the different clients in Fed-ISIC2019. \\ ~}
    \label{tab:isic-sizes}
    \resizebox{\textwidth}{!}{
    \begin{tabular}{llccc}
    \toprule
      \textbf{Number} & \textbf{Client} & \textbf{Dataset size} & \textbf{Train} & \textbf{Test} \\
    \toprule
    0 & Hospital Cl\'{i}nic de Barcelona & 12413 & 9930 & 2483\\
    \midrule
    1 & ViDIR Group, Medical University of Vienna (MoleMax HD) & 3954 & 3163 & 791\\
    \midrule
    2 & ViDIR Group, Medical University of Vienna (DermLite FOTO) & 3363 & 2691 & 672\\
    \midrule
    3 & The skin cancer practice of Cliff Rosendahl & 2259 & 1807 & 452\\
    \midrule
    4 & Memorial Sloan Kettering Cancer Center & 819 & 655 & 164\\
    \midrule
    5 & ViDIR Group, Medical University of Vienna (Heine Dermaphot) & 439 & 351 & 88\\
    
    \bottomrule
    \end{tabular}}
\end{table}

\subsection{License and Ethics}
This dataset is licensed under a Attribution-NonCommercial 4.0 International (CC BY-NC 4.0) license\footnote{\url{https://challenge.isic-archive.com/data/}}.

As per the terms of use of the ISIC archive\footnote{\url{https://challenge.isic-archive.com/terms-of-use/}},
one of the requirements for this dataset to have been hosted is that it is properly de-identified
in accordance with applicable requirements and legislations.
\subsection{Downloading and preprocessing}

Instructions for downloading and preprocessing are available in the README of the Fed-ISIC2019 dataset inside the FLamby repository.
As an offline preprocessing step, we follow recommendations and code from~\cite{aman} by resizing images to the same shorter side of 224 pixels while maintaining their aspect ratio, and by normalizing images' brightness and contrast through a color consistency algorithm.
The total size of the raw inputs is $9$ GB.

\subsection{Task}

The task consists in image classification among~8 different classes: Melanoma, Melanocytic nevus, Basal cell carcinoma, Actinic keratosis, Benign keratosis, Dermatofibroma, Vascular lesion and Squamous cell carcinoma.
Ground truth is established through histopathology, follow-up examination, expert consensus or microscopy.
The ISIC2019 dataset has a high label imbalance with prevalence ranging from 49\% to less than 1\% depending on the class.
We follow the ISIC challenge metric: we measure classification performance through balanced accuracy, defined as the average of the recalls calculated for each class.
For balanced datasets, it is equal to accuracy.
A random classifier would get a balanced accuracy equal to $1 / C$ where $C$ is the number of classes.
Using balanced accuracy prevents the model from taking advantage of an imbalanced test set.

\subsection{Baseline, loss function and evaluation}

The choices made are inspired by~\cite{aman}, ~\cite{DBLP:journals/corr/abs-1910-03910}, and an analysis of the solutions that scored well at the ISIC challenge over the years.
\paragraph{Loss function}
Our pretrained EfficientNet is fine-tuned using a weighted focal loss~\cite{DBLP:journals/corr/abs-1708-02002}.
It is calculated as follows:
\begin{equation}
FL(p_t) = - \alpha_t (1 - p_t)^\gamma \log(p_t)
\end{equation}
where $p_t$ is the probability output by our model for the ground-truth class, $\alpha_t$ is the weight of the ground-truth class (a weight is attributed to each class before training), $\gamma$ is a hyperparameter (chosen at $2$ in our work).
The focal loss is very useful where there is class imbalance.
To provide an intuition behind this focal loss, compared to Binary Cross Entropy, it gives the model a bit more freedom to take some risk when making predictions.
The weights we use for our weighted focal loss are the inverse of the class proportions calculated over the pooled dataset.
We assume these weights are available to all clients.

\paragraph{Baseline Model}
As a baseline classification model, we fine-tune an EfficientNet~\cite{DBLP:journals/corr/abs-1905-11946}.
EfficientNets are the results of a simple uniform scaling of MobileNets and ResNet on all dimensions (depth/width/resolution).
They show great accuracy and efficiency and transfer very well to other tasks.
Our EfficientNet is pretrained on ImageNet, we use it as a feature extractor ($1280$ features) by replacing the output layer by a linear layer to get an output of dimension $8$.
On top of this, we use the data augmentations listed below to regularize our model.
The model trains in under an hour on a P100, because we have to recompute EfficientNet features with dynamic data augmentations.

For training:
\begin{enumerate}
    \item Random Scaling
    \item Rotation
    \item Random Brightness Contrast
    \item Flipping
    \item Affine deformation
    \item Random crop
    \item Coarse Dropout
    \item Normalization
\end{enumerate}
At test time:
\begin{enumerate}
    \item Center cropping
    \item Normalization
\end{enumerate}
\paragraph{Optimization parameters}
For the pooled dataset benchmark, we use the Adam optimizer ~\cite{kingma2014adam} with a learning rate of $5 \times 10^{-4}$ and a batch size of $64$ for $20$ epochs.

\paragraph{Hyperparameter Search}

All the federated learning strategies are tested with the SGD optimizer. 
We performed a grid search for the federated learning strategies hyperparameters.
For FedAvg and Cyclic, we optimized the learning rate over the values \{1e-3, 1e-2.5, 1e-2, 1e-1.5, 1e-1, 1e-0.5\}.
For FedYogi, FedAdam, FedAdagrad and Scaffold, our search grid space was \{1e-3, 1e-2.5, 1e-2, 1e-1.5, 1e-1, 1e-0.5\} and \{1e-3, 1e-2.5, 1e-2, 1e-1.5, 1e-1, 1e-0.5, 1, 1e-0.5, 10\} for the learning rate and the server learning rate respectively.
For FedProx, our search space contained \{1e-3, 1e-2.5, 1e-2, 1e-1.5, 1e-1, 1e-0.5\} and \{0.001, 0.01, 0.1, 1.\} sets for learning rate and $\mu$ respectively.
The chosen hyperparameters from the HP search can be found in Sec.~\ref{sec:hp}.

\section{Fed-Heart-Disease}
\label{app:heart}
\subsection{Description}
The Heart Disease dataset contains records from 920 patients from four
hospitals in the USA, Hungary, and Switzerland. There are 13 features
before preprocessing: age, sex, chest pain type, resting blood
pressure, serum cholesterol, blood sugar, resting
electrochardiographic results, maximum heart rate, exercise induced
angina, ST depression induced by exercise, slope of the peak ST
segment, number of major vessels, and thalassemia background. All
features are continuous or binary, except for chest pain type (four
categories) and resting electrochardiographic results (three
categories).  The target is the presence of a heart disease.
After preprocessing, we are left with 740 records, each having 13~features.
They are split in train and test in a stratified manner. Distribution 
of the data records among clients is described in 
Table~\ref{tab:heart-disease-sizes}.

\begin{table}[h!]
    \centering
    \caption{Distribution of data records among the different clients in Fed-Heart-Disease. \\ ~}
    \label{tab:heart-disease-sizes}
    \begin{tabular}{llccc}
    \toprule
      \textbf{Number} & \textbf{Client} & \textbf{Dataset size} & \textbf{Train} & \textbf{Test} \\
    \toprule
    0 & Cleveland's Hospital & 303 & 199 & 104\\
    \midrule
    1 & Hungarian Hospital & 261 & 172 & 89 \\
    \midrule
    2 & Switzerland Hospital & 46 & 30 & 16 \\
    \midrule
    3 & Long Beach Hospital & 130 & 85 & 45 \\
    \bottomrule
    \end{tabular}
\end{table}

\subsection{License and Ethics}
This dataset is licensed under a Creative Commons Attribution 4.0
International (CC BY 4.0) license~\citep{janosi1988heart, dua2019Uci}.

Regarding privacy, the dataset authors~\cite{janosi1988heart} indicated that sensitive entries of the dataset (including
names and social security numbers) were removed from the database.

\subsection{Downloading and Preprocessing}
Instructions for downloading are available in the corresponding
README file on FLamby's repository. Dataset is downloaded from 
the UCI Machine Learning repository \citep{dua2019Uci}.

We preprocess the dataset by removing the three features (slope of the 
peak ST segment, number of major vessels, and thalassemia background) 
where too many entries are missing. We then remove records where at 
least one feature is missing. Finally, the two categorical (and non binary)
features (chest pain type and resting electrochardiographic results)
are encoded as binary features using dummy variables.
We also normalize features per center.

\subsection{Task}
The task consists in predicting the presence of a heart disease so the task is binary classification.
\subsection{Baseline, Loss Function, and Evaluation}

\paragraph{Loss function}
For a data record $(x_i, y_i)$, we compute the predicted label
$\hat y_i = \sigma(\beta^T x_i)$, where~$\sigma(z)=1/(1+\exp(-z))$
is the sigmoid function, and $\beta$ the parameters of the model. 
We then compute the loss over the complete dataset as
\begin{align}
    L(y, \hat y) = 
    - \frac{1}{n} \sum_{i=1}^n y_i \log(\hat y_i) 
    + (1 - y_i) \log(1 - \hat y_i)\enspace.
\end{align}

\paragraph{Baseline Model}
We fit a logistic regression model, as this is both a standard problem
in medical research, and the strongest baseline according 
to~\citep{dua2019Uci}. The model trains in a matter of seconds on modern CPUs.

\paragraph{Evaluation}
To evaluate the model, we threshold the predicted values $\hat y$ at
$0.5$, and measure the accuracy of the obtained labels as
\begin{align}
    Acc(\beta, X, y) = \frac{| \{ i \in [n] \mid y_i = (\hat y_i > 0.5) \} |}{n}\enspace.
\end{align}
\paragraph{Optimization parameters}
For the pooled benchmark, we use the Adam optimizer ~\cite{kingma2014adam} with a learning rate of~$0.001$, batch size of $4$, for $50$ epochs.
\paragraph{Hyperparameter Search}

All the federated learning strategies are tested with the SGD optimizer.
We performed a grid search for the federated learning strategies hyperparameters .
For FedAvg and Cyclic, we optimized the learning rate over the values \{0.1, 0.01, 0.001, 0.0001, 0.00001\}.
For FedYogi, FedAdam, FedAdagrad and Scaffold, our search grid space was \{0.1, 0.01, 0.001, 0.0001, 0.00001\} and \{10, 1, 0.1, 0.01, 0.001\} for the learning rate and the server learning rate respectively.
For FedProx, our search space contained \{0.001, 0.0001\} and \{1, 0.1, 0.01\} sets for learning rate and $\mu$ respectively.
The chosen hyperparameters from the HP search can be found in Section~\ref{app:experiments}.

\section{Experimental details}
\label{app:experiments}
\begin{table}[t!]
\centering
\caption{Hyperparameters used for the FedAvg strategy}
\label{app:tab:hyperFedAvg}
\begin{tabular}{ c c c }
\hline
\rowcolor{darkgraylighter}
\multicolumn{3}{c}{FedAvg}
\\
\hline
dataset & \begin{tabular}{@{}c@{}}learning\\ rate\end{tabular} & optimizer \\
\hline
Fed-Camelyon16  & 0.3162 & torch.optim.SGD \\
Fed-LIDC-IDRI  & 0.001 & torch.optim.SGD \\
Fed-IXI  & 0.001 & torch.optim.SGD \\
Fed-TCGA-BRCA  & 0.1 & torch.optim.SGD \\
Fed-KITS19  & 0.03 & torch.optim.SGD \\
Fed-ISIC2019  & 0.01 & torch.optim.SGD \\
Fed-Heart-Disease  & 0.001 & torch.optim.SGD \\
\end{tabular}
\end{table}
\begin{table}[t!]
\centering
\caption{Hyperparameters used for the FedProx strategy}
\label{app:tab:hyperFedProx}
\begin{tabular}{ c c c c }
\hline
\rowcolor{darkgraylighter}
\multicolumn{4}{c}{FedProx}
\\
\hline
dataset  & mu& \begin{tabular}{@{}c@{}}learning\\ rate\end{tabular} & optimizer \\
\hline
Fed-Camelyon16  & 0.316228 & 0.01 & torch.optim.SGD \\
Fed-LIDC-IDRI  & 0.01 & 0.001 & torch.optim.SGD \\
Fed-IXI  & 0.1 & 0.001 & torch.optim.SGD \\
Fed-TCGA-BRCA  & 0.1 & 0.1 & torch.optim.SGD \\
Fed-KITS19  & 0.001 & 0.1 & torch.optim.SGD \\
Fed-ISIC2019  & 0.001 & 0.01 & torch.optim.SGD \\
Fed-Heart-Disease  & 0.001 & 0.01 & torch.optim.SGD \\
\end{tabular}
\end{table}
\begin{table}[t!]
\centering
\caption{Hyperparameters used for the FedAdagrad strategy}
\label{app:tab:hyperFedAdagrad}
\begin{tabular}{ c c c c c c c }
\hline
\rowcolor{darkgraylighter}
\multicolumn{7}{c}{FedAdagrad}
\\
\hline
dataset & \begin{tabular}{@{}c@{}}learning\\ rate\end{tabular} & optimizer& \begin{tabular}{@{}c@{}}learning\\ rate server\end{tabular}& $\beta_1$& $\beta_2$ & $\tau$ \\
\hline
Fed-Camelyon16  & 0.01 & torch.optim.SGD & 0.003162 &   &   &   \\
Fed-LIDC-IDRI  & 0.1 & torch.optim.SGD & 0.1 &   &   &   \\
Fed-IXI  & 1e-04 & torch.optim.SGD & 0.1 & 0.9 & 0.999 & 1e-08 \\
Fed-TCGA-BRCA  & 0.01 & torch.optim.SGD & 1.0 & 0.9 & 0.999 & 1e-08 \\
Fed-KITS19  & 0.1 & torch.optim.SGD & 0.1 & 0.9 & 0.999 & 1e-08 \\
Fed-ISIC2019  & 0.01 & torch.optim.SGD & 0.0316 &   &   &   \\
Fed-Heart-Disease  & 0.003162 & torch.optim.SGD & 0.003162  & 0.9 & 0.999 & 0.3162 \\
\end{tabular}
\end{table}
\begin{table}[t!]
\centering
\caption{Hyperparameters used for the FedAdam strategy}
\label{app:tab:hyperFedAdam}
\begin{tabular}{ c c c c c c c }
\hline
\rowcolor{darkgraylighter}
\multicolumn{7}{c}{FedAdam}
\\
\hline
dataset & \begin{tabular}{@{}c@{}}learning\\ rate\end{tabular} & optimizer& \begin{tabular}{@{}c@{}}learning\\ rate server\end{tabular}& $\beta_1$ & $\beta_2$ & $\tau$ \\
\hline
Fed-Camelyon16  & 0.001 & torch.optim.SGD & 3.1622 &   &   &   \\
Fed-LIDC-IDRI  & 0.3162 & torch.optim.SGD & 0.01 &   &   &   \\
Fed-IXI  & 1e-04 & torch.optim.SGD & 0.1 & 0.9 & 0.999 & 1e-08 \\
Fed-TCGA-BRCA  & 0.01 & torch.optim.SGD & 0.1 & 0.9 & 0.999 & 1e-08 \\
Fed-KITS19  & 0.1 & torch.optim.SGD & 0.01 & 0.9 & 0.999 & 1e-08 \\
Fed-ISIC2019  & 0.01 & torch.optim.SGD & 0.0032 &   &   &   \\
Fed-Heart-Disease  & 0.01 & torch.optim.SGD & 0.01 & 0.9 & 0.999 & 1e-08 \\
\end{tabular}
\end{table}
\begin{table}[t!]
\centering
\caption{Hyperparameters used for the FedYogi strategy}
\label{app:tab:hyperFedYogi}
\begin{tabular}{ c c c c c c c }
\hline
\rowcolor{darkgraylighter}
\multicolumn{7}{c}{FedYogi}
\\
\hline
dataset & \begin{tabular}{@{}c@{}}learning\\ rate\end{tabular} & optimizer& \begin{tabular}{@{}c@{}}learning\\ rate server\end{tabular}& $\beta_1$& $\beta_2$ & $\tau$ \\
\hline
Fed-Camelyon16  & 0.003162 & torch.optim.SGD & 1.0 &   &   &   \\
Fed-LIDC-IDRI  & 0.1 & torch.optim.SGD & 0.001 &   &   &   \\
Fed-IXI  & 1e-04 & torch.optim.SGD & 0.1 & 0.9 & 0.999 & 1e-08 \\
Fed-TCGA-BRCA  & 0.01 & torch.optim.SGD & 0.1 & 0.9 & 0.999 & 1e-08 \\
Fed-KITS19  & 0.1 & torch.optim.SGD & 0.01 & 0.9 & 0.999 & 1e-08 \\
Fed-ISIC2019  & 0.01 & torch.optim.SGD & 0.0032 &   &   &   \\
Fed-Heart-Disease  & 0.0031622 & torch.optim.SGD & 0.01 & 0.9 & 0.999 & 1e-08 \\
\end{tabular}
\end{table}
\begin{table}[t!]
\centering
\caption{Hyperparameters used for the Cyclic strategy}
\label{app:tab:hyperCyclic}
\begin{tabular}{ c c c }
\hline
\rowcolor{darkgraylighter}
\multicolumn{3}{c}{Cyclic}
\\
\hline
dataset & \begin{tabular}{@{}c@{}}learning\\ rate\end{tabular} & optimizer \\
\hline
Fed-Camelyon16  & 0.01 & torch.optim.SGD \\
Fed-LIDC-IDRI  & 0.0316 & torch.optim.SGD \\
Fed-IXI  & 1e-05 & torch.optim.SGD \\
Fed-TCGA-BRCA  & 0.01 & torch.optim.SGD \\
Fed-KITS19  & 0.3 & torch.optim.SGD \\
Fed-ISIC2019  & 0.0032 & torch.optim.SGD \\
Fed-Heart-Disease  & 0.01 & torch.optim.SGD \\
\end{tabular}
\end{table}
\begin{table}[t!]
\centering
\caption{Hyperparameters used for the Scaffold strategy}
\label{app:tab:hyperScaffold}
\begin{tabular}{ c c c c }
\hline
\rowcolor{darkgraylighter}
\multicolumn{4}{c}{Scaffold}
\\
\hline
dataset & \begin{tabular}{@{}c@{}}learning\\ rate\end{tabular} & optimizer& \begin{tabular}{@{}c@{}}learning\\ rate server\end{tabular} \\
\hline
Fed-Camelyon16  & 0.1 & torch.optim.SGD & 3.1622 \\
Fed-LIDC-IDRI  & 0.0316 & torch.optim.SGD & 1.0 \\
Fed-IXI  & 0.001 & torch.optim.SGD & 1.0 \\
Fed-TCGA-BRCA  & 0.01 & torch.optim.SGD & 1.0 \\
Fed-KITS19  & 0.1 & torch.optim.SGD & 1.0 \\
Fed-ISIC2019  & 0.01 & torch.optim.SGD & 1.0 \\
Fed-Heart-Disease  & 0.001 & torch.optim.SGD & 1.0 \\
\end{tabular}
\end{table}
\subsection{Computing resources} \label{app:computing_resources}
Most experiments were performed on virtual machines equipped with NVidia P100 GPUs in Google Cloud to tune local baselines as well as searching hyperparameters.
Additional experiments were also performed on small workstations for the smallest datasets.
Overall, no more than 4k GPU-hours were used throughout the full project.

\subsection{FLamby experimental capabilities}
FLamby is designed to be a lightweight and simple codebase, to enable ease of use.
All clients run sequentially in the same python environment, without multithreading.
Datasets are assigned to clients as different python objects.
GPU acceleration is supported thanks to current PyTorch~\cite{paszke2019pytorch} backend.
In order to perform more realistic experiments, e.g. to investigate communication constraints, we encourage the usage of dedicated FL libraries, which are easy to integrate with FLamby.

\subsection{Benchmark hyperparameters \label{sec:hp}}
We used the hyperparameters detailed in Tables~\ref{app:tab:hyperFedAvg} to~\ref{app:tab:hyperScaffold} to obtain the results of Figure~\ref{fig:benchmark_results}.
These hyperparameters were found after hyper-optimization on a coarse grid.
For all strategies and datasets, we set $E=100$ the number of local updates.

\subsection{Run details}
Results of Figure~\ref{fig:benchmark_results} were obtained following 5 independent runs with different random seeds, except for the largest one (Fed-LIDC-IDRI), where computational resources prevented training.

\section{Synthetic dataset splits}
One of FLamby's strengths is that it provides datasets with natural splits.
However, due to its focus on healthcare applications, the number of clients is limited.
Thanks to the standardized API of the datasets, it is possible to create new client splits based on the provided codebase.

We provide an example of such a synthetic sampling based on a Dirichlet distribution on the original clients.
If $K$ denotes the previous number of clients and $K'$ the desired number of clients, for $\alpha \in (0, 1)$, we draw a probability distribution $\mathbf{p}_k \in \mathbb{R}^{K'}$ as
\begin{equation}
\mathbf{p}_k \sim \mathrm{Dir}(\alpha), \text{such that} \sum_{k'} p_{k k'} = 1.
\end{equation}
Each sample from client $k$ is then attributed to client $k'$ with probability $p_{kk'}$, both for the train and test sets.
The closer to $0$ $\alpha$ gets, the sharper the distribution probability $\mathbf{p}_k$ gets. In order to avoid having empty clients with the synthetic split, we recommend setting $\alpha \geq 1/2$, following previous works~\cite{yurochkin2019bayesian}.

\section{Examples of extensions possible in FLamby}
In this Appendix, we showcase the extensibility of FLamby by tackling different FL settings.
\subsection{Differential Privacy Example}
\label{app:diff-privacy}
Differential privacy (DP)~\cite{dwork2014algorithmic} is an important approach to protect update exchanges between Federated Learning participants against malicious privacy attacks~\cite{wei2020framework}. In this section, we use Fed-Heart-Disease to demonstrate the use of FLamby to study $(\epsilon, \delta)$-DP federated learning~\cite{dwork2014algorithmic}.

\begin{figure}[h!]
    \centering
    \includegraphics[width=0.7\linewidth]{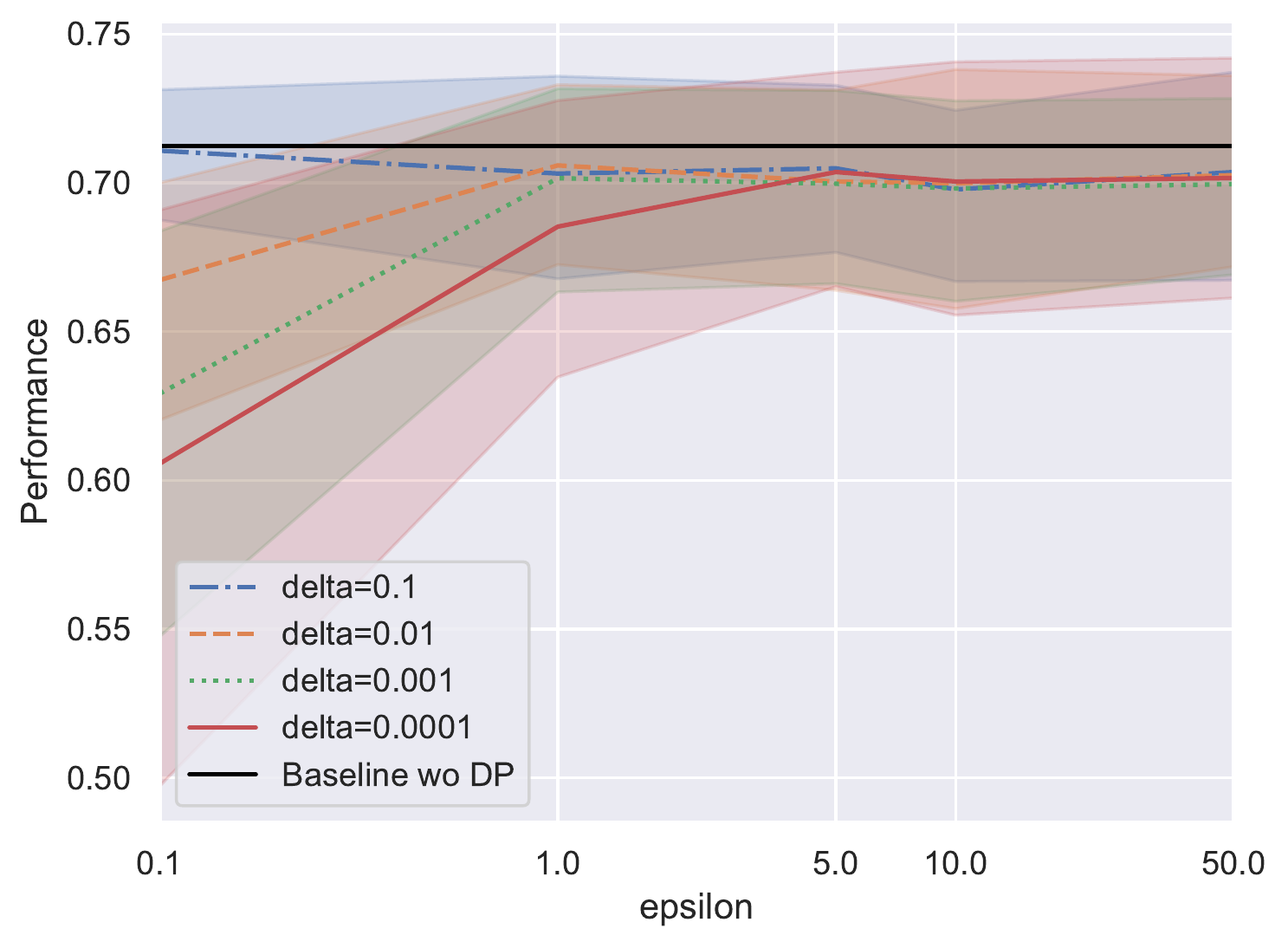}
    \caption{Impact of Differential-Privacy on average performance for DP-FedAvg on Fed-Heart-Disease. \label{fig:dp}}
\end{figure}

Figure~\ref{fig:dp} displays the average performance of a machine learning model trained in a differentially private fashion with DP-FedAvg as a function of $\epsilon$ and $\delta$.
We compare it to a regular training using the same model initialization but without privacy (``Baseline wo DP''), trained with regular FedAvg.
We see the performance diminishing when $\epsilon$ tends to $0$, especially for small values of $\delta$ often used in practice, which is a standard phenomenon.

To implement DP in FedAvg, we use on the DP-SGD mechanism and track the monitoring of privacy budget thanks to the moment accountant~\cite{abadi2016deep}. We use the Opacus library~\cite{yousefpour2021opacus}, which is easy to integrate into FLamby thanks to its modular design.

At this time of writing, Opacus does not support all Deep Learning building blocks such as normalization layers. This prevents applying DP mechanisms on some some of FLamby's baseline models such as the baseline for Fed-ISIC2019 and Fed-IXI.

\subsection{Personalized Federated Learning}
\label{app:personalized_fl}
Model personalization~\cite{fallah2020personalized} is an effective strategy to improve model performance in cross-silo settings, especially in presence of data heterogeneity.
Here, we showcase a simple example of model personalization with FLamby, which is possible thanks to its simple and modular API.

We implement the FedAvg strategy followed by local fine tuning on each center, thus producing as many models as there are clients. We test the addition of such fine-tuning process on the performances of Federated models while testing each model on its corresponding test set. For each dataset, we perform $100$ local updates after the federated averaging training has taken place.

\begin{figure}[h!]
    \centering
    \includegraphics[width=0.7\linewidth]{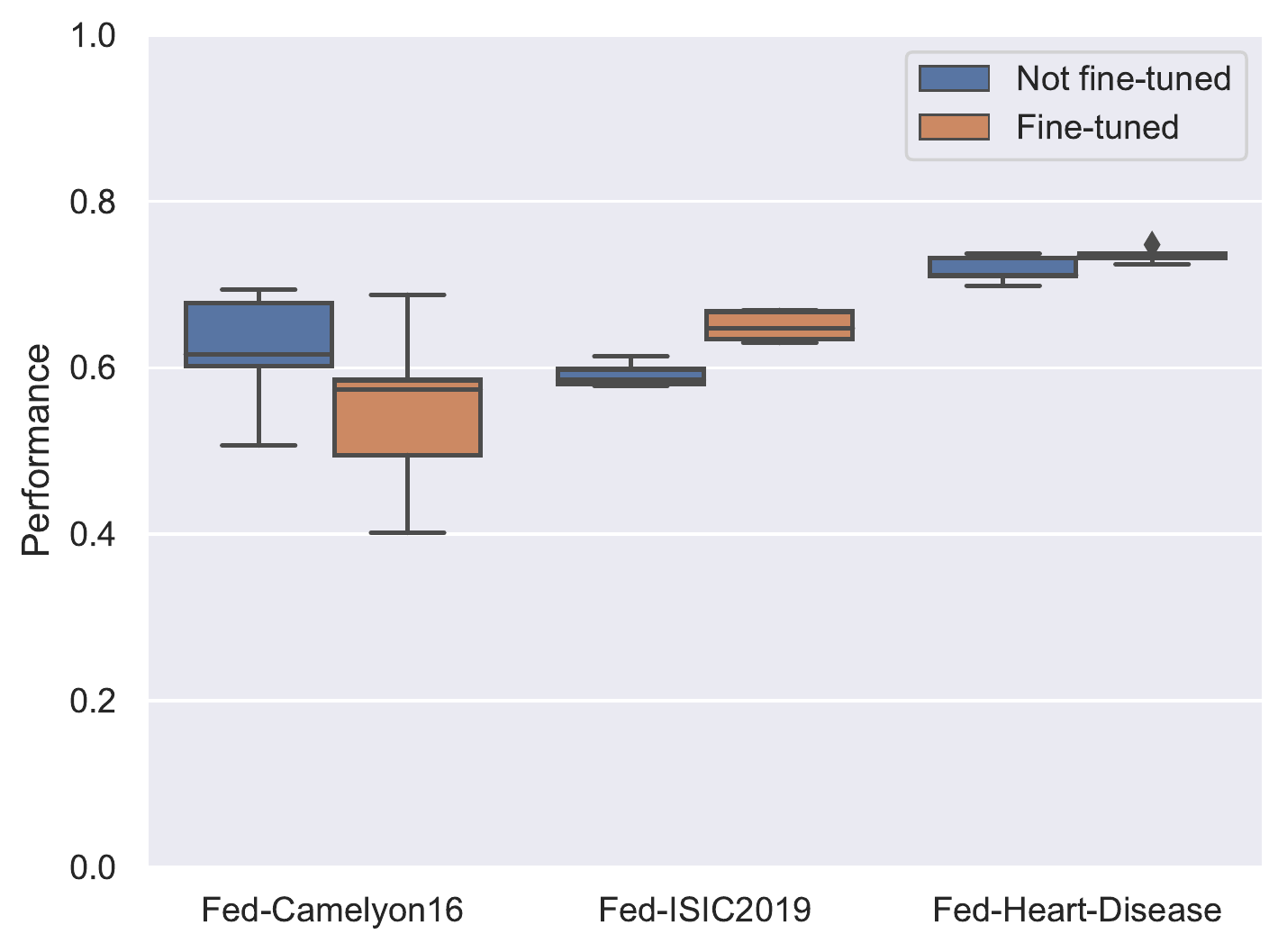}
    \caption{Impact of personalization on test average performance on three datasets of the suite (Fed-Heart-Disease, Fed-Camelyon16 and Fed-ISIC2019) after performing Federated Averaging. In the two extreme cases (Left and Right) fine-tuning is beneficial, whereas on Fed-Camelyon16, Fine-tuning degrades the performance of the resulting models. We hypothesize that, in this case, fine-tuning overfits the local training datasets.\label{fig:perso}}
\end{figure}

Figure display the training results~\ref{fig:perso}.
We see that for Fed-Heart-Disease and Fed-ISIC2019, personalization improves results, while performance is slightly degraded for Fed-Camelyon16.
We hope that researchers will be able to investigate more personalization strategies easily with FLamby.
\section{Quantitative heterogeneity benchmarks}
\label{app:heterogeneity}

We describe in this section our analysis of the heterogeneity of the Flamby datasets. Due to the variety of tasks and data, we restrict this study to generic metrics. We always compute them on the whole dataset, putting together test and train data to have the best possible estimation of the underlying distribution.

In the first subsection, we briefly describe the three sources of heterogeneity that we consider. Next, we detail the methodology used to compute statistical distance between clients. Thirdly, we apply this methodology to the FLamby's dataset. And finally, we provide some discussion on these results.

\subsection{Description of measured heterogeneity}

\paragraph{Imbalance.} The easiest quantification of heterogeneity comes from the number of samples hosted by each client, which gives natural unbalance in the training because small clients are likely to be either over-fitted or neglected in the final model. 

\paragraph{Labels distribution.} Labels can be another source of heterogeneity in case when prediction outputs vary between clients for the considered task. For instance, if a client is specialized in treating the patients with the given disease, the labels are likely to be biased (with a high number of persons with this disease), even if all clients have patients from a similar population.
In the case of the toy example of MNIST, a split with this heterogeneity is to have clients specialized on a single digit.

\paragraph{Features distribution.} Finally, in case when the features are the origin of heterogeneity, the underlying sample distribution is different in each client. It means that the same outputs can be characterized by different data depending on the client. This heterogeneity can arise from having different measurement tools (as is the case in Fed-LIDC-IDRI dataset), or different population in each client (e.g. Fed-TCGA-BRCA dataset). In the case of the toy example of MNIST, a split with this heterogeneity is to have a client where digits are in italics.

\subsection{Methodology}
For the sample division, we report the number of clients and the splits. As a way to summarize heterogeneity, we compute the entropy of the distribution of the samples across clients.
\begin{equation}
\label{app:eq:entropy}
    H(X) = - \sum_{k=1}^{K} \frac{n_k}{N} \log_2 \frac{n_k}{N}
\end{equation}
where $K$ is the number of clients, $N$ the total number of samples and $n_k$ the number of samples belonging to client $k$.

For label and features heterogeneity, when initial dimension is larger than $16$, we reduce the dimension by using PCA trained on all the centralized samples. Then, for each client, we compute the Wasserstein distance (see~Definition \ref{def:wassdist}) between each client's distribution, or the total variation distance (see~Definition \ref{def:tv}) for discrete data. 
The Wasserstein distances are computed using a minibatch-Wasserstein (without regularization)~\citep[see][]{fatras_learning_2020} implemented in the POT library~\cite{flamary2021pot} and is defined below:

\begin{definition}[Wasserstein distance]
\label{def:wassdist}
For all probability measures $\alpha$ and $\beta$ on $\mathcal{B}(\mathbb{R}^d)$, 
such that $\int_{\mathbb{R}^d} \| w\|^2 \mathrm{d} \alpha (w) < +\infty$ and $\int_{\mathbb{R}^d} \| w\|^2 \mathrm{d} \beta (w) \leq +\infty $, define the squared Wasserstein distance of order $2$ between $\alpha$ and $\beta$ by 
\begin{equation}
  \mathcal W^2_2(\alpha, \beta):= \inf
_{\xi \in  \Gamma(\alpha , \beta)} \int \| x-y\|^{2 } \xi(dx,
  dy) , \label{eq:defwass}
\end{equation}  
where $ \Gamma(\alpha , \beta)$ is the set of
probability measures $\xi$ on $\mathcal{B}(\mathbb{R}^{d} \times \mathbb{R}^{d})$ satisfying for all $\mathsf{A} \in \mathcal{B}(\mathbb{R}^d)$, $\xi(\mathsf{A} \times \mathbb{R}^d) = \beta(\mathsf{A})$, $\xi( \mathbb{R}^d \times \mathsf{A} ) = \alpha(\mathsf{A})$. 
\end{definition}

The Total-Variation distance used for discrete labels is defined as following:
\begin{definition}[Total-Variation]
\label{def:tv}
For any vector of probability $\alpha, \beta$ in $[0,1]^d$, the $\mathrm{TV}$-value is defined~by $$\mathrm{TV}(\alpha, \beta) = \frac{1}{2} \sum_{i=1}^d | \alpha_i - \beta_i| \in [0,1].$$
\end{definition}

As the tasks, dimension and characteristics differs for each dataset, there are two directions to interpret results: 1) comparing heterogeneity between clients, and 2) by contrast with a synthetic scenario where data would be identically distributed among clients. Thus, we compute two pairwise-distances matrices: one with the natural split, and one with data distributed uniformly on clients (the size of the dataset on each client is identical to the natural split). The latter is built to simulate the i.i.d.~setting and to compare the natural split with the case where we would have had homogeneous clients.

Next, we rescale the pairwise-distances matrices in order to have standardized variables in the synthetic case. Formally, we note $\mathcal{D}_{\mathrm{i.i.d.}}$ (resp. $\mathcal{D}_{\mathrm{natural}}$) the set of distances for the synthetic i.i.d.~split (resp. natural split). The cardinal of these two sets is $n(n-1) / 2$ because 1) the diagonal (distance of a client with itself) must be zero, 2) the Wasserstein distance is symmetric. Rescaling the matrices means that we standardize the i.i.d.~set by removing the mean and scaling to unit variance i.e. computing $(\mathcal{D}_{\mathrm{i.i.d.}} - \mu_{\mathrm{i.i.d.}}) / \sigma_{\mathrm{i.i.d.}}$, where $\mu_{\mathrm{i.i.d.}}, \sigma_{\mathrm{i.i.d.}}^2$ are the mean and the variance of the i.i.d.~set. Then, we apply the same transformation on the set of distance computed on the natural split i.e. $(\mathcal{D}_{\mathrm{natural}} - \mu_{\mathrm{i.i.d.}}) / \sigma_{\mathrm{i.i.d.}}$.

This is motivated by the fact that in the homogeneous case, we expect the distances to be zero. It follows that after rescaling, we are able to compare the values within the pairwise-distance matrix of the natural split. Thus, we are able to identify which clients are the closest or at odds with the others.
Additionally, the magnitudes after rescaling give an indication on the degree of heterogeneity within the dataset. The bigger their magnitudes are after rescaling, the more distant is the natural split to what would be a homogeneous split.

\subsection{Datasets analysis}

For each dataset, we report the rescaled pairwise-distances matrix for features and for labels, with the associated i.i.d. baseline in Figure~\ref{fig:allmatrices}. In order to highlight on the same plot the clients size imbalance and the clients' heterogeneity, the width of each column in the pairwise distance matrix is proportional to the client size. This helps to put heterogeneity in perspective with the size of the client and to have a better understanding of what is, in practice, the weight of the client's heterogeneity.

We also report the mean and maximum value for each dataset, both for features and labels, alongside sample distribution among client entropy in Table~\ref{app:tab:heter_figures}.

\begin{table}
\centering
\caption{Mean and max distances for features and labels, and entropy computed using eq.~\ref{app:eq:entropy}. High values correspond to important heterogeneity. \\ ~}
\label{app:tab:heter_figures}
\begin{tabular}{lllllll}
\toprule
& camelyon16 & ixi & tcga brca & kits19 & isic2019 & heart disease \\
\midrule
X mean  & 710618.57 & 188.52 &  6.38 & -0.31 &  1.62 &  7.05 \\
X max   & 710618.57 & 289.14 & 16.74 &  1.42 &  7.29 & 10.59 \\
Y mean  &     -0.01 &   3.52 & 22.26 &  0.12 & 29.40 &  2.90 \\
Y max   &     -0.01 &   6.95 & 79.42 &  2.68 & 53.87 &  5.77 \\
Entropy &      0.97 &   1.38 &  2.44 &  2.49 &  1.93 &  1.75 \\
\bottomrule
\end{tabular}
\end{table}

\newcommand{\addFeatures}[1]{\includegraphics[align=c,width=0.45\textwidth]{figures/heterogeneity/#1-X-eps-converted-to.pdf}}
\newcommand{\addLabels}[1]{\includegraphics[align=c,width=0.45\textwidth]{figures/heterogeneity/#1-Y-eps-converted-to.pdf}}
\newcolumntype{C}{>{\centering\arraybackslash}m{0.45\textwidth}}

\newcommand{\splitCell}[2]{\hspace{-2.5em} #1 \qquad #2}

\begin{table}
\caption{Heterogeneity of Flamby datasets. Each matrix is the pairwise distance matrix, the width of their column corresponds to the number of sample. \\~}
\label{fig:allmatrices}
\begin{tabular}{m{1em}cc}
\toprule
 & Distance on features & Distance on labels \\ 
  & \splitCell{(synth.) i.i.d. split}{natural split} & \splitCell{(synth.) i.i.d. split}{natural split} \\ 
\midrule
\rotatebox{90}{camelyon16} & \addFeatures{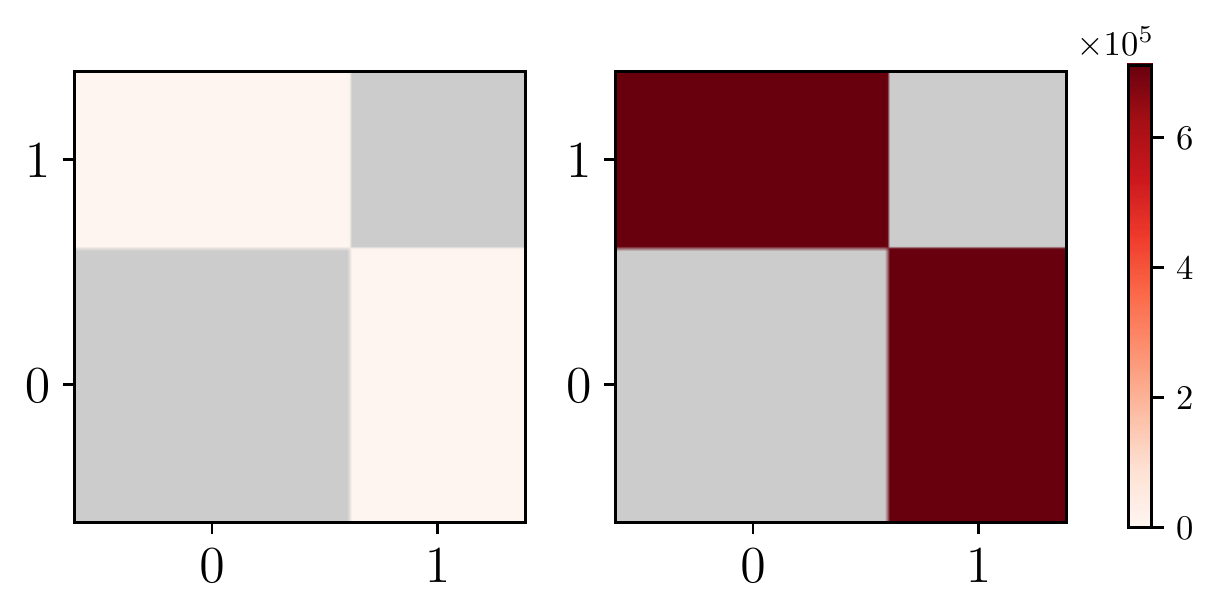} & \addLabels{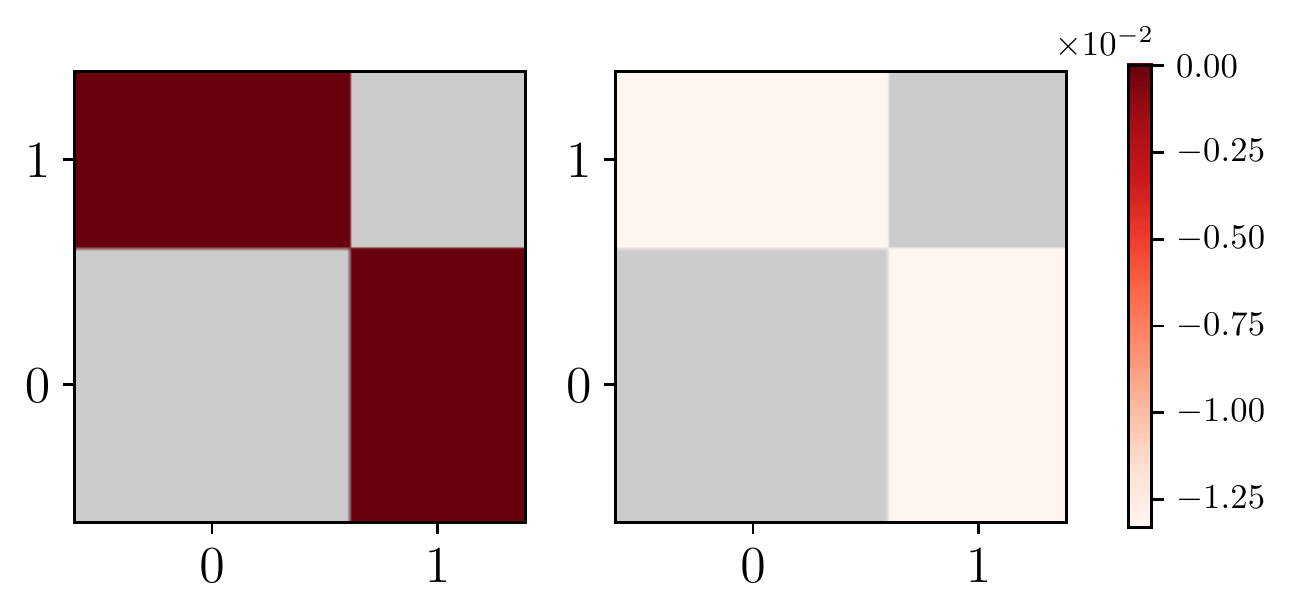} \\ 
\rotatebox{90}{ixi} & \addFeatures{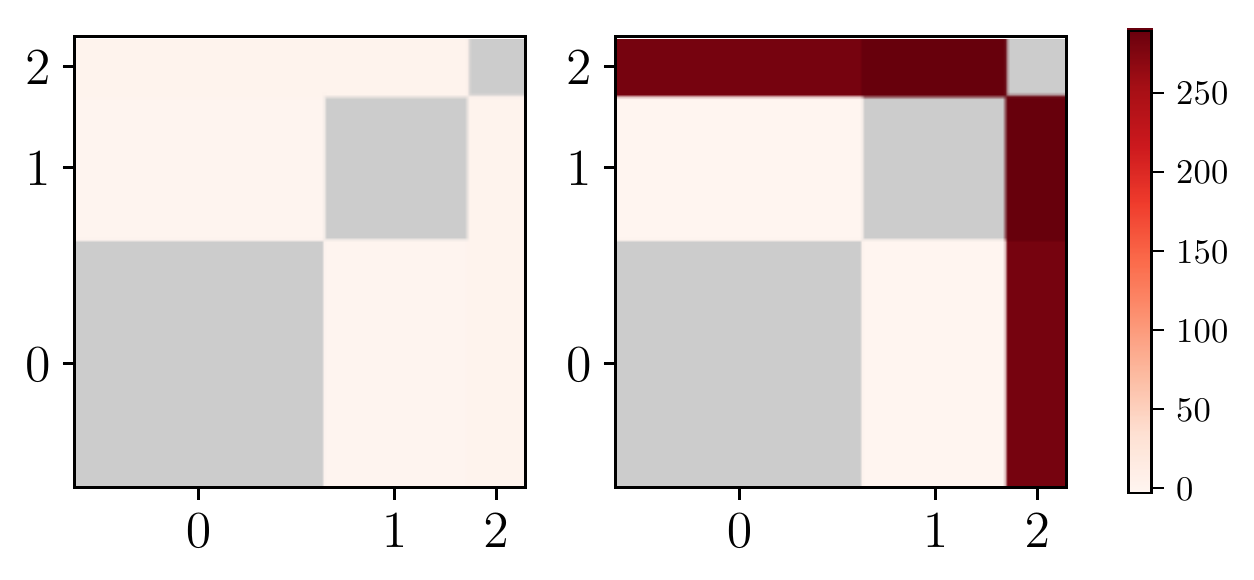} &  \addLabels{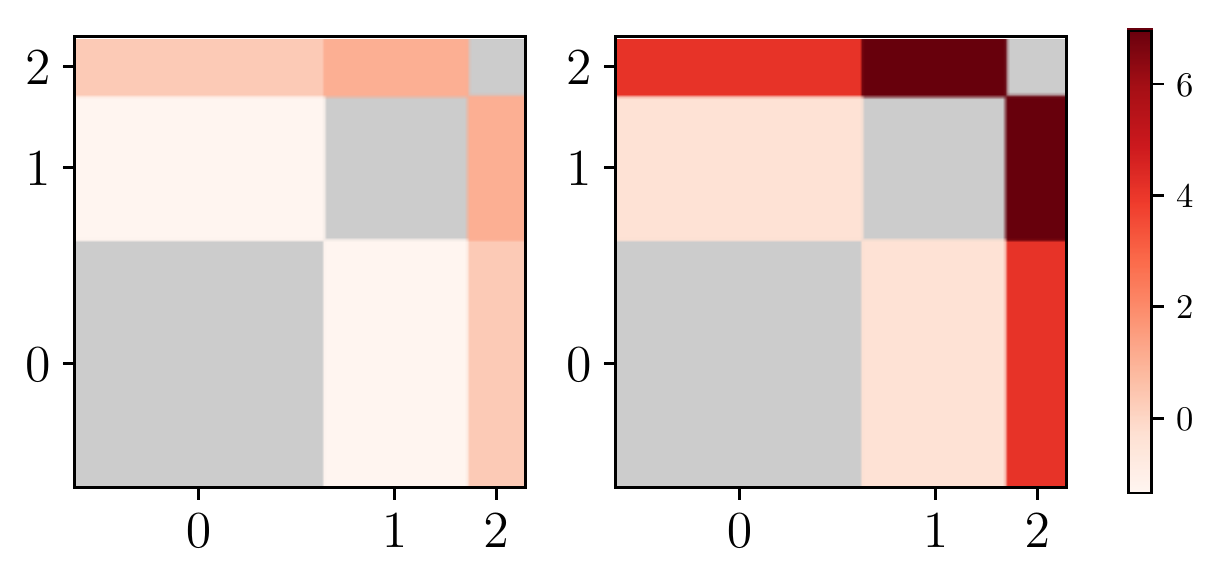} \\ 
\rotatebox{90}{tcga brca} & \addFeatures{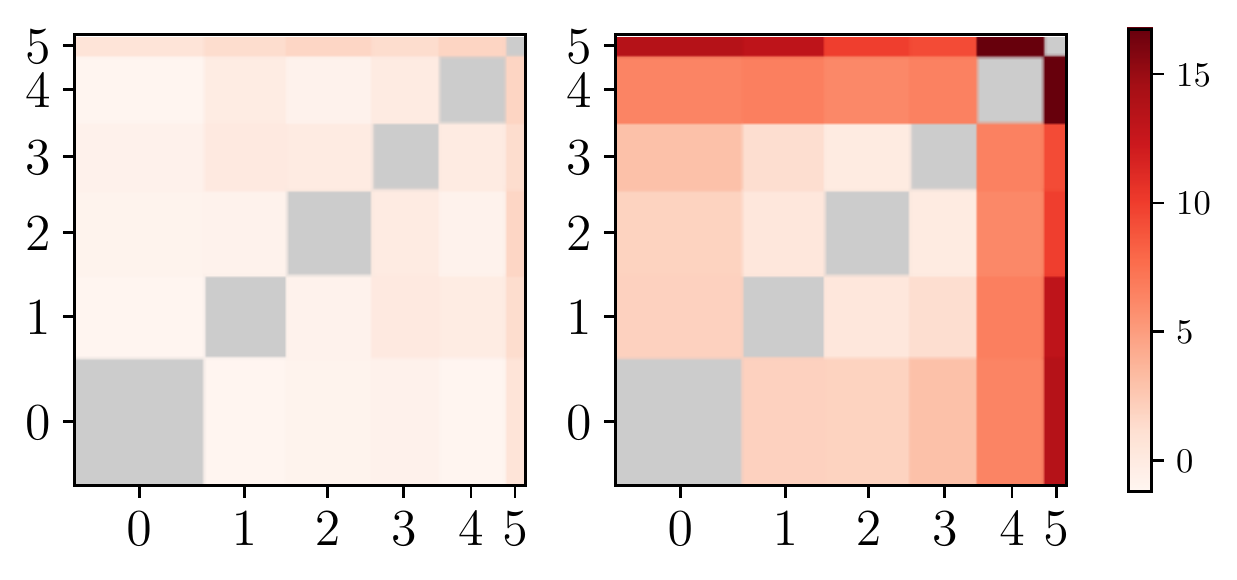} &  \addLabels{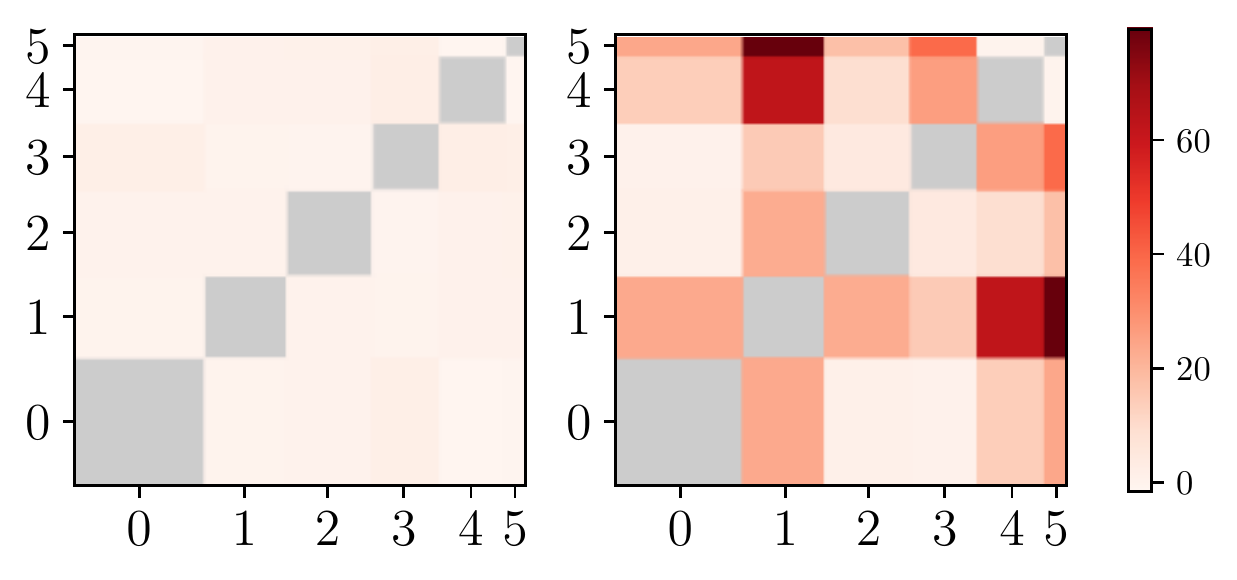} \\ \rotatebox{90}{kits19} & \addFeatures{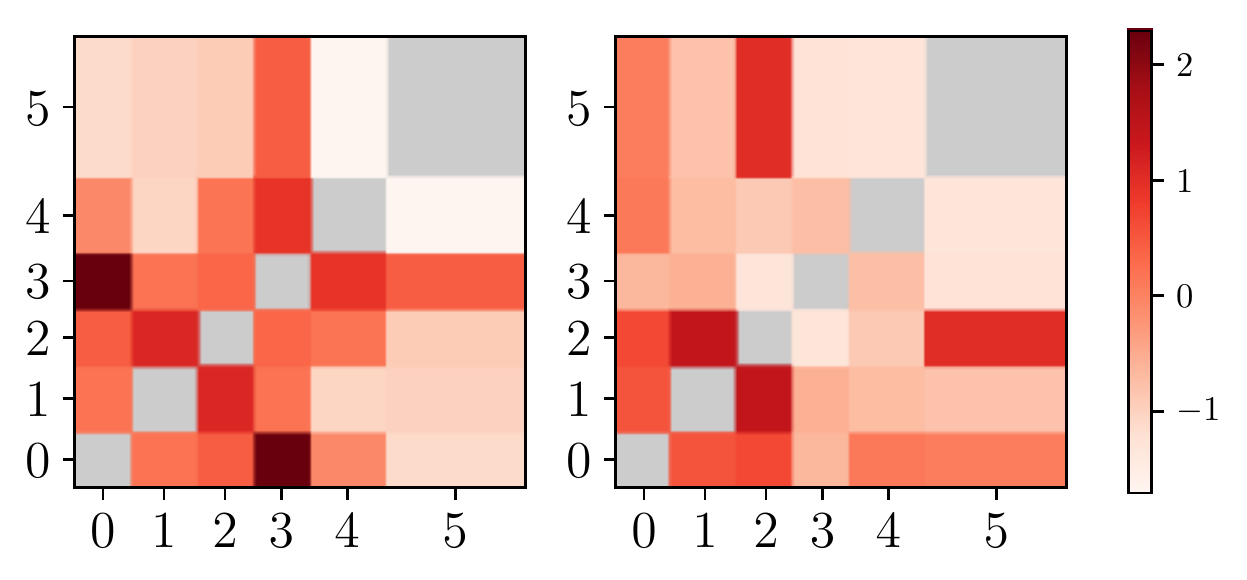} &  \addLabels{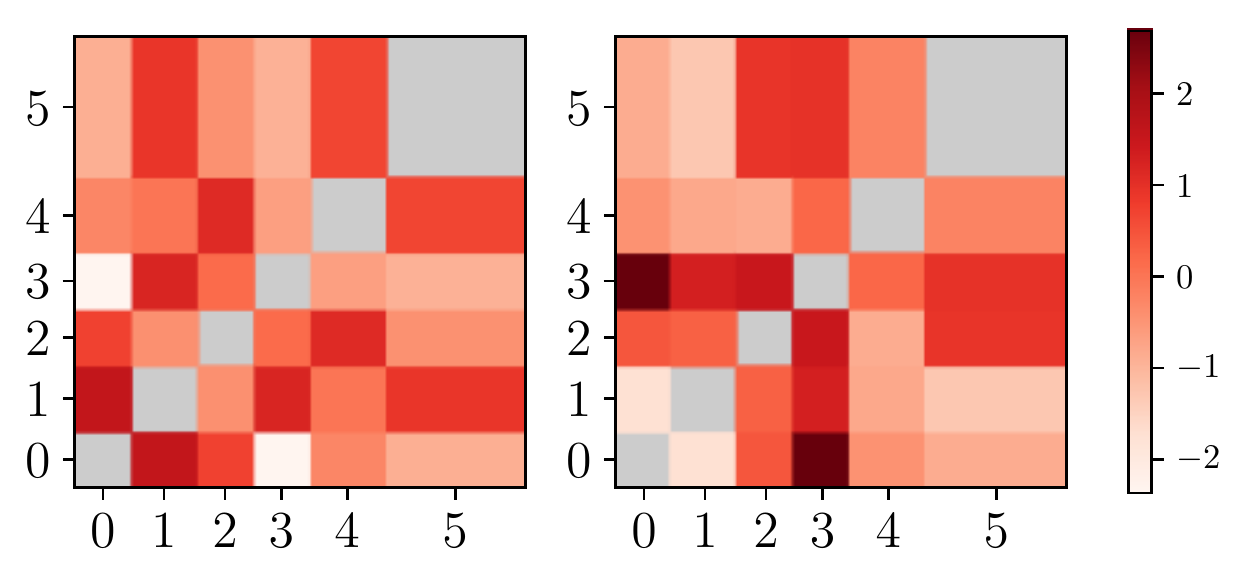} \\ 
\rotatebox{90}{isic2019} & \addFeatures{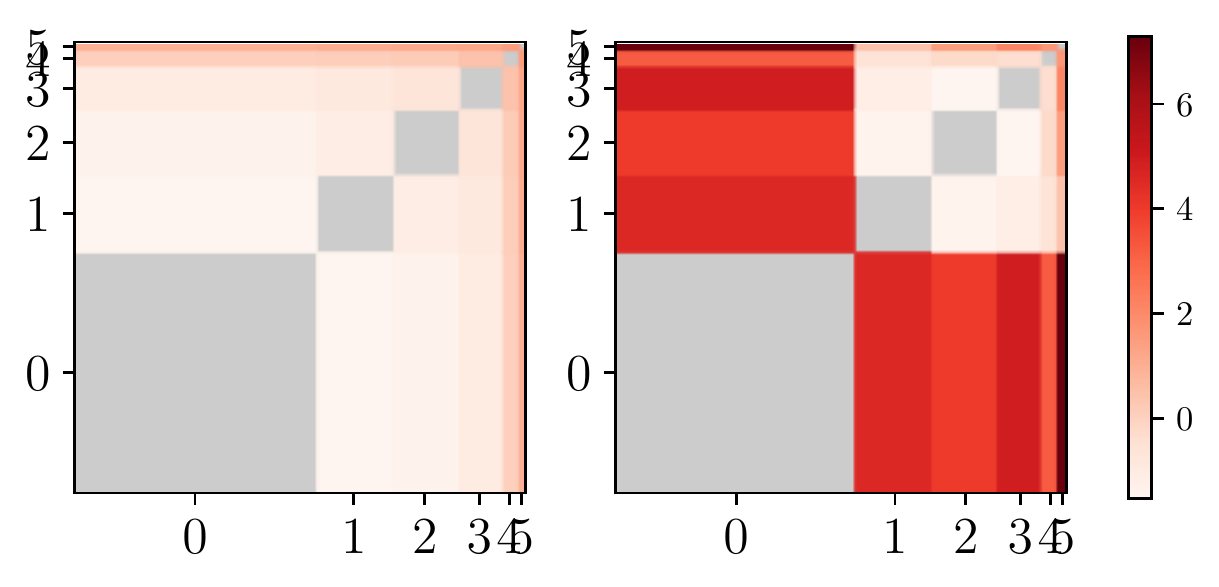} &  \addLabels{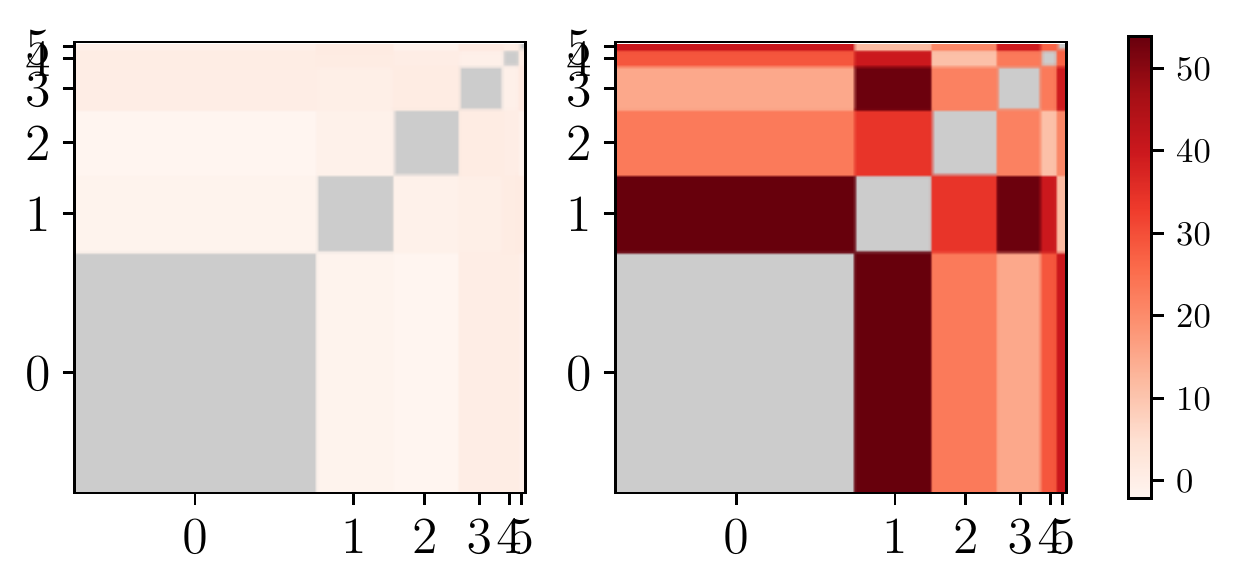} \\ 
\rotatebox{90}{heart disease} & \addFeatures{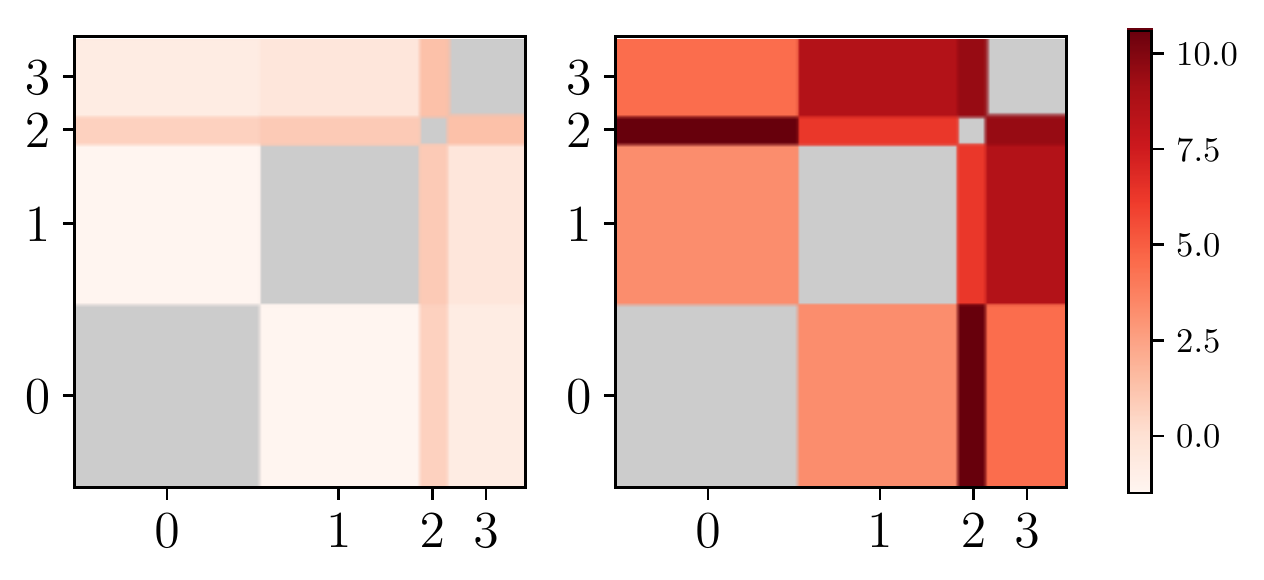} &  \addLabels{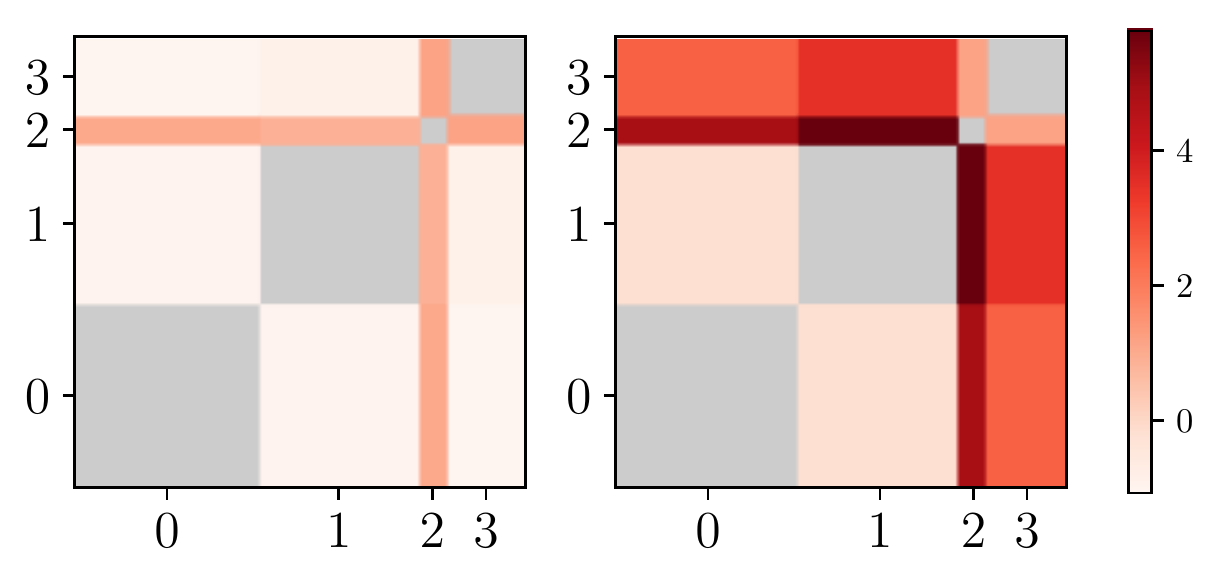} \\ 
\bottomrule 
\end{tabular}
\end{table} 	

We can make the following observations for each dataset by analysing Tables~\ref{app:tab:heter_figures} and \ref{fig:allmatrices}.
\begin{itemize}
    \item \textbf{Camelyon16.} There is a high heterogeneity on features, however labels are i.i.d.. This is of particular interest as it means that different features have led to close labels.
    \item \textbf{IXI.} Client 0 and 1 are very close, both in terms of features and labels. Compared to them, client 3 (which is also the smallest) is an outsider.
    \item \textbf{TCGA-BRCA.} In terms of features, clients 0, 1, 2, 3 are relatively close. But client 1 has labels different from the three other clients. Client 4 and 5 are apart, but it is interesting to notice that while their features are extremely different, their labels are almost identical. As for Camelyon16, this is of particular interest.
    \item \textbf{Kits19.} Clients are completely homogeneous for both features and labels. This can be derived from the fact that the distances are the same for both the natural split and the i.i.d.~split. This was already suggested by Figure~\ref{subfig:heterogeneity_kits}. 
    \item \textbf{Isic2019.} Features of client 0 are very different from the other clients. The four last clients have close features because their distances are almost zero. On the contrary, their distances on labels are far from zero. It means that close features have led to different labels. This is an element of particular interest. 
    \item \textbf{Heart disease.} The heterogeneity on features and on labels are of the same order of magnitude (up to a factor $2$) and not very important (maximum is at $10$ for features, at $5$ for labels). Client 2 is the smallest client and is an outsider. This is logical knowing that client 2 is a hospital specialized in major heart disease. We can also notice that client 1 and 2 have a moderate distance in terms of features. But however, based on their labels, they have the most significant distance. It means that relatively close features have led to very different labels. Like for isic2019, this is an element of particular interest. This could have happened if patients in hospital 2 have more severe heart disease than in other hospitals, but still have disease features close to classical cases.
    
\end{itemize}

\subsection{Discussion}

Measuring heterogeneity is an open question in machine learning, and it is beyond the scope of this paper. We provide some measurements as an indicative benchmark, with a methodology easy to reproduce. Other kind of heterogeneity could be computed and might lead to different conclusion on which clients are less or more similar, in particular as PCA representation can lead to a significant data loss. The advantages of this benchmark is its generality that allows to tackle the various data type and tasks found in FLamby. \FloatBarrier
\putbib[refs_app]
\end{bibunit}

\begin{thebibliography}{100}

\bibitem{tensorflow2015-whitepaper}
Mart\'{i}n Abadi, Ashish Agarwal, Paul Barham, Eugene Brevdo, Zhifeng Chen,
  Craig Citro, Greg~S. Corrado, Andy Davis, Jeffrey Dean, Matthieu Devin,
  Sanjay Ghemawat, Ian Goodfellow, Andrew Harp, Geoffrey Irving, Michael Isard,
  Yangqing Jia, Rafal Jozefowicz, Lukasz Kaiser, Manjunath Kudlur, Josh
  Levenberg, Dandelion Man\'{e}, Rajat Monga, Sherry Moore, Derek Murray, Chris
  Olah, Mike Schuster, Jonathon Shlens, Benoit Steiner, Ilya Sutskever, Kunal
  Talwar, Paul Tucker, Vincent Vanhoucke, Vijay Vasudevan, Fernanda Vi\'{e}gas,
  Oriol Vinyals, Pete Warden, Martin Wattenberg, Martin Wicke, Yuan Yu, and
  Xiaoqiang Zheng.
\newblock {TensorFlow}: Large-scale machine learning on heterogeneous systems,
  2015.
\newblock Software available from tensorflow.org.

\bibitem{andreux2020siloed}
Mathieu Andreux, Jean~Ogier du~Terrail, Constance Beguier, and Eric~W Tramel.
\newblock Siloed federated learning for multi-centric histopathology datasets.
\newblock In {\em Domain Adaptation and Representation Transfer, and
  Distributed and Collaborative Learning}, pages 129--139. Springer, 2020.

\bibitem{andreux2020federated}
Mathieu Andreux, Andre Manoel, Romuald Menuet, Charlie Saillard, and Chlo{\'e}
  Simpson.
\newblock Federated survival analysis with discrete-time {Cox} models.
\newblock {\em arXiv preprint arXiv:2006.08997}, 2020.

\bibitem{anguita2013public}
Davide Anguita, Alessandro Ghio, Luca Oneto, Xavier Parra~Perez, and Jorge~Luis
  Reyes~Ortiz.
\newblock A public domain dataset for human activity recognition using
  smartphones.
\newblock In {\em Proceedings of the 21th international European symposium on
  artificial neural networks, computational intelligence and machine learning},
  pages 437--442, 2013.

\bibitem{armato2011lidc}
Samuel~G Armato~III, Geoffrey McLennan, Luc Bidaut, Michael~F McNitt-Gray,
  Charles~R Meyer, Anthony~P Reeves, Binsheng Zhao, Denise~R Aberle, Claudia~I
  Henschke, Eric~A Hoffman, et~al.
\newblock The lung image database consortium ({LIDC}) and image database
  resource initiative ({IDRI}): a completed reference database of lung nodules
  on ct scans.
\newblock {\em Medical physics}, 38(2):915--931, 2011.

\bibitem{aman}
Aman Arora.
\newblock Siim-isic melanoma classification - my journey to a top 5\% solution
  and first silver medal on kaggle.
\newblock \url{https://amaarora.github.io/2020/08/23/siimisic.html}.
\newblock Accessed: 2022-02-02.

\bibitem{badano2015consistency}
Aldo Badano, Craig Revie, Andrew Casertano, Wei-Chung Cheng, Phil Green, Tom
  Kimpe, Elizabeth Krupinski, Christye Sisson, Stein Skr{\o}vseth, Darren
  Treanor, et~al.
\newblock Consistency and standardization of color in medical imaging: a
  consensus report.
\newblock {\em Journal of digital imaging}, 28(1):41--52, 2015.

\bibitem{baheti2020federated}
Pragati Baheti, Mukul Sikka, KV~Arya, and R~Rajesh.
\newblock Federated learning on distributed medical records for detection of
  lung nodules.
\newblock In {\em VISIGRAPP (4: VISAPP)}, pages 445--451, 2020.

\bibitem{bandi2018detection}
Peter Bandi, Oscar Geessink, Quirine Manson, Marcory Van~Dijk, Maschenka
  Balkenhol, Meyke Hermsen, Babak~Ehteshami Bejnordi, Byungjae Lee, Kyunghyun
  Paeng, Aoxiao Zhong, et~al.
\newblock From detection of individual metastases to classification of lymph
  node status at the patient level: the camelyon17 challenge.
\newblock {\em IEEE transactions on medical imaging}, 38(2):550--560, 2018.

\bibitem{bejnordi2017diagnostic}
Babak~Ehteshami Bejnordi, Mitko Veta, Paul~Johannes Van~Diest, Bram
  Van~Ginneken, Nico Karssemeijer, Geert Litjens, Jeroen~AWM Van Der~Laak,
  Meyke Hermsen, Quirine~F Manson, Maschenka Balkenhol, et~al.
\newblock Diagnostic assessment of deep learning algorithms for detection of
  lymph node metastases in women with breast cancer.
\newblock {\em Jama}, 318(22):2199--2210, 2017.

\bibitem{bhowmick2018protection}
Abhishek Bhowmick, John Duchi, Julien Freudiger, Gaurav Kapoor, and Ryan
  Rogers.
\newblock Protection against reconstruction and its applications in private
  federated learning.
\newblock {\em arXiv preprint arXiv:1812.00984}, 2018.

\bibitem{bonawitz2019towards}
Keith Bonawitz, Hubert Eichner, Wolfgang Grieskamp, Dzmitry Huba, Alex
  Ingerman, Vladimir Ivanov, Chloe Kiddon, Jakub Kone{\v{c}}n{\`y}, Stefano
  Mazzocchi, Brendan McMahan, et~al.
\newblock Towards federated learning at scale: System design.
\newblock {\em Proceedings of Machine Learning and Systems}, 1:374--388, 2019.

\bibitem{bonawitz2017practical}
Keith Bonawitz, Vladimir Ivanov, Ben Kreuter, Antonio Marcedone, H~Brendan
  McMahan, Sarvar Patel, Daniel Ramage, Aaron Segal, and Karn Seth.
\newblock Practical secure aggregation for privacy-preserving machine learning.
\newblock In {\em proceedings of the 2017 ACM SIGSAC Conference on Computer and
  Communications Security}, pages 1175--1191, 2017.

\bibitem{jax2018github}
James Bradbury, Roy Frostig, Peter Hawkins, Matthew~James Johnson, Chris Leary,
  Dougal Maclaurin, George Necula, Adam Paszke, Jake Vander{P}las, Skye
  Wanderman-{M}ilne, and Qiao Zhang.
\newblock {JAX}: composable transformations of {P}ython+{N}um{P}y programs,
  2018.

\bibitem{burki2019pharma}
Talha Burki.
\newblock Pharma blockchains {AI} for drug development.
\newblock {\em The Lancet}, 393(10189):2382, 2019.

\bibitem{caldas2018leaf}
Sebastian Caldas, Sai Meher~Karthik Duddu, Peter Wu, Tian Li, Jakub
  Kone{\v{c}}n{\`y}, H~Brendan McMahan, Virginia Smith, and Ameet Talwalkar.
\newblock Leaf: A benchmark for federated settings.
\newblock {\em arXiv preprint arXiv:1812.01097}, 2018.

\bibitem{caruana2004kdd}
Rich Caruana, Thorsten Joachims, and Lars Backstrom.
\newblock {KDD-Cup} 2004: results and analysis.
\newblock {\em ACM SIGKDD Explorations Newsletter}, 6(2):95--108, 2004.

\bibitem{chakravarty2021federated}
Arunava Chakravarty, Avik Kar, Ramanathan Sethuraman, and Debdoot Sheet.
\newblock Federated learning for site aware chest radiograph screening.
\newblock In {\em 2021 IEEE 18th International Symposium on Biomedical Imaging
  (ISBI)}, pages 1077--1081. IEEE, 2021.

\bibitem{chang2018distributed}
Ken Chang, Niranjan Balachandar, Carson Lam, Darvin Yi, James Brown, Andrew
  Beers, Bruce Rosen, Daniel~L Rubin, and Jayashree Kalpathy-Cramer.
\newblock Distributed deep learning networks among institutions for medical
  imaging.
\newblock {\em Journal of the American Medical Informatics Association},
  25(8):945--954, 2018.

\bibitem{charles2021large}
Zachary Charles, Zachary Garrett, Zhouyuan Huo, Sergei Shmulyian, and Virginia
  Smith.
\newblock On large-cohort training for federated learning.
\newblock {\em arXiv preprint arXiv:2106.07820}, 2021.

\bibitem{chowdhery2022palm}
Aakanksha Chowdhery, Sharan Narang, Jacob Devlin, Maarten Bosma, Gaurav Mishra,
  Adam Roberts, Paul Barham, Hyung~Won Chung, Charles Sutton, Sebastian
  Gehrmann, et~al.
\newblock Palm: Scaling language modeling with pathways.
\newblock {\em arXiv preprint arXiv:2204.02311}, 2022.

\bibitem{cciccek20163d}
{\"O}zg{\"u}n {\c{C}}i{\c{c}}ek, Ahmed Abdulkadir, Soeren~S Lienkamp, Thomas
  Brox, and Olaf Ronneberger.
\newblock 3d u-net: learning dense volumetric segmentation from sparse
  annotation.
\newblock In {\em International conference on medical image computing and
  computer-assisted intervention}, pages 424--432. Springer, 2016.

\bibitem{clark2013cancer}
Kenneth Clark, Bruce Vendt, Kirk Smith, John Freymann, Justin Kirby, Paul
  Koppel, Stephen Moore, Stanley Phillips, David Maffitt, Michael Pringle,
  et~al.
\newblock The cancer imaging archive ({TCIA}): maintaining and operating a
  public information repository.
\newblock {\em Journal of digital imaging}, 26(6):1045--1057, 2013.

\bibitem{codella2018skin}
Noel~CF Codella, David Gutman, M~Emre Celebi, Brian Helba, Michael~A Marchetti,
  Stephen~W Dusza, Aadi Kalloo, Konstantinos Liopyris, Nabin Mishra, Harald
  Kittler, et~al.
\newblock Skin lesion analysis toward melanoma detection: A challenge at the
  2017 international symposium on biomedical imaging ({ISBI}), hosted by the
  international skin imaging collaboration ({ISIC}).
\newblock In {\em 2018 IEEE 15th international symposium on biomedical imaging
  (ISBI 2018)}, pages 168--172. IEEE, 2018.

\bibitem{cohen2017emnist}
Gregory Cohen, Saeed Afshar, Jonathan Tapson, and Andre Van~Schaik.
\newblock {EMNIST}: Extending {MNIST} to handwritten letters.
\newblock In {\em 2017 International Joint Conference on Neural Networks
  (IJCNN)}, pages 2921--2926. IEEE, 2017.

\bibitem{combalia2019bcn20000}
Marc Combalia, Noel~CF Codella, Veronica Rotemberg, Brian Helba, Veronica
  Vilaplana, Ofer Reiter, Cristina Carrera, Alicia Barreiro, Allan~C Halpern,
  Susana Puig, et~al.
\newblock Bcn20000: Dermoscopic lesions in the wild.
\newblock {\em arXiv preprint arXiv:1908.02288}, 2019.

\bibitem{Cordts2016Cityscapes}
Marius Cordts, Mohamed Omran, Sebastian Ramos, Timo Rehfeld, Markus Enzweiler,
  Rodrigo Benenson, Uwe Franke, Stefan Roth, and Bernt Schiele.
\newblock The cityscapes dataset for semantic urban scene understanding.
\newblock In {\em Proc. of the IEEE Conference on Computer Vision and Pattern
  Recognition (CVPR)}, 2016.

\bibitem{corinzia2019variational}
Luca Corinzia, Ami Beuret, and Joachim~M Buhmann.
\newblock Variational federated multi-task learning.
\newblock {\em arXiv preprint arXiv:1906.06268}, 2019.

\bibitem{courtiol2018classification}
Pierre Courtiol, Eric~W Tramel, Marc Sanselme, and Gilles Wainrib.
\newblock Classification and disease localization in histopathology using only
  global labels: A weakly-supervised approach.
\newblock {\em arXiv preprint arXiv:1802.02212}, 2018.

\bibitem{cox1972regression}
David~R Cox.
\newblock Regression models and life-tables.
\newblock {\em Journal of the Royal Statistical Society: Series B
  (Methodological)}, 34(2):187--202, 1972.

\bibitem{dayan2021federated}
Ittai Dayan, Holger~R Roth, Aoxiao Zhong, Ahmed Harouni, Amilcare Gentili,
  Anas~Z Abidin, Andrew Liu, Anthony~Beardsworth Costa, Bradford~J Wood,
  Chien-Sung Tsai, et~al.
\newblock Federated learning for predicting clinical outcomes in patients with
  covid-19.
\newblock {\em Nature medicine}, 27(10):1735--1743, 2021.

\bibitem{de2021deep}
Kevin de~Haan, Yijie Zhang, Jonathan~E Zuckerman, Tairan Liu, Anthony~E Sisk,
  Miguel~FP Diaz, Kuang-Yu Jen, Alexander Nobori, Sofia Liou, Sarah Zhang,
  et~al.
\newblock Deep learning-based transformation of {H}\&{E} stained tissues into
  special stains.
\newblock {\em Nature communications}, 12(1):1--13, 2021.

\bibitem{dehaene2020self}
Olivier Dehaene, Axel Camara, Olivier Moindrot, Axel de~Lavergne, and Pierre
  Courtiol.
\newblock Self-supervision closes the gap between weak and strong supervision
  in histology.
\newblock {\em arXiv preprint arXiv:2012.03583}, 2020.

\bibitem{deng2009imagenet}
Jia Deng, Wei Dong, Richard Socher, Li-Jia Li, Kai Li, and Li~Fei-Fei.
\newblock Imagenet: A large-scale hierarchical image database.
\newblock In {\em 2009 IEEE conference on computer vision and pattern
  recognition}, pages 248--255. Ieee, 2009.

\bibitem{ixi}
Brain development team.
\newblock Ixi dataset.
\newblock \url{https://brain-development.org/ixi-dataset/}.
\newblock Accessed: 2022-02-02.

\bibitem{dice1945measures}
Lee~R Dice.
\newblock Measures of the amount of ecologic association between species.
\newblock {\em Ecology}, 26(3):297--302, 1945.

\bibitem{du2021collaborative}
Jean~Ogier du~Terrail, Armand Leopold, Cl{\'e}ment Joly, Constance Beguier,
  Mathieu Andreux, Charles Maussion, Benoit Schmauch, Eric~W Tramel, Etienne
  Bendjebbar, Mikhail Zaslavskiy, et~al.
\newblock Collaborative federated learning behind hospitals' firewalls for
  predicting histological response to neoadjuvant chemotherapy in
  triple-negative breast cancer.
\newblock {\em medRxiv}, 2021.

\bibitem{duarte2004vehicle}
Marco~F Duarte and Yu~Hen Hu.
\newblock Vehicle classification in distributed sensor networks.
\newblock {\em Journal of Parallel and Distributed Computing}, 64(7):826--838,
  2004.

\bibitem{dwork2014algorithmic}
Cynthia Dwork, Aaron Roth, et~al.
\newblock The algorithmic foundations of differential privacy.
\newblock {\em Found. Trends Theor. Comput. Sci.}, 9(3-4):211--407, 2014.

\bibitem{fallah2020personalized}
Alireza Fallah, Aryan Mokhtari, and Asuman Ozdaglar.
\newblock Personalized federated learning: A meta-learning approach.
\newblock {\em arXiv preprint arXiv:2002.07948}, 2020.

\bibitem{fate}
FedAI-maintainers.
\newblock Fate (federated ai technology enabler).
\newblock \url{https://github.com/FederatedAI/FATE}.
\newblock Accessed: 2022-10-12.

\bibitem{fraboni2021impact}
Yann Fraboni, Richard Vidal, Laetitia Kameni, and Marco Lorenzi.
\newblock On the impact of client sampling on federated learning convergence.
\newblock {\em arXiv preprint arXiv:2107.12211}, 2021.

\bibitem{fu2020pan}
Yu~Fu, Alexander~W Jung, Ramon~Vi{\~n}as Torne, Santiago Gonzalez, Harald
  V{\"o}hringer, Artem Shmatko, Lucy~R Yates, Mercedes Jimenez-Linan, Luiza
  Moore, and Moritz Gerstung.
\newblock Pan-cancer computational histopathology reveals mutations, tumor
  composition and prognosis.
\newblock {\em Nature Cancer}, 1(8):800--810, 2020.

\bibitem{galtier2019substra}
Mathieu~N Galtier and Camille Marini.
\newblock Substra: a framework for privacy-preserving, traceable and
  collaborative machine learning.
\newblock {\em arXiv preprint arXiv:1910.11567}, 2019.

\bibitem{go2009twitter}
Alec Go, Richa Bhayani, and Lei Huang.
\newblock Twitter sentiment classification using distant supervision.
\newblock {\em CS224N project report, Stanford}, 1(12):2009, 2009.

\bibitem{gong2021ensemble}
Xuan Gong, Abhishek Sharma, Srikrishna Karanam, Ziyan Wu, Terrence Chen, David
  Doermann, and Arun Innanje.
\newblock Ensemble attention distillation for privacy-preserving federated
  learning.
\newblock In {\em Proceedings of the IEEE/CVF International Conference on
  Computer Vision}, pages 15076--15086, 2021.

\bibitem{graham2015kaggle}
Ben Graham.
\newblock Kaggle diabetic retinopathy detection competition report.
\newblock {\em University of Warwick}, pages 24--26, 2015.

\bibitem{gunesli2021feddropoutavg}
Gozde~N Gunesli, Mohsin Bilal, Shan E~Ahmed Raza, and Nasir~M Rajpoot.
\newblock Feddropoutavg: Generalizable federated learning for histopathology
  image classification.
\newblock {\em arXiv preprint arXiv:2111.13230}, 2021.

\bibitem{haddadpour2021federated}
Farzin Haddadpour, Mohammad~Mahdi Kamani, Aryan Mokhtari, and Mehrdad Mahdavi.
\newblock Federated learning with compression: Unified analysis and sharp
  guarantees.
\newblock In {\em International Conference on Artificial Intelligence and
  Statistics}, pages 2350--2358. PMLR, 2021.

\bibitem{hahn2006adrenal}
Peter~F Hahn, Michael~A Blake, and Giles~WL Boland.
\newblock Adrenal lesions: attenuation measurement differences between ct
  scanners.
\newblock {\em Radiology}, 240(2):458--463, 2006.

\bibitem{he2020group}
Chaoyang He, Murali Annavaram, and Salman Avestimehr.
\newblock Group knowledge transfer: Federated learning of large {CNN}s at the
  edge.
\newblock {\em arXiv preprint arXiv:2007.14513}, 2020.

\bibitem{he2021fedgraphnn}
Chaoyang He, Keshav Balasubramanian, Emir Ceyani, Carl Yang, Han Xie, Lichao
  Sun, Lifang He, Liangwei Yang, Philip~S Yu, Yu~Rong, et~al.
\newblock Fedgraphnn: A federated learning system and benchmark for graph
  neural networks.
\newblock {\em arXiv preprint arXiv:2104.07145}, 2021.

\bibitem{he2020fedml}
Chaoyang He, Songze Li, Jinhyun So, Xiao Zeng, Mi~Zhang, Hongyi Wang, Xiaoyang
  Wang, Praneeth Vepakomma, Abhishek Singh, Hang Qiu, et~al.
\newblock Fedml: A research library and benchmark for federated machine
  learning.
\newblock {\em arXiv preprint arXiv:2007.13518}, 2020.

\bibitem{he2021fedcv}
Chaoyang He, Alay~Dilipbhai Shah, Zhenheng Tang, Di~Fan1Adarshan~Naiynar
  Sivashunmugam, Keerti Bhogaraju, Mita Shimpi, Li~Shen, Xiaowen Chu, Mahdi
  Soltanolkotabi, and Salman Avestimehr.
\newblock Fedcv: A federated learning framework for diverse computer vision
  tasks.
\newblock {\em arXiv preprint arXiv:2111.11066}, 2021.

\bibitem{heller2020state}
Nicholas Heller, Fabian Isensee, Klaus~H Maier-Hein, Xiaoshuai Hou, Chunmei
  Xie, Fengyi Li, Yang Nan, Guangrui Mu, Zhiyong Lin, Miofei Han, et~al.
\newblock The state of the art in kidney and kidney tumor segmentation in
  contrast-enhanced ct imaging: Results of the kits19 challenge.
\newblock {\em Medical Image Analysis}, page 101821, 2020.

\bibitem{heller2019kits19}
Nicholas Heller, Niranjan Sathianathen, Arveen Kalapara, Edward Walczak, Keenan
  Moore, Heather Kaluzniak, Joel Rosenberg, Paul Blake, Zachary Rengel, Makinna
  Oestreich, et~al.
\newblock The kits19 challenge data: 300 kidney tumor cases with clinical
  context, ct semantic segmentations, and surgical outcomes.
\newblock {\em arXiv preprint arXiv:1904.00445}, 2019.

\bibitem{howard2021impact}
Frederick~M Howard, James Dolezal, Sara Kochanny, Jefree Schulte, Heather Chen,
  Lara Heij, Dezheng Huo, Rita Nanda, Olufunmilayo~I Olopade, Jakob~N Kather,
  et~al.
\newblock The impact of site-specific digital histology signatures on deep
  learning model accuracy and bias.
\newblock {\em Nature communications}, 12(1):1--13, 2021.

\bibitem{icml2020_3152}
Kevin Hsieh, Amar Phanishayee, Onur Mutlu, and Phillip Gibbons.
\newblock The non-{IID} data quagmire of decentralized machine learning.
\newblock In {\em International Conference on Machine Learning ({ICML})}, pages
  5819--5830. PMLR, 2020.

\bibitem{hsu2019measuring}
Tzu-Ming~Harry Hsu, Hang Qi, and Matthew Brown.
\newblock Measuring the effects of non-identical data distribution for
  federated visual classification.
\newblock {\em arXiv preprint arXiv:1909.06335}, 2019.

\bibitem{hsu2020federated}
Tzu-Ming~Harry Hsu, Hang Qi, and Matthew Brown.
\newblock Federated visual classification with real-world data distribution.
\newblock In {\em European Conference on Computer Vision}, pages 76--92.
  Springer, 2020.

\bibitem{iglesias2011robust}
Juan~Eugenio Iglesias, Cheng-Yi Liu, Paul~M Thompson, and Zhuowen Tu.
\newblock Robust brain extraction across datasets and comparison with publicly
  available methods.
\newblock {\em IEEE transactions on medical imaging}, 30(9):1617--1634, 2011.

\bibitem{lidcdata}
S.~G.~Armato III, G.~McLennan, L.~Bidaut, M.~F. McNitt-Gray, C.~R. Meyer, A.~P.
  Reeves, B.~Zhao, D.~R. Aberle, C.~I. Henschke, E.~A. Hoffman, E.~A.
  Kazerooni, H.~MacMahon, E.~J. R.~Van Beek, D.~Yankelevitz, A.~M. Biancardi,
  P.~H. Bland, M.~S. Brown, R.~M. Engelmann, G.~E. Laderach, D.~Max, R.~C.
  Pais, D.~P.~Y. Qing, R.~Y. Roberts, A.~R. Smith, A.~Starkey, P.~Batra,
  P.~Caligiuri, A.~Farooqi, G.~W. Gladish, C.~M. Jude, R.~F. Munden,
  I.~Petkovska, L.~E. Quint, L.~H. Schwartz, B.~Sundaram, L.~E. Dodd,
  C.~Fenimore, D.~Gur, N.~Petrick, J.~Freymann, J.~Kirby, B.~Hughes, A.~V.
  Casteele, S.~Gupte ans M.~Sallam, M.~D. Heath, M.~H. Kuhn, E.~Dharaiya,
  R.~Burns, D.~S. Fryd, M.~Salganicoff, V.~Anand, U.~Shreter, S.~Vastagh, B.~Y.
  Croft, and L.~P. Clarke.
\newblock Data from lidc-idri [data set]. the cancer imaging archive., 2015.

\bibitem{ilse2018attention}
Maximilian Ilse, Jakub Tomczak, and Max Welling.
\newblock Attention-based deep multiple instance learning.
\newblock In {\em International conference on machine learning}, pages
  2127--2136. PMLR, 2018.

\bibitem{deepmil}
Maximilian Ilse, Jakub~M. Tomczak, and Max Welling.
\newblock Attention-based deep multiple instance learning.
\newblock \url{https://github.com/AMLab-Amsterdam/AttentionDeepMIL}.
\newblock Accessed: 2022-02-02.

\bibitem{irvin2019chexpert}
Jeremy Irvin, Pranav Rajpurkar, Michael Ko, Yifan Yu, Silviana Ciurea-Ilcus,
  Chris Chute, Henrik Marklund, Behzad Haghgoo, Robyn Ball, Katie Shpanskaya,
  et~al.
\newblock Chexpert: A large chest radiograph dataset with uncertainty labels
  and expert comparison.
\newblock In {\em Proceedings of the AAAI conference on artificial
  intelligence}, volume~33, pages 590--597, 2019.

\bibitem{isensee2020batchgenerators}
F~Isensee, P~J{\"a}ger, J~Wasserthal, D~Zimmerer, J~Petersen, S~Kohl, J~Schock,
  A~Klein, T~RoSS, S~Wirkert, et~al.
\newblock batchgenerators—a python framework for data augmentation.
\newblock {\em Zenodo https://doi. org/10.5281/zenodo}, 3632567, 2020.

\bibitem{isensee2021nnu}
Fabian Isensee, Paul~F Jaeger, Simon~AA Kohl, Jens Petersen, and Klaus~H
  Maier-Hein.
\newblock nnu-net: a self-configuring method for deep learning-based biomedical
  image segmentation.
\newblock {\em Nature methods}, 18(2):203--211, 2021.

\bibitem{janosi1988heart}
Andras Janosi, William Steinbrunn, Matthias Pfisterer, and Robert Detrano.
\newblock Heart disease data set, 1988.

\bibitem{janowczyk2019histoqc}
Andrew Janowczyk, Ren Zuo, Hannah Gilmore, Michael Feldman, and Anant
  Madabhushi.
\newblock Histoqc: an open-source quality control tool for digital pathology
  slides.
\newblock {\em JCO clinical cancer informatics}, 3:1--7, 2019.

\bibitem{jenkins2005survival}
Stephen~P Jenkins.
\newblock Survival analysis.
\newblock {\em Unpublished manuscript, Institute for Social and Economic
  Research, University of Essex, Colchester, UK}, 42:54--56, 2005.

\bibitem{kairouz2019advances}
Peter Kairouz, H.~Brendan McMahan, Brendan Avent, Aurélien Bellet, Mehdi
  Bennis, Arjun~Nitin Bhagoji, Kallista Bonawitz, Zachary Charles, Graham
  Cormode, Rachel Cummings, Rafael G.~L. D’Oliveira, Hubert Eichner, Salim~El
  Rouayheb, David Evans, Josh Gardner, Zachary Garrett, Adrià Gascón, Badih
  Ghazi, Phillip~B. Gibbons, Marco Gruteser, Zaid Harchaoui, Chaoyang He, Lie
  He, Zhouyuan Huo, Ben Hutchinson, Justin Hsu, Martin Jaggi, Tara Javidi,
  Gauri Joshi, Mikhail Khodak, Jakub Konecný, Aleksandra Korolova, Farinaz
  Koushanfar, Sanmi Koyejo, Tancrède Lepoint, Yang Liu, Prateek Mittal,
  Mehryar Mohri, Richard Nock, Ayfer Özgür, Rasmus Pagh, Hang Qi, Daniel
  Ramage, Ramesh Raskar, Mariana Raykova, Dawn Song, Weikang Song, Sebastian~U.
  Stich, Ziteng Sun, Ananda~Theertha Suresh, Florian Tramèr, Praneeth
  Vepakomma, Jianyu Wang, Li~Xiong, Zheng Xu, Qiang Yang, Felix~X. Yu, Han Yu,
  and Sen Zhao.
\newblock Advances and open problems in federated learning.
\newblock {\em Foundations and Trends® in Machine Learning}, 14(1–2):1--210,
  2021.

\bibitem{kaissis2021end}
Georgios Kaissis, Alexander Ziller, Jonathan Passerat-Palmbach, Th{\'e}o
  Ryffel, Dmitrii Usynin, Andrew Trask, Ion{\'e}sio Lima, Jason Mancuso,
  Friederike Jungmann, Marc-Matthias Steinborn, et~al.
\newblock End-to-end privacy preserving deep learning on multi-institutional
  medical imaging.
\newblock {\em Nature Machine Intelligence}, 3(6):473--484, 2021.

\bibitem{kaplan1958nonparametric}
Edward~L Kaplan and Paul Meier.
\newblock Nonparametric estimation from incomplete observations.
\newblock {\em Journal of the American statistical association},
  53(282):457--481, 1958.

\bibitem{kaplan2020scaling}
Jared Kaplan, Sam McCandlish, Tom Henighan, Tom~B Brown, Benjamin Chess, Rewon
  Child, Scott Gray, Alec Radford, Jeffrey Wu, and Dario Amodei.
\newblock Scaling laws for neural language models.
\newblock {\em arXiv preprint arXiv:2001.08361}, 2020.

\bibitem{karimireddy2020scaffold}
Sai~Praneeth Karimireddy, Satyen Kale, Mehryar Mohri, Sashank Reddi, Sebastian
  Stich, and Ananda~Theertha Suresh.
\newblock Scaffold: Stochastic controlled averaging for federated learning.
\newblock In {\em International Conference on Machine Learning}, pages
  5132--5143. PMLR, 2020.

\bibitem{kermany2018identifying}
Daniel~S Kermany, Michael Goldbaum, Wenjia Cai, Carolina~CS Valentim, Huiying
  Liang, Sally~L Baxter, Alex McKeown, Ge~Yang, Xiaokang Wu, Fangbing Yan,
  et~al.
\newblock Identifying medical diagnoses and treatable diseases by image-based
  deep learning.
\newblock {\em Cell}, 172(5):1122--1131, 2018.

\bibitem{kingma2014adam}
Diederik~P Kingma and Jimmy Ba.
\newblock Adam: A method for stochastic optimization.
\newblock {\em arXiv preprint arXiv:1412.6980}, 2014.

\bibitem{Krizhevsky09learningmultiple}
Alex Krizhevsky, Geoffrey Hinton, et~al.
\newblock Learning multiple layers of features from tiny images, 2009.

\bibitem{lahiani2020seamless}
Amal Lahiani, Irina Klaman, Nassir Navab, Shadi Albarqouni, and Eldad Klaiman.
\newblock Seamless virtual whole slide image synthesis and validation using
  perceptual embedding consistency.
\newblock {\em IEEE Journal of Biomedical and Health Informatics},
  25(2):403--411, 2020.

\bibitem{lai2022fedscale}
Fan Lai, Yinwei Dai, Sanjay Singapuram, Jiachen Liu, Xiangfeng Zhu, Harsha
  Madhyastha, and Mosharaf Chowdhury.
\newblock Fedscale: Benchmarking model and system performance of federated
  learning at scale.
\newblock In {\em International Conference on Machine Learning}, pages
  11814--11827. PMLR, 2022.

\bibitem{lecun-mnisthandwrittendigit-2010}
Yann LeCun and Corinna Cortes.
\newblock {MNIST} handwritten digit database.
\newblock http://yann.lecun.com/exdb/mnist/, 2010.

\bibitem{li2020suvey}
Tian Li, Anit~Kumar Sahu, Ameet Talwalkar, and Virginia Smith.
\newblock Federated learning: Challenges, methods, and future directions.
\newblock {\em IEEE Signal Processing Magazine}, 37(3):50--60, 2020.

\bibitem{li2020federated}
Tian Li, Anit~Kumar Sahu, Manzil Zaheer, Maziar Sanjabi, Ameet Talwalkar, and
  Virginia Smith.
\newblock Federated optimization in heterogeneous networks.
\newblock {\em Proceedings of Machine Learning and Systems}, 2:429--450, 2020.

\bibitem{li2006pachinko}
Wei Li and Andrew McCallum.
\newblock Pachinko allocation: {DAG}-structured mixture models of topic
  correlations.
\newblock In {\em Proceedings of the 23rd international conference on Machine
  learning}, pages 577--584, 2006.

\bibitem{lim2020federated}
Wei Yang~Bryan Lim, Nguyen~Cong Luong, Dinh~Thai Hoang, Yutao Jiao, Ying-Chang
  Liang, Qiang Yang, Dusit Niyato, and Chunyan Miao.
\newblock Federated learning in mobile edge networks: A comprehensive survey.
\newblock {\em IEEE Communications Surveys \& Tutorials}, 22(3):2031--2063,
  2020.

\bibitem{DBLP:journals/corr/abs-1708-02002}
Tsung{-}Yi Lin, Priya Goyal, Ross~B. Girshick, Kaiming He, and Piotr
  Doll{\'{a}}r.
\newblock Focal loss for dense object detection.
\newblock {\em CoRR}, abs/1708.02002, 2017.

\bibitem{litjens20181399}
Geert Litjens, Peter Bandi, Babak Ehteshami~Bejnordi, Oscar Geessink, Maschenka
  Balkenhol, Peter Bult, Altuna Halilovic, Meyke Hermsen, Rob van~de Loo, Rob
  Vogels, et~al.
\newblock 1399 {H}\&{E}-stained sentinel lymph node sections of breast cancer
  patients: the {CAMELYON} dataset.
\newblock {\em GigaScience}, 7(6):giy065, 2018.

\bibitem{liu2015deep}
Ziwei Liu, Ping Luo, Xiaogang Wang, and Xiaoou Tang.
\newblock Deep learning face attributes in the wild.
\newblock In {\em Proceedings of the IEEE international conference on computer
  vision}, pages 3730--3738, 2015.

\bibitem{lu2022federated}
Ming~Y Lu, Richard~J Chen, Dehan Kong, Jana Lipkova, Rajendra Singh, Drew~FK
  Williamson, Tiffany~Y Chen, and Faisal Mahmood.
\newblock Federated learning for computational pathology on gigapixel whole
  slide images.
\newblock {\em Medical image analysis}, 76:102298, 2022.

\bibitem{lu2021data}
Ming~Y Lu, Drew~FK Williamson, Tiffany~Y Chen, Richard~J Chen, Matteo Barbieri,
  and Faisal Mahmood.
\newblock Data-efficient and weakly supervised computational pathology on
  whole-slide images.
\newblock {\em Nature biomedical engineering}, 5(6):555--570, 2021.

\bibitem{lu2020low}
Yunlong Lu, Xiaohong Huang, Ke~Zhang, Sabita Maharjan, and Yan Zhang.
\newblock Low-latency federated learning and blockchain for edge association in
  digital twin empowered 6g networks.
\newblock {\em IEEE Transactions on Industrial Informatics}, 17(7):5098--5107,
  2020.

\bibitem{luo2019real}
Jiahuan Luo, Xueyang Wu, Yun Luo, Anbu Huang, Yunfeng Huang, Yang Liu, and
  Qiang Yang.
\newblock Real-world image datasets for federated learning.
\newblock {\em arXiv preprint arXiv:1910.11089}, 2019.

\bibitem{luofediris}
Zhengquan Luo, Yunlong Wang, Zilei Wang, Zhenan Sun, and Tieniu Tan.
\newblock Fediris: Towards more accurate and privacy-preserving iris
  recognition via federated template communication.
\newblock {\em CVPRW}, 2022.

\bibitem{lydia2019adagrad}
Agnes Lydia and Sagayaraj Francis.
\newblock Adagrad—an optimizer for stochastic gradient descent.
\newblock {\em Int. J. Inf. Comput. Sci}, 6(5):566--568, 2019.

\bibitem{marfoq20neurips}
Othmane Marfoq, Chuan Xu, Giovanni Neglia, and Richard Vidal.
\newblock {Throughput-Optimal Topology Design for Cross-Silo Federated
  Learning}.
\newblock In {\em {34th Conference on Neural Information Processing Systems
  (NeurIPS 2020)}}, Vancouver, Canada, December 2020.
\newblock NeurIPS 2020.

\bibitem{mcinnes2018umap}
Leland McInnes, John Healy, and James Melville.
\newblock Umap: Uniform manifold approximation and projection for dimension
  reduction.
\newblock {\em arXiv preprint arXiv:1802.03426}, 2018.

\bibitem{mcmahan2017communication}
Brendan McMahan, Eider Moore, Daniel Ramage, Seth Hampson, and Blaise~Aguera
  y~Arcas.
\newblock Communication-efficient learning of deep networks from decentralized
  data.
\newblock In {\em Artificial intelligence and statistics}, pages 1273--1282.
  PMLR, 2017.

\bibitem{milletari2016v}
Fausto Milletari, Nassir Navab, and Seyed-Ahmad Ahmadi.
\newblock V-net: Fully convolutional neural networks for volumetric medical
  image segmentation.
\newblock In {\em 2016 fourth international conference on 3D vision (3DV)},
  pages 565--571. IEEE, 2016.

\bibitem{tcga}
TCGA~Research Network.
\newblock Tensorflow federated stack overflow dataset.
\newblock \url{https://www.cancer.gov/tcga}.
\newblock Accessed: 2022-05-18.

\bibitem{adaloglou2019MRIsegmentation}
Adaloglou Nikolaos.
\newblock Deep learning in medical image analysis: a comparative analysis of
  multi-modal brain-mri segmentation with 3d deep neural networks.
\newblock Master's thesis, University of Patras, 2019.
\newblock \url{https://github.com/black0017/MedicalZooPytorch}.

\bibitem{niu2020billion}
Chaoyue Niu, Fan Wu, Shaojie Tang, Lifeng Hua, Rongfei Jia, Chengfei Lv, Zhihua
  Wu, and Guihai Chen.
\newblock Billion-scale federated learning on mobile clients: a submodel design
  with tunable privacy.
\newblock In {\em Proceedings of the 26th Annual International Conference on
  Mobile Computing and Networking}, pages 1--14, 2020.

\bibitem{paszke2019pytorch}
Adam Paszke, Sam Gross, Francisco Massa, Adam Lerer, James Bradbury, Gregory
  Chanan, Trevor Killeen, Zeming Lin, Natalia Gimelshein, Luca Antiga, et~al.
\newblock Pytorch: An imperative style, high-performance deep learning library.
\newblock {\em Advances in neural information processing systems}, 32, 2019.

\bibitem{pati2021federated}
Sarthak Pati, Ujjwal Baid, Maximilian Zenk, Brandon Edwards, Micah Sheller,
  G~Anthony Reina, Patrick Foley, Alexey Gruzdev, Jason Martin, Shadi
  Albarqouni, et~al.
\newblock The federated tumor segmentation ({FeTS}) challenge.
\newblock {\em arXiv preprint arXiv:2105.05874}, 2021.

\bibitem{perez2021torchio}
Fernando P{\'e}rez-Garc{\'\i}a, Rachel Sparks, and Sebastien Ourselin.
\newblock Torchio: a python library for efficient loading, preprocessing,
  augmentation and patch-based sampling of medical images in deep learning.
\newblock {\em Computer Methods and Programs in Biomedicine}, 208:106236, 2021.

\bibitem{philippenko2020bidirectional}
Constantin Philippenko and Aymeric Dieuleveut.
\newblock Bidirectional compression in heterogeneous settings for distributed
  or federated learning with partial participation: tight convergence
  guarantees.
\newblock {\em arXiv preprint arXiv:2006.14591}, 2020.

\bibitem{phillips2008iris}
P~Jonathon Phillips, Kevin~W Bowyer, Patrick~J Flynn, Xiaomei Liu, and W~Todd
  Scruggs.
\newblock The iris challenge evaluation 2005.
\newblock In {\em 2008 IEEE Second International Conference on Biometrics:
  Theory, Applications and Systems}, pages 1--8. IEEE, 2008.

\bibitem{iximodel}
unet 0.7.7.
\newblock \url{https://pypi.org/project/unet/0.7.7/}.
\newblock Accessed: 2022-02-02.

\bibitem{ixitiny}
Fernando Pérez-García.
\newblock Ixitiny dataset.
\newblock
  \url{https://torchio.readthedocs.io/datasets.html#torchio.datasets.ixi.IXITiny}.
\newblock Accessed: 2022-05-18.

\bibitem{dalle2}
Aditya Ramesh, Prafulla Dhariwal, Alex Nichol, Casey Chu, and Mark Chen.
\newblock Hierarchical text-conditional image generation with clip latents,
  2022.

\bibitem{reddi2020adaptive}
Sashank Reddi, Zachary Charles, Manzil Zaheer, Zachary Garrett, Keith Rush,
  Jakub Kone{\v{c}}n{\`y}, Sanjiv Kumar, and H~Brendan McMahan.
\newblock Adaptive federated optimization.
\newblock {\em arXiv preprint arXiv:2003.00295}, 2020.

\bibitem{flsim}
Meta~AI Research.
\newblock Federated learning simulator (flsim).
\newblock \url{https://github.com/facebookresearch/FLSim/tree/main/examples},
  2012.

\bibitem{rieke2020future}
Nicola Rieke, Jonny Hancox, Wenqi Li, Fausto Milletari, Holger~R Roth, Shadi
  Albarqouni, Spyridon Bakas, Mathieu~N Galtier, Bennett~A Landman, Klaus
  Maier-Hein, et~al.
\newblock The future of digital health with federated learning.
\newblock {\em NPJ digital medicine}, 3(1):1--7, 2020.

\bibitem{sattler2019sparse}
Felix Sattler, Simon Wiedemann, Klaus-Robert M{\"u}ller, and Wojciech Samek.
\newblock Sparse binary compression: Towards distributed deep learning with
  minimal communication.
\newblock In {\em 2019 International Joint Conference on Neural Networks
  (IJCNN)}, pages 1--8. IEEE, 2019.

\bibitem{setio2017validation}
Arnaud Arindra~Adiyoso Setio, Alberto Traverso, Thomas De~Bel, Moira~SN Berens,
  Cas Van Den~Bogaard, Piergiorgio Cerello, Hao Chen, Qi~Dou, Maria~Evelina
  Fantacci, Bram Geurts, et~al.
\newblock Validation, comparison, and combination of algorithms for automatic
  detection of pulmonary nodules in computed tomography images: the {LUNA16}
  challenge.
\newblock {\em Medical image analysis}, 42:1--13, 2017.

\bibitem{sheller2020federated}
Micah~J Sheller, Brandon Edwards, G~Anthony Reina, Jason Martin, Sarthak Pati,
  Aikaterini Kotrotsou, Mikhail Milchenko, Weilin Xu, Daniel Marcus, Rivka~R
  Colen, et~al.
\newblock Federated learning in medicine: facilitating multi-institutional
  collaborations without sharing patient data.
\newblock {\em Scientific reports}, 10(1):1--12, 2020.

\bibitem{sheller2018multi}
Micah~J Sheller, G~Anthony Reina, Brandon Edwards, Jason Martin, and Spyridon
  Bakas.
\newblock Multi-institutional deep learning modeling without sharing patient
  data: A feasibility study on brain tumor segmentation.
\newblock In {\em International MICCAI Brainlesion Workshop}, pages 92--104.
  Springer, 2018.

\bibitem{silva2020fed}
Santiago Silva, Andre Altmann, Boris Gutman, and Marco Lorenzi.
\newblock Fed-{BioMed}: A general open-source frontend framework for federated
  learning in healthcare.
\newblock In {\em Domain Adaptation and Representation Transfer, and
  Distributed and Collaborative Learning}, pages 201--210. Springer, 2020.

\bibitem{sun2017revisiting}
Chen Sun, Abhinav Shrivastava, Saurabh Singh, and Abhinav Gupta.
\newblock Revisiting unreasonable effectiveness of data in deep learning era.
\newblock In {\em Proceedings of the IEEE international conference on computer
  vision}, pages 843--852, 2017.

\bibitem{DBLP:journals/corr/abs-1905-11946}
Mingxing Tan and Quoc~V. Le.
\newblock Efficientnet: Rethinking model scaling for convolutional neural
  networks.
\newblock {\em CoRR}, abs/1905.11946, 2019.

\bibitem{tenso2019stack}
Tensorflow.
\newblock Tensorflow federated stack overflow dataset.
\newblock \url{https:
  //www.tensorflow.org/federated/api_docs/python/tff/simulation/datasets/
  stackoverflow/load_data}, 2019.

\bibitem{tomczak2015cancer}
Katarzyna Tomczak, Patrycja Czerwi{\'n}ska, and Maciej Wiznerowicz.
\newblock The cancer genome atlas ({TCGA}): an immeasurable source of
  knowledge.
\newblock {\em Contemporary oncology}, 19(1A):A68, 2015.

\bibitem{torralba2011unbiased}
Antonio Torralba and Alexei~A Efros.
\newblock Unbiased look at dataset bias.
\newblock In {\em CVPR 2011}, pages 1521--1528. IEEE, 2011.

\bibitem{tschandl2018ham10000}
Philipp Tschandl, Cliff Rosendahl, and Harald Kittler.
\newblock The {HAM10000} dataset, a large collection of multi-source
  dermatoscopic images of common pigmented skin lesions.
\newblock {\em Scientific data}, 5(1):1--9, 2018.

\bibitem{van2008visualizing}
Laurens Van~der Maaten and Geoffrey Hinton.
\newblock Visualizing data using t-{SNE}.
\newblock {\em Journal of machine learning research}, 9(11), 2008.

\bibitem{horn2018inaturalist}
G.~{Van Horn}, O.~{Mac Aodha}, Y.~{Song}, Y.~{Cui}, C.~{Sun}, A.~{Shepard},
  H.~{Adam}, P.~{Perona}, and S.~{Belongie}.
\newblock The inaturalist species classification and detection dataset.
\newblock In {\em 2018 IEEE/CVF Conference on Computer Vision and Pattern
  Recognition}, pages 8769--8778, 2018.

\bibitem{veta2014breast}
Mitko Veta, Josien~PW Pluim, Paul~J Van~Diest, and Max~A Viergever.
\newblock Breast cancer histopathology image analysis: A review.
\newblock {\em IEEE transactions on biomedical engineering}, 61(5):1400--1411,
  2014.

\bibitem{volskeetal2017tl}
Michael V{\"o}lske, Martin Potthast, Shahbaz Syed, and Benno Stein.
\newblock {TL};{DR}: Mining {R}eddit to learn automatic summarization.
\newblock In {\em Proceedings of the Workshop on New Frontiers in
  Summarization}, pages 59--63, Copenhagen, Denmark, September 2017.
  Association for Computational Linguistics.

\bibitem{wei2007nonlinear}
Zhuoshi Wei, Tieniu Tan, and Zhenan Sun.
\newblock Nonlinear iris deformation correction based on gaussian model.
\newblock In {\em International Conference on Biometrics}, pages 780--789.
  Springer, 2007.

\bibitem{xia2021survey}
Qi~Xia, Winson Ye, Zeyi Tao, Jindi Wu, and Qun Li.
\newblock A survey of federated learning for edge computing: Research problems
  and solutions.
\newblock {\em High-Confidence Computing}, page 100008, 2021.

\bibitem{xie2022fedmed}
Guoyang Xie, Jinbao Wang, Yawen Huang, Yefeng Zheng, Feng Zheng, Jingkuang
  Song, and Yaochu Jin.
\newblock {FedMed-GAN}: Federated multi-modal unsupervised brain image
  synthesis.
\newblock {\em arXiv preprint arXiv:2201.08953}, 2022.

\bibitem{yang2021achieving}
Haibo Yang, Minghong Fang, and Jia Liu.
\newblock Achieving linear speedup with partial worker participation in non-iid
  federated learning.
\newblock {\em arXiv preprint arXiv:2101.11203}, 2021.

\bibitem{yuchen2022fednlp}
Bill Yuchen~Lin, Chaoyang He, Zihang Zeng, Hulin Wang, Yufen Huang, Christophe
  Dupuy, Rahul Gupta, Mahdi Soltanolkotabi, Xiang Ren, and Salman Avestimehr.
\newblock Fednlp: Benchmarking federated learning methods for natural language
  processing tasks.
\newblock {\em Findings of NAACL}, 2022.

\bibitem{yurochkin2019bayesian}
Mikhail Yurochkin, Mayank Agarwal, Soumya Ghosh, Kristjan Greenewald, Nghia
  Hoang, and Yasaman Khazaeni.
\newblock Bayesian nonparametric federated learning of neural networks.
\newblock In {\em International Conference on Machine Learning}, pages
  7252--7261. PMLR, 2019.

\bibitem{zaheer2018adaptive}
Manzil Zaheer, Sashank Reddi, Devendra Sachan, Satyen Kale, and Sanjiv Kumar.
\newblock Adaptive methods for nonconvex optimization.
\newblock {\em Advances in neural information processing systems}, 31, 2018.

\bibitem{zhang2010contact}
Hui Zhang, Zhenan Sun, and Tieniu Tan.
\newblock Contact lens detection based on weighted lbp.
\newblock In {\em 2010 20th International Conference on Pattern Recognition},
  pages 4279--4282. IEEE, 2010.

\bibitem{zhang2018deep}
Qi~Zhang, Haiqing Li, Zhenan Sun, and Tieniu Tan.
\newblock Deep feature fusion for iris and periocular biometrics on mobile
  devices.
\newblock {\em IEEE Transactions on Information Forensics and Security},
  13(11):2897--2912, 2018.

\end{thebibliography}


\newcommand{\etalchar}[1]{$^{#1}$}
\begin{thebibliography}{VDWCV11}

\bibitem[ACG{\etalchar{+}}16]{abadi2016deep}
Martin Abadi, Andy Chu, Ian Goodfellow, H~Brendan McMahan, Ilya Mironov, Kunal
  Talwar, and Li~Zhang.
\newblock Deep learning with differential privacy.
\newblock In {\em Proceedings of the 2016 ACM SIGSAC conference on computer and
  communications security}, pages 308--318, 2016.

\bibitem[AIMB{\etalchar{+}}11]{armato2011lidc}
Samuel~G Armato~III, Geoffrey McLennan, Luc Bidaut, Michael~F McNitt-Gray,
  Charles~R Meyer, Anthony~P Reeves, Binsheng Zhao, Denise~R Aberle, Claudia~I
  Henschke, Eric~A Hoffman, et~al.
\newblock The lung image database consortium ({LIDC}) and image database
  resource initiative ({IDRI}): a completed reference database of lung nodules
  on ct scans.
\newblock {\em Medical physics}, 38(2):915--931, 2011.

\bibitem[AMM{\etalchar{+}}20]{andreux2020federated}
Mathieu Andreux, Andre Manoel, Romuald Menuet, Charlie Saillard, and Chlo{\'e}
  Simpson.
\newblock Federated survival analysis with discrete-time {Cox} models.
\newblock {\em arXiv preprint arXiv:2006.08997}, 2020.

\bibitem[Aro]{aman}
Aman Arora.
\newblock Siim-isic melanoma classification - my journey to a top 5\% solution
  and first silver medal on kaggle.
\newblock \url{https://amaarora.github.io/2020/08/23/siimisic.html}.
\newblock Accessed: 2022-02-02.

\bibitem[Bra97]{bradley1997use}
Andrew~P Bradley.
\newblock The use of the area under the roc curve in the evaluation of machine
  learning algorithms.
\newblock {\em Pattern recognition}, 30(7):1145--1159, 1997.

\bibitem[CCR{\etalchar{+}}19]{combalia2019bcn20000}
Marc Combalia, Noel~CF Codella, Veronica Rotemberg, Brian Helba, Veronica
  Vilaplana, Ofer Reiter, Cristina Carrera, Alicia Barreiro, Allan~C Halpern,
  Susana Puig, et~al.
\newblock Bcn20000: Dermoscopic lesions in the wild.
\newblock {\em arXiv preprint arXiv:1908.02288}, 2019.

\bibitem[CGC{\etalchar{+}}18]{codella2018skin}
Noel~CF Codella, David Gutman, M~Emre Celebi, Brian Helba, Michael~A Marchetti,
  Stephen~W Dusza, Aadi Kalloo, Konstantinos Liopyris, Nabin Mishra, Harald
  Kittler, et~al.
\newblock Skin lesion analysis toward melanoma detection: A challenge at the
  2017 international symposium on biomedical imaging ({ISBI}), hosted by the
  international skin imaging collaboration ({ISIC}).
\newblock In {\em 2018 IEEE 15th international symposium on biomedical imaging
  (ISBI 2018)}, pages 168--172. IEEE, 2018.

\bibitem[Cox72]{cox1972regression}
David~R Cox.
\newblock Regression models and life-tables.
\newblock {\em Journal of the Royal Statistical Society: Series B
  (Methodological)}, 34(2):187--202, 1972.

\bibitem[CVS{\etalchar{+}}13]{clark2013cancer}
Kenneth Clark, Bruce Vendt, Kirk Smith, John Freymann, Justin Kirby, Paul
  Koppel, Stephen Moore, Stanley Phillips, David Maffitt, Michael Pringle,
  et~al.
\newblock The cancer imaging archive ({TCIA}): maintaining and operating a
  public information repository.
\newblock {\em Journal of digital imaging}, 26(6):1045--1057, 2013.

\bibitem[DCM{\etalchar{+}}20]{dehaene2020self}
Olivier Dehaene, Axel Camara, Olivier Moindrot, Axel de~Lavergne, and Pierre
  Courtiol.
\newblock Self-supervision closes the gap between weak and strong supervision
  in histology.
\newblock {\em arXiv preprint arXiv:2012.03583}, 2020.

\bibitem[DG17]{dua2019Uci}
Dheeru Dua and Casey Graff.
\newblock {UCI} machine learning repository, 2017.

\bibitem[Dic45]{dice1945measures}
Lee~R Dice.
\newblock Measures of the amount of ecologic association between species.
\newblock {\em Ecology}, 26(3):297--302, 1945.

\bibitem[DR{\etalchar{+}}14]{dwork2014algorithmic}
Cynthia Dwork, Aaron Roth, et~al.
\newblock The algorithmic foundations of differential privacy.
\newblock {\em Found. Trends Theor. Comput. Sci.}, 9(3-4):211--407, 2014.

\bibitem[dt]{ixi}
Brain development team.
\newblock Ixi dataset.
\newblock \url{https://brain-development.org/ixi-dataset/}.
\newblock Accessed: 2022-02-02.

\bibitem[FCG{\etalchar{+}}21]{flamary2021pot}
Rémi Flamary, Nicolas Courty, Alexandre Gramfort, Mokhtar~Z. Alaya, Aurélie
  Boisbunon, Stanislas Chambon, Laetitia Chapel, Adrien Corenflos, Kilian
  Fatras, Nemo Fournier, Léo Gautheron, Nathalie~T.H. Gayraud, Hicham Janati,
  Alain Rakotomamonjy, Ievgen Redko, Antoine Rolet, Antony Schutz, Vivien
  Seguy, Danica~J. Sutherland, Romain Tavenard, Alexander Tong, and Titouan
  Vayer.
\newblock {POT}: {Python} optimal transport.
\newblock {\em Journal of Machine Learning Research}, 22(78):1--8, 2021.

\bibitem[FDZ{\etalchar{+}}15]{favazza2015cross}
Christopher~P Favazza, Xinhui Duan, Yi~Zhang, Lifeng Yu, Shuai Leng, James~M
  Kofler, Michael~R Bruesewitz, and Cynthia~H McCollough.
\newblock A cross-platform survey of ct image quality and dose from routine
  abdomen protocols and a method to systematically standardize image quality.
\newblock {\em Physics in Medicine \& Biology}, 60(21):8381, 2015.

\bibitem[Fee10]{feeman2010mathematics}
Timothy~G Feeman.
\newblock {\em The mathematics of medical imaging}.
\newblock Springer, 2010.

\bibitem[FMO20]{fallah2020personalized}
Alireza Fallah, Aryan Mokhtari, and Asuman Ozdaglar.
\newblock Personalized federated learning: A meta-learning approach.
\newblock {\em arXiv preprint arXiv:2002.07948}, 2020.

\bibitem[FZF{\etalchar{+}}20]{fatras_learning_2020}
Kilian Fatras, Younes Zine, R{\'e}mi Flamary, R{\'e}mi Gribonval, and Nicolas
  Courty.
\newblock {Learning with minibatch Wasserstein : asymptotic and gradient
  properties}.
\newblock In {\em {AISTATS 2020 - 23nd International Conference on Artificial
  Intelligence and Statistics}}, volume volume 108 of {\em PMLR}, pages 1--20,
  Palermo, Italy, June 2020.

\bibitem[GNS{\etalchar{+}}19]{DBLP:journals/corr/abs-1910-03910}
Nils Gessert, Maximilian Nielsen, Mohsin Shaikh, Ren{\'{e}} Werner, and
  Alexander Schlaefer.
\newblock Skin lesion classification using ensembles of multi-resolution
  efficientnets with meta data.
\newblock {\em CoRR}, abs/1910.03910, 2019.

\bibitem[Goo92]{good1992rational}
Irving~John Good.
\newblock Rational decisions.
\newblock In {\em Breakthroughs in statistics}, pages 365--377. Springer, 1992.

\bibitem[HIMH{\etalchar{+}}20]{heller2020state}
Nicholas Heller, Fabian Isensee, Klaus~H Maier-Hein, Xiaoshuai Hou, Chunmei
  Xie, Fengyi Li, Yang Nan, Guangrui Mu, Zhiyong Lin, Miofei Han, et~al.
\newblock The state of the art in kidney and kidney tumor segmentation in
  contrast-enhanced ct imaging: Results of the kits19 challenge.
\newblock {\em Medical Image Analysis}, page 101821, 2020.

\bibitem[HSK{\etalchar{+}}19]{heller2019kits19}
Nicholas Heller, Niranjan Sathianathen, Arveen Kalapara, Edward Walczak, Keenan
  Moore, Heather Kaluzniak, Joel Rosenberg, Paul Blake, Zachary Rengel, Makinna
  Oestreich, et~al.
\newblock The kits19 challenge data: 300 kidney tumor cases with clinical
  context, ct semantic segmentations, and surgical outcomes.
\newblock {\em arXiv preprint arXiv:1904.00445}, 2019.

\bibitem[HZRS16]{he2016deep}
Kaiming He, Xiangyu Zhang, Shaoqing Ren, and Jian Sun.
\newblock Deep residual learning for image recognition.
\newblock In {\em Proceedings of the IEEE conference on computer vision and
  pattern recognition}, pages 770--778, 2016.

\bibitem[IJK{\etalchar{+}}21]{isensee2021nnu}
Fabian Isensee, Paul~F Jaeger, Simon~AA Kohl, Jens Petersen, and Klaus~H
  Maier-Hein.
\newblock nnu-net: a self-configuring method for deep learning-based biomedical
  image segmentation.
\newblock {\em Nature methods}, 18(2):203--211, 2021.

\bibitem[ILTT11]{iglesias2011robust}
Juan~Eugenio Iglesias, Cheng-Yi Liu, Paul~M Thompson, and Zhuowen Tu.
\newblock Robust brain extraction across datasets and comparison with publicly
  available methods.
\newblock {\em IEEE transactions on medical imaging}, 30(9):1617--1634, 2011.

\bibitem[IMB{\etalchar{+}}15]{lidcdata}
S.~G.~Armato III, G.~McLennan, L.~Bidaut, M.~F. McNitt-Gray, C.~R. Meyer, A.~P.
  Reeves, B.~Zhao, D.~R. Aberle, C.~I. Henschke, E.~A. Hoffman, E.~A.
  Kazerooni, H.~MacMahon, E.~J. R.~Van Beek, D.~Yankelevitz, A.~M. Biancardi,
  P.~H. Bland, M.~S. Brown, R.~M. Engelmann, G.~E. Laderach, D.~Max, R.~C.
  Pais, D.~P.~Y. Qing, R.~Y. Roberts, A.~R. Smith, A.~Starkey, P.~Batra,
  P.~Caligiuri, A.~Farooqi, G.~W. Gladish, C.~M. Jude, R.~F. Munden,
  I.~Petkovska, L.~E. Quint, L.~H. Schwartz, B.~Sundaram, L.~E. Dodd,
  C.~Fenimore, D.~Gur, N.~Petrick, J.~Freymann, J.~Kirby, B.~Hughes, A.~V.
  Casteele, S.~Gupte ans M.~Sallam, M.~D. Heath, M.~H. Kuhn, E.~Dharaiya,
  R.~Burns, D.~S. Fryd, M.~Salganicoff, V.~Anand, U.~Shreter, S.~Vastagh, B.~Y.
  Croft, and L.~P. Clarke.
\newblock Data from lidc-idri [data set]. the cancer imaging archive., 2015.

\bibitem[ITW]{deepmil}
Maximilian Ilse, Jakub~M. Tomczak, and Max Welling.
\newblock Attention-based deep multiple instance learning.
\newblock \url{https://github.com/AMLab-Amsterdam/AttentionDeepMIL}.
\newblock Accessed: 2022-02-02.

\bibitem[Jad20]{jadon2020survey}
Shruti Jadon.
\newblock A survey of loss functions for semantic segmentation.
\newblock In {\em 2020 IEEE Conference on Computational Intelligence in
  Bioinformatics and Computational Biology (CIBCB)}, pages 1--7. IEEE, 2020.

\bibitem[Jen05]{jenkins2005survival}
Stephen~P Jenkins.
\newblock Survival analysis.
\newblock {\em Unpublished manuscript, Institute for Social and Economic
  Research, University of Essex, Colchester, UK}, 42:54--56, 2005.

\bibitem[JSPD88]{janosi1988heart}
Andras Janosi, William Steinbrunn, Matthias Pfisterer, and Robert Detrano.
\newblock Heart disease data set, 1988.

\bibitem[KB14]{kingma2014adam}
Diederik~P Kingma and Jimmy Ba.
\newblock Adam: A method for stochastic optimization.
\newblock {\em arXiv preprint arXiv:1412.6980}, 2014.

\bibitem[LBEB{\etalchar{+}}18]{litjens20181399}
Geert Litjens, Peter Bandi, Babak Ehteshami~Bejnordi, Oscar Geessink, Maschenka
  Balkenhol, Peter Bult, Altuna Halilovic, Meyke Hermsen, Rob van~de Loo, Rob
  Vogels, et~al.
\newblock 1399 {H}\&{E}-stained sentinel lymph node sections of breast cancer
  patients: the {CAMELYON} dataset.
\newblock {\em GigaScience}, 7(6):giy065, 2018.

\bibitem[LGG{\etalchar{+}}17]{DBLP:journals/corr/abs-1708-02002}
Tsung{-}Yi Lin, Priya Goyal, Ross~B. Girshick, Kaiming He, and Piotr
  Doll{\'{a}}r.
\newblock Focal loss for dense object detection.
\newblock {\em CoRR}, abs/1708.02002, 2017.

\bibitem[LLH{\etalchar{+}}18]{liu2018integrated}
Jianfang Liu, Tara Lichtenberg, Katherine~A Hoadley, Laila~M Poisson,
  Alexander~J Lazar, Andrew~D Cherniack, Albert~J Kovatich, Christopher~C Benz,
  Douglas~A Levine, Adrian~V Lee, et~al.
\newblock An integrated tcga pan-cancer clinical data resource to drive
  high-quality survival outcome analytics.
\newblock {\em Cell}, 173(2):400--416, 2018.

\bibitem[MAB]{histolab}
Alessia Marcolini, Ernesto Arbitrio, and Nicole Bussola.
\newblock https://histolab.readthedocs.io/en/latest/.
\newblock \url{https://histolab.readthedocs.io/en/latest/}.
\newblock Accessed: 2022-05-18.

\bibitem[MLI{\etalchar{+}}14]{itk}
Matthew McCormick, Xiaoxiao Liu, Luis Ibanez, Julien Jomier, and Charles
  Marion.
\newblock Itk: enabling reproducible research and open science.
\newblock {\em Frontiers in Neuroinformatics}, 8, 2014.

\bibitem[MNA16]{milletari2016v}
Fausto Milletari, Nassir Navab, and Seyed-Ahmad Ahmadi.
\newblock V-net: Fully convolutional neural networks for volumetric medical
  image segmentation.
\newblock In {\em 2016 fourth international conference on 3D vision (3DV)},
  pages 565--571. IEEE, 2016.

\bibitem[Mod]{niftyreg}
Marc Modat.
\newblock Nifty reg.
\newblock \url{https://sourceforge.net/p/niftyreg/git/ci/master/tree/}.
\newblock Accessed: 2022-02-02.

\bibitem[Net]{tcga}
TCGA~Research Network.
\newblock Tensorflow federated stack overflow dataset.
\newblock \url{https://www.cancer.gov/tcga}.
\newblock Accessed: 2022-05-18.

\bibitem[PG]{ixitiny}
Fernando Pérez-García.
\newblock Ixitiny dataset.
\newblock
  \url{https://torchio.readthedocs.io/datasets.html#torchio.datasets.ixi.IXITiny}.
\newblock Accessed: 2022-05-18.

\bibitem[PGM{\etalchar{+}}19]{paszke2019pytorch}
Adam Paszke, Sam Gross, Francisco Massa, Adam Lerer, James Bradbury, Gregory
  Chanan, Trevor Killeen, Zeming Lin, Natalia Gimelshein, Luca Antiga, et~al.
\newblock Pytorch: An imperative style, high-performance deep learning library.
\newblock {\em Advances in neural information processing systems}, 32, 2019.

\bibitem[RFB15]{ronneberger2015u}
Olaf Ronneberger, Philipp Fischer, and Thomas Brox.
\newblock U-net: Convolutional networks for biomedical image segmentation.
\newblock In {\em International Conference on Medical image computing and
  computer-assisted intervention}, pages 234--241. Springer, 2015.

\bibitem[TL19]{DBLP:journals/corr/abs-1905-11946}
Mingxing Tan and Quoc~V. Le.
\newblock Efficientnet: Rethinking model scaling for convolutional neural
  networks.
\newblock {\em CoRR}, abs/1905.11946, 2019.

\bibitem[TRK18]{tschandl2018ham10000}
Philipp Tschandl, Cliff Rosendahl, and Harald Kittler.
\newblock The {HAM10000} dataset, a large collection of multi-source
  dermatoscopic images of common pigmented skin lesions.
\newblock {\em Scientific data}, 5(1):1--9, 2018.

\bibitem[VDWCV11]{van2011numpy}
Stefan Van Der~Walt, S~Chris Colbert, and Gael Varoquaux.
\newblock The numpy array: a structure for efficient numerical computation.
\newblock {\em Computing in science \& engineering}, 13(2):22--30, 2011.

\bibitem[WLL{\etalchar{+}}20]{wei2020framework}
Wenqi Wei, Ling Liu, Margaret Loper, Ka-Ho Chow, Mehmet~Emre Gursoy, Stacey
  Truex, and Yanzhao Wu.
\newblock A framework for evaluating gradient leakage attacks in federated
  learning.
\newblock {\em arXiv preprint arXiv:2004.10397}, 2020.

\bibitem[YAG{\etalchar{+}}19]{yurochkin2019bayesian}
Mikhail Yurochkin, Mayank Agarwal, Soumya Ghosh, Kristjan Greenewald, Nghia
  Hoang, and Yasaman Khazaeni.
\newblock Bayesian nonparametric federated learning of neural networks.
\newblock In {\em International Conference on Machine Learning}, pages
  7252--7261. PMLR, 2019.

\bibitem[YSS{\etalchar{+}}21]{yousefpour2021opacus}
Ashkan Yousefpour, Igor Shilov, Alexandre Sablayrolles, Davide Testuggine,
  Karthik Prasad, Mani Malek, John Nguyen, Sayan Ghosh, Akash Bharadwaj,
  Jessica Zhao, et~al.
\newblock Opacus: User-friendly differential privacy library in pytorch.
\newblock {\em arXiv preprint arXiv:2109.12298}, 2021.

\end{thebibliography}
\end{document}